\documentclass[11pt]{article}

\usepackage{ifthen}
\newif\ifshortver

\usepackage{geometry}
\usepackage{soul}
\usepackage[normalem]{ulem}
\usepackage[makeroom]{cancel}

\usepackage{amsmath}
\usepackage{amssymb}
\usepackage{dsfont}
\usepackage{bbm}
\usepackage{graphicx}
\usepackage{float}
\usepackage{algorithmic}
\usepackage{algorithm}
\usepackage{listings,lstautogobble}
\usepackage{tikz}
\usepackage{courier}
\usetikzlibrary{decorations.markings}
\usepackage{graphicx}
\usepackage{color}
\usepackage{hyperref}
\usepackage{cleveref}
\hypersetup{
    colorlinks=true, 
    citecolor=black,
     linkcolor=blue,
     urlcolor=blue,
    linktoc=all,     
    linkcolor=black,  
}
\usepackage[round]{natbib}
\usepackage[nottoc]{tocbibind}  
\usepackage{caption}
\usepackage{subcaption}

\def\ddefloop#1{\ifx\ddefloop#1\else\ddef{#1}\expandafter\ddefloop\fi}
\def\ddef#1{\expandafter\def\csname bb#1\endcsname{\ensuremath{\mathbb{#1}}}}
\ddefloop ABCDEFGHIJKLMNOPQRSTUVWXYZ\ddefloop

\def\ddefloop#1{\ifx\ddefloop#1\else\ddef{#1}\expandafter\ddefloop\fi}
\def\ddef#1{\expandafter\def\csname frak#1\endcsname{\ensuremath{\mathfrak{#1}}}}
\ddefloop ABCDEFGHIJKLMNOPQRSTUVWXYZ\ddefloop

\def\ddefloop#1{\ifx\ddefloop#1\else\ddef{#1}\expandafter\ddefloop\fi}
\def\ddef#1{\expandafter\def\csname fr#1\endcsname{\ensuremath{\mathfrak{#1}}}}
\ddefloop ABCDEFGHIJKLMNOPQRSTUVWXYZ\ddefloop

\def\ddefloop#1{\ifx\ddefloop#1\else\ddef{#1}\expandafter\ddefloop\fi}
\def\ddef#1{\expandafter\def\csname eul#1\endcsname{\ensuremath{\EuScript{#1}}}}
\ddefloop ABCDEFGHIJKLMNOPQRSTUVWXYZ\ddefloop

\def\ddefloop#1{\ifx\ddefloop#1\else\ddef{#1}\expandafter\ddefloop\fi}
\def\ddef#1{\expandafter\def\csname scr#1\endcsname{\ensuremath{\mathscr{#1}}}}
\ddefloop ABCDEFGHIJKLMNOPQRSTUVWXYZ\ddefloop

\def\ddefloop#1{\ifx\ddefloop#1\else\ddef{#1}\expandafter\ddefloop\fi}
\def\ddef#1{\expandafter\def\csname b#1\endcsname{\ensuremath{\mathbf{#1}}}}
\ddefloop ABCDEFGHIJKLMNOPQRSTUVWXYZ\ddefloop

\def\ddefloop#1{\ifx\ddefloop#1\else\ddef{#1}\expandafter\ddefloop\fi}
\def\ddef#1{\expandafter\def\csname bhat#1\endcsname{\ensuremath{\hat{\mathbf{#1}}}}}
\ddefloop ABCDEFGHIJKLMNOPQRSTUVWXYZ\ddefloop

\def\ddefloop#1{\ifx\ddefloop#1\else\ddef{#1}\expandafter\ddefloop\fi}
\def\ddef#1{\expandafter\def\csname btil#1\endcsname{\ensuremath{\tilde{\mathbf{#1}}}}}
\ddefloop ABCDEFGHIJKLMNOPQRSTUVWXYZ\ddefloop

\def\ddefloop#1{\ifx\ddefloop#1\else\ddef{#1}\expandafter\ddefloop\fi}
\def\ddef#1{\expandafter\def\csname bst#1\endcsname{\ensuremath{\mathbf{#1}^\star}}}
\ddefloop ABCDEFGHIJKLMNOPQRSTUVWXYZ\ddefloop

\def\ddefloop#1{\ifx\ddefloop#1\else\ddef{#1}\expandafter\ddefloop\fi}
\def\ddef#1{\expandafter\def\csname bst#1\endcsname{\ensuremath{\mathbf{#1}^\star}}}
\ddefloop abcdeghijklmnopqrstuvwxyz\ddefloop

\def\ddefloop#1{\ifx\ddefloop#1\else\ddef{#1}\expandafter\ddefloop\fi}
\def\ddef#1{\expandafter\def\csname bhat#1\endcsname{\ensuremath{\hat{\mathbf{#1}}}}}
\ddefloop abcdefghijklmnopqrstuvwxyz\ddefloop

\def\ddefloop#1{\ifx\ddefloop#1\else\ddef{#1}\expandafter\ddefloop\fi}
\def\ddef#1{\expandafter\def\csname b#1\endcsname{\ensuremath{\mathbf{#1}}}}
\ddefloop abcdefghijklnopqrstuvwxyz\ddefloop

\def\ddefloop#1{\ifx\ddefloop#1\else\ddef{#1}\expandafter\ddefloop\fi}
\def\ddef#1{\expandafter\def\csname barb#1\endcsname{\ensuremath{\bar{\mathbf{#1}}}}}
\ddefloop abcdefghijklmnopqrstuvwxyz\ddefloop

\def\ddef#1{\expandafter\def\csname c#1\endcsname{\ensuremath{\mathcal{#1}}}}
\ddefloop ABCDEFGHIJKLMNOPQRSTUVWXYZ\ddefloop
\def\ddef#1{\expandafter\def\csname h#1\endcsname{\ensuremath{\widehat{#1}}}}
\ddefloop ABCDEFGHIJKLMNOPQRSTUVWXYZ\ddefloop
\def\ddef#1{\expandafter\def\csname hc#1\endcsname{\ensuremath{\widehat{\mathcal{#1}}}}}
\ddefloop ABCDEFGHIJKLMNOPQRSTUVWXYZ\ddefloop
\def\ddef#1{\expandafter\def\csname t#1\endcsname{\ensuremath{\widetilde{#1}}}}
\ddefloop ABCDEFGHIJKLMNOPQRSTUVWXYZ\ddefloop
\def\ddef#1{\expandafter\def\csname tc#1\endcsname{\ensuremath{\widetilde{\mathcal{#1}}}}}
\ddefloop ABCDEFGHIJKLMNOPQRSTUVWXYZ\ddefloop

\def\ddefloop#1{\ifx\ddefloop#1\else\ddef{#1}\expandafter\ddefloop\fi}
\def\ddef#1{\expandafter\def\csname rm#1\endcsname{\ensuremath{\mathrm{#1}}}}
\ddefloop ABCDEFGHIJKLMNOPQRSTUVWXYZabcdefghijklnopqrstuvwxyz\ddefloop

\def\ddefloop#1{\ifx\ddefloop#1\else\ddef{#1}\expandafter\ddefloop\fi}
\def\ddef#1{\expandafter\def\csname sf#1\endcsname{\ensuremath{\mathsf{#1}}}}
\ddefloop ABCDEFGHIJKLMNOPQRSTUVWXYZ\ddefloop

\title{{\Large\bfseries TOWARDS GUIDED DESCENT: OPTIMIZATION ALGORITHMS FOR TRAINING NEURAL NETWORKS AT SCALE\par}}
\date{\vspace{-1cm}}
\author{
    Ansh Nagwekar\\
    University of Pennsylvania\\
    \texttt{nagwekar@upenn.edu}
}

\usepackage{Shorthands}

\begin{document}
\maketitle













\vspace{0.5em}
\begin{center}
\large\textbf{Abstract}
\end{center}

Neural network optimization remains one of the most consequential yet poorly understood challenges in modern AI research, where improvements in training algorithms can lead to enhanced feature learning in foundation models, order-of-magnitude reductions in training time, and improved interpretability into how networks learn. While stochastic gradient descent (SGD) and its variants have become the de facto standard for training deep networks, their success in these overparameterized regimes often appears more empirical than principled. This thesis investigates this apparent paradox by tracing the evolution of optimization algorithms from classical first-order methods to modern higher-order techniques, revealing how principled algorithmic design can demystify the training process. Starting from first principles with SGD and adaptive gradient methods, the analysis progressively uncovers the limitations of these conventional approaches when confronted with anisotropy that is representative of real-world data. These breakdowns motivate the exploration of sophisticated alternatives rooted in curvature information: second-order approximation techniques, layer-wise preconditioning, adaptive learning rates, and more. Next, the interplay between these optimization algorithms and the broader neural network training toolkit, which includes prior and recent developments such as maximal update parametrization, learning rate schedules, and exponential moving averages, emerges as equally essential to empirical success. To bridge the gap between theoretical understanding and practical deployment, this paper offers practical prescriptions and implementation strategies for integrating these methods into modern deep learning workflows. The ultimate goal of this work is to highlight how designing optimization algorithms tailored toward neural networks work to illuminate the "black-boxes" of deep learning, and how many of these ideas are rooted in the mature fields of mathematical optimization and statistical learning theory.

\clearpage

\tableofcontents  

\clearpage

\section{Introduction}

The training of neural networks presents a fundamental tension: while modern deep learning achieves remarkable empirical success across domains from computer vision to natural language processing, the optimization algorithms that enable this success often seem more like carefully tuned recipes than principled mathematical procedures. The revival of deep neural networks over the past few decades, catalyzed by breakthroughs in convolutional architectures for vision and more recently by the dominance of Transformer-based models in language understanding, has only amplified this puzzle \citep{vaswani2017attention, krizhevsky2012imagenet}. How do we reconcile the fact that simple algorithms like SGD or Adam can train networks with billions of parameters, when classical optimization theory would predict failure in such massively non-convex landscapes \citep{duchi2011adaptive,li2017psgd}? This thesis confronts this paradox head-on, revealing that the apparent disconnect between theory and practice dissolves when we properly account for the unique structure of neural network optimization --- from the implicit geometry induced by network parameterization to the crucial role of adaptive preconditioning in navigating anisotropic loss surfaces.

Higher-order methods have been of great interest to the AI research community but are rarely spoken about in circles due to a variety of reasons. Some may say there's a high barrier to entry --- understanding the core underpinnings of neural network training requires a deep background in optimization theory, statistical learning, and numerical linear algebra. However, renewed interest in these methods is evident from initiatives like MLCommons' AlgoPerf benchmark, in which the higher-order optimizer ``Shampoo'' outperformed other algorithms in reaching a target performance level in the least amount of time  \citep{dahl2023algoperf,gupta2018shampoo}. Moreover, deeper research into the industry-leading optimizer Adam reveals core flaws: it suffers from poor generalization compared to SGD, sensitivity to hyperparameters across scales, and inability to exploit the natural geometry of neural network loss surfaces \citep{wilson2017marginal,choi2019empirical,martens2020ngm}. Some critiques argue that aggressive tuning is all you need to get Adam to work, or at the very least, match their performance on SGD and other momentum-based methods \citep{choi2019empirical}. However, even a perfectly tuned bicycle can't match the speed of a car --- true perhaps on certain routes, but missing the fundamental advantage of a different paradigm.

Kronecker-factored approximate curvature (KFAC) methods have demonstrated compelling advantages across diverse applications. In reinforcement learning, where sample efficiency is paramount, OpenAI's ACKTR algorithm achieved significant improvements over first-order methods in continuous control tasks, while  extended these benefits to value-based learning \citep{wu2017acktr,beltiukov2019qfac}. More recently, the transformer revolution has sparked renewed interest in scaling KFAC to attention-based architectures. Early work by \citet{zhang2019kfac} adapted KFAC to BERT pretraining, achieving 18\% wall-clock speedup over Adam. Subsequent efforts have pushed these boundaries further: \citet{pauloski2021kaisa} demonstrated distributed KFAC training on large-scale transformer models, \citet{osawa2022kfac} achieved competitive results on GPT-2 scale models, and \citet{grosse2023kfac} provided theoretical analysis of KFAC's behavior in the infinite-width limit of transformers. Most recently, \citet{eschenhagen2023kfac} developed a unified theoretical framework showing how KFAC can be systematically derived for any architecture expressible as linear layers with weight-sharing --- encompassing attention, convolutional, and graph neural network layers under a single formalism. These successes suggest that the additional computational cost of approximate second-order methods can be justified when the structure of the problem aligns with their theoretical advantages, particularly in scenarios where better per-iteration progress outweighs the overhead of curvature estimation.

Our journey begins with the foundations of neural network optimization, examining how momentum styles, (quasi-)Newton methods, and adaptive gradient algorithms evolved to address the limitations of vanilla SGD. We then dive deep into the world of curvature-aware optimization, where methods leverage structured approximations of second-order information to achieve both theoretical guarantees and practical speedups. We will then work on reframing these diverse algorithms through a recently-proposed unifying lens --- the modular norm framework --- showing how optimizer design reduces to choosing the right geometry for steepest descent for neural network loss functions. Next, we explore the broader ecosystem of training techniques that interact with these optimizers: maximal update parameterization for scale-invariant hyperparameter transfer, learning rate schedules that balance exploration and exploitation, exponential moving averages for stable gradient aggregation, and weight decay's careful role in maintaining optimization stability. To validate these theoretical insights and demonstrate their practical implications, we complement the literature survey with comprehensive experiments ranging from controlled studies on classic optimization landscapes to head-to-head comparisons on standard datasets, revealing how different algorithmic choices interact and exploring novel hybrid approaches that leverage multiple optimization paradigms. Through this systematic exploration, we aim to transform neural network optimization from an empirical art into a principled science, providing both theoretical insights and practical prescriptions for the next generation of deep learning systems.

\section{Classical Methods for Neural Network Optimization }\label{sec:classicdlo}

We begin our study by framing the problem of training neural networks in optimization-native language. Given a parametric model $f_{\mathbf{w}}: \mathcal{X} \to \mathcal{Y}$ with weights $\mathbf{w} \in \mathbb{R}^d$, our goal is to minimize a loss function $\ell: \mathcal{Y} \times \mathcal{Y} \to \mathbb{R}$ over a data distribution $\mathcal{D}$ on $\mathcal{X} \times \mathcal{Y}$, where $\mathcal{X}$ denotes the input space and $\mathcal{Y}$ the output space (labels for classification, real values for regression). In the stochastic optimization framework, we seek to minimize the expected risk $L(\mathbf{w}) = \mathbb{E}_{(\mathbf{x}, y) \sim \mathcal{D}}[\ell(f_{\mathbf{w}}(\mathbf{x}), y)]$, but we only have access to a finite training dataset $\mathcal{S} = \{(\mathbf{x}_i, y_i)\}_{i=1}^n$ sampled from $\mathcal{D}$. At each iteration $t$ over a training horizon $T$, we update the weights $\mathbf{w}_t$ using (noisy) gradient estimates computed on mini-batches $\mathcal{B}_t \subset \mathcal{S}$, where the stochastic gradient is $g_t = \frac{1}{|\mathcal{B}_t|} \sum_{(\mathbf{x}, y) \in \mathcal{B}_t} \nabla \ell(f_{\mathbf{w}_t}(\mathbf{x}), y)$ or simply just $\nabla \ell (\mathbf{w}_t)$. While our perspective focuses primarily on a single epoch (one complete pass through the dataset), in practice training involves multiple epochs, and the methods we discuss --- ranging from basic stochastic gradient descent (SGD) and steepest descent variants to momentum-based methods (Polyak's heavy ball, Nesterov's accelerated gradient), second-order methods (Newton's method, quasi-Newton methods), and adaptive learning rate algorithms (AdaGrad, RMSProp, Adam) --- form the foundation for understanding modern deep learning optimization. 

Throughout this, we maintain that while classical optimization theory provides essential intuition and convergence guarantees, the unique properties of neural network loss landscapes and the interplay between optimization and generalization necessitate careful empirical validation of theoretical prescription. Unlike optimization, the goal of deep learning can't be boiled down to simply finding the best solution under a single objective; instead, we must simultaneously optimize for training performance, generalization to unseen data, computational efficiency, robustness to distribution shift, and often additional constraints like fairness, interpretability, or deployment requirements.

\subsection{How to Descend?}

In classical optimization, the \emph{steepest descent} direction is defined as the update that achieves the maximal decrease in the objective function per unit step with respect to a chosen norm. Formally, given a differentiable loss $\ell(\mathbf{w})$ and a norm $\|\cdot\|$, the steepest descent direction at $\mathbf{w}_t$ is
\[
d_t = \arg\min_{\|d\| = 1} \langle \nabla \ell(\mathbf{w}_t), d \rangle.
\]
By duality of norms, this yields the update
\[
d_t = - \frac{\nabla \ell(\mathbf{w}_t)}{\|\nabla \ell(\mathbf{w}_t)\|_*},
\]
where $\|\cdot\|_*$ is the dual norm of $\|\cdot\|$. The notion of steepest descent is therefore \emph{norm-dependent}: in the $\ell_2$ norm it coincides with the negative gradient, while in other geometries, the direction may differ.

\begin{figure}
    \centering
    \includegraphics[width=0.5\linewidth]{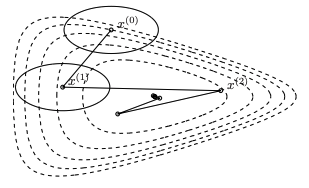}
    \caption{Figure 9.11 from \citep{boyd2004convex} depicting steepest descent. The ellipses represent the boundaries of the ``norm ball'' produced under the quadratic norm.}
    \label{fig:gd_boyd}
\end{figure}

Although steepest descent provides a principled way to define an``ideal'' update direction, it is often impractical to consider outside the Euclidean setting. The special case of the $\ell_2$ norm recovers the familiar gradient descent algorithm which has become the workhorse of large-scale optimization. This perspective situates gradient descent and its stochastic and momentum-based variants as descendants of steepest descent in the Euclidean norm. Classical gradient descent arises from optimization of a smooth convex function $\ell(\mathbf{w})$ by iteratively updating
\[
\mathbf{w}_{t+1} = \mathbf{w}_t - \eta \nabla \ell(\mathbf{w}_t),
\]
where $\eta > 0$ is the learning rate. For convex $\ell$, convergence to an $\epsilon$-neighborhood of the global optimum is guaranteed in $\mathcal{O}(1/\epsilon)$ iterations if $\eta \leq 1/L$, where $L$ is the Lipschitz constant of the gradient. For strongly convex functions, the rate improves to $\mathcal{O}(\kappa \log(1/\epsilon))$, with $\kappa = L/m$ the condition number \citep{boyd2004convex}.  

In large-scale machine learning, computing full gradients $\nabla \ell(\mathbf{w}_t)$ is costly. Instead, \emph{stochastic gradient descent} (SGD) replaces the full gradient with an unbiased estimate from a random mini-batch:
\[
\mathbf{w}_{t+1} = \mathbf{w}_t - \eta \nabla \ell_{i_t}(\mathbf{w}_t),
\]
where $\ell_{i_t}$ denotes the loss on a sampled data point (or small batch). Under diminishing learning rates, SGD converges to the optimum in expectation. The choice of learning rate is crucial: too large causes divergence, too small slows convergence.

\subsection{Momentum, Polyak’s Heavy Ball, and Nesterov's Method}
To accelerate GD, Polyak introduced the \emph{heavy ball method} which incorporates the first known idea of ``momentum'' \citep{polyak1964some}:
\[
\mathbf{w}_{t+1} = \mathbf{w}_t - \eta \nabla \ell(\mathbf{w}_t) + \rho (\mathbf{w}_t - w_{t-1}), \quad \rho \in (0,1).
\]
Here $\rho$ acts as a ``momentum'' coefficient, effectively giving the updates an ``inertia'' term. Intuitively, if successive gradients point in a consistent direction, the update gains extra speed; if not, momentum damps oscillations. Early works proved that this can substantially reduce the number of iterations: $\mathcal{O}(1/\sqrt{\epsilon})$ for convex functions, and $\mathcal{O}(\sqrt{\kappa}\log(1/\epsilon))$ for strongly convex functions.

Nesterov later proposed a refined method that improves stability near the optimum \citep{nesterov1983method}. Its update rule is
\[
u_{t+1} = \rho u_t - \nabla \ell(\mathbf{w}_t + \eta \rho u_t), \quad
\mathbf{w}_{t+1} = \mathbf{w}_t + \eta u_{t+1}.
\]
Unlike Polyak’s method, the gradient is evaluated at a \emph{look-ahead point} $\mathbf{w}_t + \eta \rho u_t$, which can be interpreted as computing the gradient at the anticipated future location with the belief that averaging the previous descent direction $u_t$ with the direction at this seemingly arbitrary jump in parameter space is more aligned with the ``optimal'' descent direction (the idea of optimality is vague in this context --- see section 2.4 for an in-depth discussion).

While this result has been evaluated empirically, classical convex analysis also shows that both Polyak’s and Nesterov’s methods achieve accelerated rates: $\mathcal{O}(1/\sqrt{\epsilon})$ in the convex case and $\mathcal{O}(\sqrt{\kappa}\log(1/\epsilon))$ in the strongly convex case.

We can also interpret momentum as performing averaging on the gradients as we descend. From an optimization-theoretic perspective, dual averaging (DA) maintains a running \emph{average} of past gradients and chooses the next iterate by solving a proximal subproblem in the \emph{primal} space \citep{nesterov2005primal}:
\[
\bar g_t \;=\; \tfrac{1}{t}\sum_{\tau=1}^t g_\tau,\qquad
\mathbf{w}_{t+1}\;=\;\arg\min_{w}\Big\{\langle \bar g_t,\, w\rangle \;+\; \tfrac{1}{\alpha_t}\,\lambda(\mathbf{w})\Big\},
\]
where $\lambda$ is a strongly convex regularizer (e.g., $\tfrac12\|w\|_2^2$) and $\alpha_t>0$ is a stepsize schedule \citep{nesterov2005primal, xiao2010dual}. DA can be read as a ``global momentum'' scheme: instead of the exponential moving average (EMA) used by Polyak and Nesterov, DA aggregates gradients uniformly over time and then takes a single steepest-descent step. For an in-depth discussion on the roles of EMA and momentum in modern deep learning, skip to section \ref{sec:ema}. While momentum remains crucial to deep learning, its exact usage still remains varied and open. Whether or not to perform momentum updates using look-ahead points or to simply track prior gradient information using a moving average is still theoretically open; modern momentum schemes focus on reducing the computational memory footprint and opt for EMA solutions. 

\subsection{Newton’s Method and Quasi-Newton Methods}

\begin{figure}
    \centering
    \includegraphics[width=0.5\linewidth]{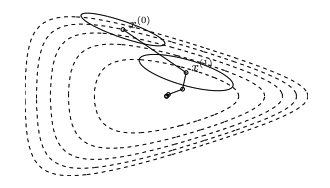}
    \caption{Figure 9.19 from \citep{boyd2004convex}. Here, the elipses look curvature-aware --- they seem to mold their shape based the local geometry of the loss function.}
    \label{fig:steepest_desc}
\end{figure}

Gradient descent updates scale the step size uniformly across all directions, which can have clear disadvantages. Consider the simple quadratic bowl in Figure \ref{fig:gd_boyd}: gradient descent zigzags slowly along a narrow valley, taking many iterations to converge. Now consider the same landscape when Newton's method performs descent in Figure \ref{fig:steepest_desc}: a method that accounts for curvature can proceed quickly toward the minimum. 

As hinted by the figures, convex optimization theory has long proposed superior methods for descent: curvature information, or what is usually referred to as second-order information in vector calculus, or simply acceleration from the physical laws of motion. Newton's method instead rescales, or "preconditions," the gradient using curvature information:
\[
\mathbf{w}_{t+1} = \mathbf{w}_t - \eta_t [\nabla^2 \ell(\mathbf{w}_t)]^{-1} \nabla \ell(\mathbf{w}_t),
\]
where $\eta_t$ was classically interpreted as the step size resulting from a line search (which essentially systematically does a local search for the best possible loss-decreasing descent step). This is motivated by the fact that computing and inverting the Hessian $[\nabla^2 \ell(\mathbf{w}_t)]^{-1}$ is so computationally heavy that conducting a line search of small order doesn't affect the runtime significantly. The Newton step can be interpreted as minimizing a second-order Taylor expansion of $\ell$ around $\mathbf{w}_t$ or as solving the linearized optimality condition $\nabla \ell(\mathbf{w}_t + \mathbf{v}) \approx \mathbf{0}$ with respect to $\mathbf{v}$. For a much deeper look into Newton's method and proof of convergence, please see the canonical Convex Optimization course by \citet{boyd2004convex}.

Newton's method enjoys \emph{quadratic convergence} near a minimizer: once close, the number of accurate digits roughly doubles with each iteration. In contrast to the $\mathcal{O}(1/\epsilon)$ or $\mathcal{O}(\kappa \log(1/\epsilon))$ rates of first-order methods, Newton's method typically requires only $\mathcal{O}(\log \log (1/\epsilon))$ iterations to achieve $\epsilon$-accuracy. Preconditioning here is crucial --- by scaling each gradient direction by the inverse curvature, Newton's method avoids slow progress in flat directions and overshooting in steep ones. Moreover, Newton's method is attractive from a theoretical standpoint as it is affine-invariant, meaning the optimized solution is independent of linear change of coordinates \citep{boyd2004convex}.

An important variant arises when we consider the structure of statistical learning problems. When the loss is chosen to be the negative log-likelihood $\ell(\mathbf{w}) = -\log p(y|\mathbf{x}; \mathbf{w})$ in a convex setting, the Hessian of the loss function (over the full data distribution) is equivalent to the Fisher information matrix:
$$\mathbf{F}_t = \mathbb{E}_{y \sim p(y|\mathbf{x}; \mathbf{w}_t)}[\nabla \log p(y|\mathbf{x}; \mathbf{w}_t) \nabla \log p(y|\mathbf{x}; \mathbf{w}_t)^\top]$$
Thus, when approaching a statistical learning problem, using the Fisher information as the matrix to invert and scale the gradient at the current weights by is know as \emph{natural gradient descent}  (NGD) \citep{amari1998natural}. The natural gradient $\mathbf{F}_t^{-1} g_t$ represents the steepest descent direction in the space of probability distributions (measured by KL divergence) rather than in Euclidean parameter space which coincides in these special cases. This is the theoretically ``correct'' gradient for statistical models as it optimally estimates the maximum likelihood that best describes the data distribution.

The connection between natural gradient descent and Newton's method in high-dimension deep learning is important: when the model is well-specified and near convergence, these matrices approximately coincide, making natural gradient descent behave similarly to Newton's method. However, a crucial practical distinction emerges: the Fisher is defined as an expectation over the model's predicted distribution, while in practice we compute an \emph{empirical Fisher} using the empirical data distribution. The choice between the theoretical Fisher (expectation under model predictions) and empirical Fisher (using actual labels) can significantly impact the theoretical analysis following the resulting algorithm's behavior, but in an empirical setting, we have seen promising results from using the empirical Fisher as specified. We defer a proper treatment of these distinctions and their implications for deep learning in Section \ref{sec:fisher}, where we discuss curvature-aware optimization in depth.

Despite significantly reducing the number of iterations needed to converge, Newton's method in practice can be intractable for modern machine learning models as computing and inverting the Hessian is computationally expensive. We present some practical alternatives that have been discussed for decades.

\paragraph{Hessian-free optimization (HF).} Martens’ work on HF optimization demonstrated that by approximating Hessian-vector products through finite differences of gradients and solving the resulting linear systems using conjugate gradient (CG) methods, one can achieve effective second-order updates with only first-order computational cost \citep{martens2010deep, nocedal2006numerical}. At iteration $t$, the update direction $\mathbf{p}_t$ is found as the solution to:
\[
\left( H(\mathbf{w}_t) + \lambda I \right)\mathbf{p}_t = -\nabla f(\mathbf{w}_t)
\]
where:
\begin{itemize}
    \item $H(\mathbf{w}_t)$ is the Hessian of the objective function at parameters $\mathbf{w}_t$,
    \item $\lambda$ is a damping parameter, added for numerical stability and regularization,
    \item $\nabla f(\mathbf{w}_t)$ is the gradient at $\mathbf{w}_t$.
\end{itemize}
The Hessian-vector product $H(\mathbf{w}_t)\mathbf{v}$ is approximated efficiently as:
\[
H(\mathbf{w}_t)\mathbf{v} \approx \frac{\nabla f(\mathbf{w}_t + \epsilon \mathbf{v}) - \nabla f(\mathbf{w}_t)}{\epsilon}
\]
where $\epsilon$ is a small scalar. The update rule is then:
\[
\mathbf{w}_{n+1} = \mathbf{w}_t + \mathbf{p}_t
\]
with $\mathbf{p}_t$ found by running CG to minimize the quadratic model:
\[
q_{\mathbf{w}_t}(\mathbf{p}) = \nabla f(\mathbf{w}_t)^{T}\mathbf{p} + \frac{1}{2}\mathbf{p}^{T}H(\mathbf{w}_t)\mathbf{p} + \frac{\lambda}{2}\mathbf{p}^{T}\mathbf{p}
\]
This approach is especially well suited for deep neural networks, where the loss surface’s curvature varies dramatically across layers. By using damping and structural approximations, the Hessian-free method can adaptively correct ill-conditioning while avoiding direct matrix storage, making it a scalable substitute for full Newton updates.

\paragraph{BFGS Family.} The BFGS algorithm is arguably the most widely used quasi-Newton method. The core idea is elegant: instead of computing the true Hessian, BFGS maintains an approximation $B_k$ that is updated at each iteration to satisfy the \textit{secant equation} (also called the quasi-Newton condition):
\[
B_{k+1} \Delta \mathbf{w}_k = \mathbf{y}_k
\]
where $\Delta \mathbf{w}_k = \mathbf{w}_{k+1} - \mathbf{w}_k$ and $\mathbf{y}_k = \nabla f(\mathbf{w}_{k+1}) - \nabla f(\mathbf{w}_k)$. This condition ensures that the approximate Hessian correctly captures the curvature along the most recent search direction. BFGS updates the inverse Hessian approximation $H_k \approx B_k^{-1}$ directly via a rank-two update:
\[
H_{k+1} = \left(I - \rho_k \Delta \mathbf{w}_k \mathbf{y}_k^\top \right) H_k \left(I - \rho_k \mathbf{y}_k \Delta \mathbf{w}_k^\top \right) + \rho_k \Delta \mathbf{w}_k \Delta \mathbf{w}_k^\top
\]
where $\rho_k = 1 / (\mathbf{y}_k^\top \Delta \mathbf{w}_k)$. This rank-two structure is the minimal update that preserves both symmetry and positive-definiteness of the Hessian approximation. The parameter update then follows the Newton-like form:
\[
\mathbf{w}_{k+1} = \mathbf{w}_k - \eta H_k \nabla f(\mathbf{w}_k)
\]
where $\eta$ is typically determined by a line search. BFGS achieves superlinear convergence on convex problems while requiring only $\mathcal{O}(n^2)$ storage and computation per iteration --- a substantial improvement over Newton's $\mathcal{O}(n^3)$ cost, though still quadratic in the number of parameters.

For large-scale problems where even $\mathcal{O}(n^2)$ storage becomes prohibitive, the Limited-memory BFGS (L-BFGS) method provides a practical alternative \citep{liu1989limited}. Instead of storing the full $n \times n$ inverse Hessian approximation, L-BFGS maintains only the $m$ most recent pairs $\{(\Delta \mathbf{w}_j, \mathbf{y}_j)\}_{j=k-m}^{k-1}$, where $m$ is typically small (e.g., $m = 10$ to $20$). The key insight is that the product $H_k \nabla f(\mathbf{w}_k)$ can be computed implicitly through the \textit{two-loop recursion} --- a sequence of vector operations that applies the stored curvature corrections to the gradient without ever forming $H_k$ explicitly. This reduces storage from $\mathcal{O}(n^2)$ to $\mathcal{O}(mn)$ and computation to $\mathcal{O}(mn)$ per iteration, making quasi-Newton optimization feasible for problems with millions of parameters. In essence, L-BFGS approximates the effect of multiplying by the inverse Hessian by successively adjusting the gradient using curvature information from recent steps, producing a descent direction that accounts for both magnitude and orientation of past curvature. 

In deep learning, L-BFGS has shown mixed but instructive results --- while it can achieve smoother convergence and better conditioning for smaller networks or fine-tuning stages, its batch dependencies and memory demands limit its practicality at scale. Nevertheless, both methods exemplify how classical curvature approximations can bridge the gap between first-order and second-order optimization, preserving much of Newton’s acceleration with dramatically reduced runtime expense.

\paragraph{Hybrid CG-BFGS Methods.} Buckley and LeNir proposed an innovative hybrid that combines BFGS updates with conjugate gradient (CG) steps to further reduce memory requirements while retaining quasi-Newton efficiency \citep{liu1989limited}. The method operates in cycles: it begins by performing standard BFGS updates, but stores the corrections to an initial matrix separately rather than forming the full $\mathcal{O}(n^2)$ approximation. When the allocated storage is exhausted, the current BFGS matrix is frozen as a fixed preconditioner, and the method switches to preconditioned conjugate gradient steps. These CG iterations continue until Powell's restart criterion --- based on the observation that gradients are orthogonal for quadratic objectives under exact line search --- indicates that a restart is beneficial. At this point, all BFGS corrections are discarded and a new BFGS cycle begins. This approach generalizes earlier work by Shanno and allows flexible memory allocation: with minimal storage it behaves like preconditioned CG, while with ample storage it approaches full BFGS performance. The method is regarded as effective for large-scale optimization where memory constraints preclude standard quasi-Newton approaches.

Despite their theoretical elegance and success in classical optimization, the BFGS family has shown mixed results in deep learning. Several factors limit their practicality: (1) the non-convex loss landscapes of neural networks violate the positive-definiteness assumptions underlying quasi-Newton convergence guarantees; (2) stochastic gradients, essential for scaling to large datasets, introduce noise that corrupts the curvature estimates built from gradient differences; (3) even L-BFGS's $\mathcal{O}(mn)$ storage becomes burdensome for models with billions of parameters; and (4) the line search procedures typically required for quasi-Newton convergence add computational overhead incompatible with mini-batch training. Nevertheless, these methods remain valuable for fine-tuning stages, smaller networks, and full-batch optimization scenarios. More importantly, the BFGS family exemplifies how principled curvature approximations can bridge the gap between first-order and second-order optimization --- a theme that resurfaces in modern methods like KFAC and Shampoo, which adapt these ideas to the structure of neural networks.

\subsection{Improvements to Stochastic Gradient Descent}

While standard SGD remains a cornerstone of deep learning optimization, it often discovers solutions that don't always generalize well and can be sensitive to data perturbations. Recent work has drawn from optimization theory and classical machine learning to motivate several algorithmic refinements that explicitly target the geometric properties of the loss landscape.

\begin{figure}
    \centering
    \includegraphics[width=0.5\linewidth]{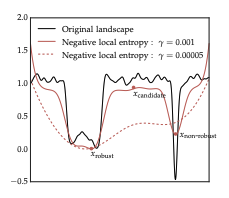}
    \caption{Figure 2 from \citep{chaudhari2017entropy} depicting how local entropy tend to gather around wider valleys and away from narrow regions, corresponding to the robustness of the solution.}
    \label{fig:entropy_descent}
\end{figure}

\paragraph{Entropy-SGD.} One such approach to improving SGD emerges from the observation that well-generalizing solutions tend to occupy large flat regions of the energy landscape --- Entropy-SGD addresses this insight by reformulating the optimization problem through the lens of statistical physics \citep{chaudhari2017entropy}. The authors of the Entropy-SGD paper start by arguing that vanilla SGD often discovers solutions at sharp minima that generalize poorly. Entropy-SGD addresses this by reformulating the optimization problem to exploit landscape geometry. Rather than minimizing training loss directly, it minimizes a local entropy objective:
\[
F(x, \gamma) = \log \int_{x'} \exp\left(-f(x') - \frac{\gamma}{2}\|x - x'\|_2^2\right) dx'
\]
This objective concentrates probability mass on flat regions of the loss landscape, where well-generalizing solutions tend to reside. The hyperparameter $\gamma$ controls the scope of exploration. The intuition here is key: under a Bayesian prior on the parameters, wide valleys have higher marginalized likelihood than sharp, isolated minima even if the latter achieve lower training loss (see Figure \ref{fig:entropy_descent}). By biasing optimization toward these flat regions, the algorithm inherently favors solutions with better generalization properties while avoiding narrow minima that are brittle to input and parameter perturbations.

Specifically, the algorithm is structured such that an outer loop performs parameter updates while an inner loop uses stochastic gradient Langevin dynamics (SGLD) to estimate the entropy gradient.
The inner loop samples from a locally-focused Gibbs distribution using mini-batch gradient estimates and Langevin noise. The outer loop then steps toward the estimated local expectation. This architecture creates an implicit averaging effect that stabilizes optimization and admits connections to regularization schemes. For more information on implementation details, see the original paper \citep{chaudhari2017entropy}.

Crucially, Entropy-SGD prioritizes \emph{generalization} over training loss. Empirically, it achieves comparable or superior test error despite higher training loss while also achieving large speedups over standard SGD on RNNs. Theoretically, the local entropy objective smooths the loss landscape by a factor controlled by $\gamma$, improving generalization error bounds via uniform stability analysis.

\paragraph{Elastic Averaging SGD (EA-SGD).} Another idea is to try and parallelize SGD and develop a scheme to use averaged results from the learned weights. The central idea from optimization theory that this draws from is ``alternating direction method of multipliers'' (ADMM) \citep{boyd2010admm}. ADMM is a decomposition-coordination algorithm that solves convex optimization problems of the form minimize $f(x) + g(z)$ subject to $Ax + Bz = c$ by alternately minimizing the augmented Lagrangian over $x$ and $z$ while updating dual variables. The algorithm blends the decomposability of dual ascent methods with the superior convergence properties of augmented Lagrangian methods, making it well-suited for distributed optimization where large problems can be split into smaller subproblems solved in parallel. ADMM has proven particularly effective for machine learning and statistical problems involving loss functions coupled with regularization terms, such as the lasso, sparse inverse covariance selection, and consensus problems across multiple processors.

EA-SGD tackles parallel SGD training by enabling multiple worker processes to maintain local parameters while coordinating through an elastic force linking them to a center variable \citep{zhang2015Eeasgd}:
\[
x_t^{i+1} = x_t^i - \eta(\nabla f(x_t^i) + \rho(x_t^i - \tilde{x}_t))
\]
The center variable is updated as a moving average in both time and space:
\[
\tilde{x}_{t+1} = (1-\beta)\tilde{x}_t + \beta \frac{1}{p}\sum_{i=1}^p x_t^i
\]
where $\beta = p\alpha$ and $\alpha = \eta\rho$ create ``elastic symmetry.'' The penalty parameter $\rho$ controls exploration versus exploitation: small $\rho$ allows local workers to explore further from the center, discovering different regions of the loss landscape. In deep learning, where many local optima exist, this exploration can lead to better generalization.

A key advantage of EA-SGD is stability under asynchronous updates and infrequent communication. Unlike ADMM, which can exhibit chaotic behavior in parallel round-robin schemes, EA-SGD stability depends on simple conditions: $0 \leq \eta \leq 2$ and $0 \leq \alpha \leq \frac{4-2\eta}{4-\eta}$, independent of the number of workers. This decoupling makes EA-SGD scalable and robust to communication delays. Experiments on CIFAR-10 and ImageNet with up to 16 workers show that EA-SGD significantly outperforms ``DOWNPOUR'' and other parallelized gradient descent baselines, particularly when communication periods between workers are large (e.g. 16, 64). The algorithm achieves faster convergence, better test error, and reduced communication overhead, demonstrating that allowing local exploration through less frequent synchronization can improve both generalization and efficiency in distributed deep learning.

\paragraph{Equivalence under Stochastic Homogenization.} As we can see, classical ideas from optimization theory have motivated algorithmic refinements to standard SGD that explicitly exploit the geometric structure of deep learning loss landscapes. Entropy-SGD addresses the challenge of discovering solutions that generalize well by reformulating optimization to maximize local entropy --- favoring large flat regions of the landscape over sharp isolated minima --- using an efficient nested SGD structure in the inner loop. Parallel and distributed training presents complementary challenges that EA-SGD tackles through elastic forces linking local worker parameters to a center variable, enabling effective exploration of different regions of the loss landscape while reducing communication overhead. 

Remarkably, \citet{chaudhari2017deeprelax} proved these two seemingly disparate approaches to be mathematically equivalent under the framework of stochastic homogenization. The key insight comes from analyzing systems of stochastic differential equations with multiple time-scales. In Entropy-SGD, a fast inner loop samples from a locally-focused distribution while a slow outer loop performs parameter updates. Through homogenization --- a technique for analyzing dynamical systems with ``fast'' and ``slow'' variables --- the authors rigorously prove that as the time scales separate (formally, as $\varepsilon \to 0$ in their formulation), the Entropy-SGD dynamics converge to gradient descent on the local entropy function $f_\gamma(x)$. For a proper treatment, please see the original paper \citep{chaudhari2017deeprelax}.

Roughly, for EA-SGD, the objective can be written as minimizing over $n$ workers:
\[
\min_x \frac{1}{n} \sum_{i=1}^{n} \left( f(x_i) + \frac{1}{2\gamma}|x_i - \bar{x}|^2 \right)
\]
where $\bar{x}$ is the average across workers. Each worker $i$ maintains local parameters that are coupled to the center variable through an elastic penalty term. The equivalence emerges through ergodicity: the temporal averaging performed by Entropy-SGD's inner loop can be replaced by spatial averaging across EA-SGD's multiple workers. Specifically, both methods minimize variants of the same underlying local objective
\[
H(x, y; \gamma) = f(y) + \frac{1}{2\gamma}|x - y|^2
\]
but access it through different computational paths --- one through time (Entropy-SGD's inner loop iterations) and one through space (EA-SGD's parallel workers). This equivalence explains why both methods exhibit similar generalization properties and why the $\rho$ parameter in EA-SGD plays an analogous role to $\gamma$ in Entropy-SGD: both control the strength of exploration around the current solution.

Furthermore, through a stochastic control interpretation, \citet{chaudhari2017deeprelax}\ prove that local entropy dynamics improve the expected value of the loss function compared to standard SGD (Theorem 12), providing theoretical justification for the empirical success of both approaches. In the paper, ``semiconcavity'' estimates show that the smoothing is controlled by $\gamma$ and that high-curvature local minima widen faster than flat ones under this evolution, which, under the arguments behind Entropy-SGD, directly relates to improved generalization properties. These advances underscore a broader principle: in deep learning, the choice of what objective to optimize proves as important as the choice of how to optimize it.

\paragraph{Preconditioned Stochastic Gradient Descent.} While Entropy-SGD and EA-SGD exploit landscape geometry through indirect mechanisms, preconditioned SGD approaches the problem more directly by incorporating curvature information of the loss landscape into the optimization dynamics \citep{li2017psgd}. Preconditioned SGD replaces the standard update rule with:
\[
\mathbf{w}_{t+1} = \mathbf{w}_t - \eta_t P_t g_t
\]
where $P_t$ is a symmetric positive definite preconditioner that scales the stochastic gradient adaptively based on curvature information. The key insight is that the preconditioner should reduce the ``eigenvalue spread'' of the Hessian-preconditioner product $P_t H_t$ while simultaneously normalizing ``eigenvalue amplitudes'' to near unity and suppressing gradient noise.

The paper proposes three preconditioner estimation criteria, of which criterion 3 proves most effective for general non-convex optimization:
\[
c_3(P_t) = \mathbb{E}[\delta \hat{g}_t^T P_t \delta \hat{g}_t + \delta \mathbf{w}_t^T P_t^{-1} \delta \mathbf{w}_t]
\]
where $\delta \hat{g}_t = g_t(\mathbf{w}_t + \delta \mathbf{w}_t) - g_t(\mathbf{w}_t)$ denotes the difference in stochastic gradients evaluated at perturbed parameters, with both gradients computed on the same mini-batch to reduce noise. This criterion ensures that preconditioned gradient perturbations match parameter perturbations in amplitude, scaling the stochastic gradient in a manner comparable to Newton's method without requiring explicit Hessian computation. Crucially, the optimal preconditioner $P_t^*$ satisfies $P_t^* R_{g,t} P_t^* = R_{\mathbf{w},t}$, where $R_{g,t} = \mathbb{E}[\delta \hat{g}_t \delta \hat{g}_t^T]$ and $R_{\mathbf{w},t} = \mathbb{E}[\delta \mathbf{w}_t \delta \mathbf{w}_t^T]$ are empirical covariance matrices estimated from these stochastic gradient differences. This early paper touches on the application of curvature matrices to adapt gradient steps, which as we will explore is a central part to understanding the future of neural network training algorithms.

The method exhibits three desirable properties: reduced eigenvalue spread for uniform convergence rates, normalized eigenvalue amplitudes enabling fixed step size selection without tuning, and natural gradient noise damping through the preconditioner structure. Through Cholesky factorization $P_t = Q_t^T Q_t$ and relative (natural) gradient descent updates to the triangular factor $Q_t$, the algorithm remains computationally tractable even for high-dimensional problems while inheriting equivariance properties that ensure robustness across different numerical scales.

Experimental results demonstrate that criterion 3 succeeds on challenging problems where standard SGD fails --- including learning long-term dependencies in RNNs and training on ill-conditioned problems --- by directly addressing the role of curvature. This approach sets the stage for later sections, which will establish formal connections between preconditioning via curvature matrices, generalization guarantees, and efficient feature learning rooted in second-order optimization theory and information geometry.

\subsection{Adaptive Methods for Neural Networks}

Building off the idea of preconditioners, we now switch gears to adaptive methods, which focus on scaling each parameter with element-wise operations instead of storing an entire preconditioner matrix $P_t \in \mathbb{R}^{d \times d}$. While this simplification dramatically reduces memory from $\mathcal{O}(d^2)$ to $\mathcal{O}(d)$ and computation from matrix-vector products to coordinate-wise multiplications, it retains an interesting connection: these diagonal scaling factors can be interpreted as approximating the diagonal of a curvature matrix. This approximation strategy, though seemingly crude, proves remarkably effective in practice and will be examined extensively in Section \ref{sec:curvmat}, where we explore the theoretical underpinnings of diagonal preconditioning and its relationship to full second-order methods.

Adaptive methods form the backbone of modern deep learning optimization. Their appeal lies in several practical properties: each iteration requires only gradient information and cheap operations on it, making updates computationally cheap; the operations naturally parallelize across large batches and distributed hardware; memory requirements are modest compared to second-order methods; and the resulting updates empirically perform well on foundational model training in many domains. The most popular method is Adam(W), which is motivated by a variety of empirical results, and many theorists have considered theoretical justifications post-hoc.

\subsubsection{AdaGrad + RMSProp = Adam}
Concretely, two adaptive first-order methods motivated the development of Adam: AdaGrad, which accumulates squared gradients to scale learning rates and thereby adapts to the geometry of the data, and RMSProp, which uses an exponential moving average of squared gradients to avoid AdaGrad’s rapid decay in effective learning rate \citep{duchi2011adaptive,tieleman2012lecture}.

\paragraph{AdaGrad} was originally conceived as a method for second moment estimation of the stochatic gradient, but due to computational restrictions, its diagonal counterpart (which only captures interactions between the gradient terms and themselves) became widely popular \citep{duchi2011adaptive}. For a proper discussion on the role of the AdaGrad matrix, see Section \ref{sec:adagrad}. For now, we can view the AdaGrad matrix as a method to accumulate the squared gradients of each parameter in order to rescale future updates. For parameters $\theta \in \mathbb{R}^d$ and gradient vector $g_t = \nabla \ell (\mathbf{w}_t)$, the update is
\[
v_t = \sum_{i=1}^t g_i \odot g_i, \quad v_t \in \mathbb{R}^d 
\]
\[
\mathbf{w}_{t+1} = \mathbf{w}_t - \eta \, v_t^{-1/2} \odot g_t
\]
where $\odot$ denotes element-wise operations (also known as Hadamard product). Intuitively, coordinates that have seen large gradients receive smaller effective step sizes, while infrequently updated coordinates retain larger step sizes. This makes AdaGrad particularly effective in sparse settings such as natural language processing. However, since $v_t$ grows monotonically, learning rates shrink continuously, which often leads to premature convergence.

\paragraph{RMSProp} addresses AdaGrad’s overly aggressive learning-rate decay by replacing the cumulative sum with an exponential moving average of squared gradients, thus stabilizing updates while maintaining adaptivity \citep{tieleman2012lecture}:
\[
v_t = \beta v_{t-1} + (1 - \beta) \, (g_t \odot g_t)
\]
Kingma and Ba observed that the eigenspectrum of weight updates in deep neural networks exhibits heavy tails and anisotropy, suggesting that an effective optimizer should combine per-parameter adaptivity with momentum. They theorized that an optimizer that maintains an exponential moving averaging on both the first and second moments of the gradient should lead to stable updates while adapting step sizes across coordinates based on gradient magnitude.

\paragraph{Adam} maintains exponential moving averages of both the first moment (mean) and second moment (uncentered variance) of the gradient, with bias corrections to account for initialization \citep{kingma2015adam}:
\[
m_t = \beta_1 m_{t-1} + (1-\beta_1) g_t
\]
\[
v_t = \beta_2 v_{t-1} + (1-\beta_2) g_t^2
\]
\[
\hat{m}_t = \frac{m_t}{1 - \beta_1^t}
\]
\[
\hat{v}_t = \frac{v_t}{1 - \beta_2^t}
\]
\[
\mathbf{w}_{t+1} = \mathbf{w}_t - \eta \frac{\hat{m}_t}{\sqrt{\hat{v}_t} + \epsilon}
\]
Here $m_t$ and $v_t$ are the moving averages, $\beta_1 \approx 0.9$, $\beta_2 \approx 0.999$, and $\epsilon$ is a small constant for numerical stability (note that this was used in RMSProp but omitted for brevity). Adam thus combines momentum with adaptive per-coordinate learning rates, making it robust and efficient in practice. Note that setting $\beta_1 = 0$ recovers RMSProp.
 
\paragraph{AdamW} introduces one critical modification: decoupling weight decay from the adaptive update \citep{loshchilov2019decoupled}. In Adam, $\ell_2$ regularization was intertwined with the adaptive denominator, which distorted the effect of weight decay. AdamW applies decay directly to the parameters:
\[
\mathbf{w}_{t+1} = (1 - \eta \lambda)\mathbf{w}_t - \eta \frac{\hat{m}_t}{\sqrt{\hat{v}_t} + \epsilon},
\]
where $\lambda$ is the weight decay coefficient. This simple adjustment has a large effect on generalization, and AdamW is now the standard choice in large-scale deep learning.

Adam(W) is widely deployed in industrial training pipelines due to its robustness and efficiency. Implementation optimizations include fused CUDA kernels for vectorized moment updates, mixed-precision arithmetic for speed and memory savings, and memory-efficient variants designed for large models. Hyperparameter choices remain critical: subsequent work has analyzed how tuning these values affects stability and generalization \citep{defazio2025gradients}.

Adam’s empirical effectiveness has been most visible in natural language processing and large sequence models. In recurrent neural networks, its adaptive per-parameter scaling helps mitigate exploding and vanishing gradients, leading to more stable long-horizon training. For transformer-based architectures, Adam(W) has become the de facto choice. \citet{orvieto2025adamsecret} argue that Adam’s ``secret sauce'' lies in the interplay between bias correction and adaptive scaling, which aligns well with the heavy-tailed gradient noise observed in large models. \citet{zhou2019adaptive} provide complementary evidence in the context of attention models, showing that Adam’s coordinate-wise normalization is particularly well suited to highly anisotropic curvature, thereby accelerating convergence and improving stability \citep{zhou2019adaptive}.

Alternative approaches to momentum-fused averaging schemes such as ``SignSGD'' highlight that precise scales for parameter updates is not strictly necessary for progress \citep{bernstein2019signsgd}. Observe its update rule:
\[
\mathbf{w}_{t+1} = \mathbf{w}_t - \eta *\text{sign}(\hat{m}_t)
\]
However, these methods often underperform well-tuned Adam on large-scale NLP and transformer tasks. \citet{zhang2024transformers} provide a geometric explanation for this gap through the lens of Hessian structure: Transformers exhibit \emph{block-heterogeneous} Hessians at initialization, where the eigenvalue spectra differ dramatically across parameter blocks (e.g., query, key, value, and MLP layers). This contrasts sharply with CNNs, which display block-homogeneous Hessians due to their repetitive convolutional structure \citep{zhang2024transformers}. The authors demonstrate that this block heterogeneity --- measurable before training begins --- strongly predicts whether SGD will succeed or fail. In controlled experiments, they construct simple MLPs with artificially induced block heterogeneity (by scaling layer outputs with different constants) and show that SGD fails even without attention mechanisms, while Adam remains robust. Crucially, they show that the largest eigenvalue ratios across blocks correlate with the Adam-SGD performance gap, suggesting that adaptive learning rates are essential for navigating the disparate curvature scales inherent to Transformer architectures.

\subsubsection{Critiques of Adaptive Gradient Optimizer Methods} 

\citep{wilson2017marginal} deliver a sharp critique on adaptive methods (including AdaGrad, RMSProp, and Adam): these optimizers may converge quickly on the training objective but often generalize worse than SGD or SGD+M. They construct toy separable classification tasks where Adam and RMSProp approach solutions with high test error (near random guessing), while SGD converges to a minimum-norm separator with zero test error. Moreover, they show empirically (on deep nets) that even when the adaptive methods achieve lower training loss, their validation/test error often plateaus earlier than that of SGD. Their findings challenge the ``less tuning, better convergence'' folklore of adaptive methods, showing that in practice adaptive methods often require similarly careful hyperparameter tuning as SGD. \citep{wilson2017marginal}. 

A key insight is that per-coordinate rescaling, while effective in accelerating convergence, induces a distinct \emph{algorithmic bias} in solution selection. Unlike SGD, which tends to converge to minimum-norm solutions in linearly separable settings, adaptive methods distort the underlying geometry of the optimization problem through their diagonal preconditioning. As a result, they systematically overweight coordinates with historically large gradients and underweight others, potentially leading to fits on spurious or less informative directions.

Formally, consider the separable linear classification setting where SGD converges to the minimum-norm separator under standard assumptions. \citet{wilson2017marginal}\ prove that adaptive methods such as AdaGrad and Adam do not share this implicit bias: the direction of the limiting classifier depends strongly on the history of gradient magnitudes rather than on the geometry of the data (see section 3.2 and Appendix C for a proper treatment). This mismatch can yield solutions with arbitrarily poor generalization, despite achieving low training error. Empirically, their work shows that adaptive optimizers often exhibit what can be described as \emph{plateauing generalization}: although training loss decreases quickly, test error levels off at values strictly worse than SGD, even when tuned carefully. This reflects the fact that rapid early progress is made along gradient-dominant directions, but the optimizer subsequently fails to fully explore flatter directions associated with improved generalization \citep{wilson2017marginal}.

Taken together, these results provide a theoretical and empirical counterpoint to the widespread adoption of Adam: while it excels in early convergence and stability, its long-term generalization behavior diverges significantly from SGD. This motivates later refinements such as hybrid schedules that attempt to combine Adam’s stability with the generalization properties of SGD.

\citet{keskar2017improving}\ empirically demonstrate that, while Adam typically converges quickly in training error, it often underperforms SGD in test accuracy due to an implicit generalization gap. They propose a hybrid strategy: start with Adam and then switch to SGD when the test error plateaus. Their experiments on CIFAR-10 with a DenseNet architecture reveal that pure SGD ultimately achieves lower test error ($\approx 5 \%$) compared to Adam’s plateau ($\approx 7\%$). By constraining Adam’s internal step size range via clipped momentum ratios (Adam-Clip variants), they partially close the gap, but the clearest improvement comes from switching to SGD midway --- achieving test error comparable to SGD but retaining Adam’s fast early convergence \citep{keskar2017improving}.

\begin{figure}
    \centering
    \includegraphics[width=0.5\linewidth]{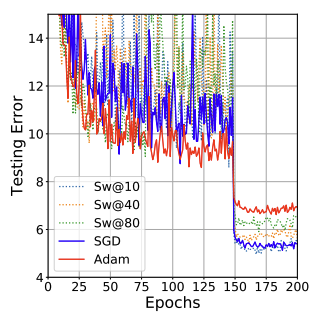}
    \caption{Figure 2 from \citep{keskar2017improving} showing the testing error on DenseNet architecture for CIFAR-10 with a varying switchover point (switching from Adam to SGD during training).}
    \label{fig:swats}
\end{figure}

The SWATS results underscore a critical point: Adam’s adaptivity may lead to overly conservative effective step sizes in later stages, limiting exploration of flat directions important for generalization. Thus, although Adam provides fast initial progress, its default trajectory can impede ultimate generalization performance, motivating mixed optimizer schedules, transitions to SGD for fine-tuning, or more advanced higher order optimizers.

\subsection{Limitations of Classical Methods and the Optimization-Learning Disconnect}
\begin{figure}
    \centering
    \includegraphics[width=0.5\linewidth]{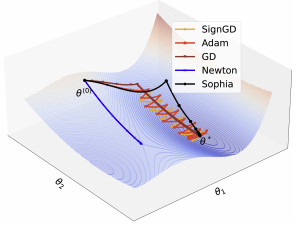}
    \caption{Figure 2 from \citep{liu2023sophia}. This toy descent graph highlights how adaptive methods may bounce around a single parameter dimension, while full second order methods (Newton) converge to a saddle point. Sophia is depicted as the winner in this scenario due to its ability to make reliable estimates of local curvature.}
    \label{fig:sophia_toy_ex}
\end{figure}

As touched on in the previous section, much of optimization theory focuses on minimizing a loss function as quickly as possible, e.g., proving convergence rates for convex functions. In deep learning, however, the true goal is \emph{generalization}: strong out-of-sample performance. In other words, an algorithm that optimizes the training loss most aggressively may not yield the best test accuracy. Second-order methods highlight this disconnect. In theory, Newton’s method provides dramatic acceleration and convergence to global minima. In practice, we see that in non-convex settings (such as loss functions for deep neural networks), Netwon's method can converge to a set of parameters that generalizes poorly and is sensitive to stochasticity of gradients (for a proper illustration, see Figure \ref{fig:sophia_toy_ex}).

We also notice that the computation costs associated with the classical methods that we have discussed so far:
\begin{itemize}
    \item \textbf{SGD}: requires one gradient computation per step, i.e.\ $\mathcal{O}(P)$ FLOPs for $P$ parameters.  
    \item \textbf{SGD+M}: essentially the same cost, $\mathcal{O}(P)$, with only a few additional vector operations.  
    \item \textbf{Newton’s}: computing the Hessian costs $\mathcal{O}(P^2)$ FLOPs, and inverting it costs $\mathcal{O}(P^3)$ FLOPs (via dense linear algebra). This is infeasible when $P \gg 10^6$.  
    \item \textbf{HF}: avoids forming the Hessian but requires iterative conjugate-gradient solves; each iteration needs repeated Hessian–vector products at cost $\mathcal{O}(P)$ each, multiplied by tens of inner iterations.  
    \item \textbf{L-BFGS}: stores and updates a low-rank approximation; the per-step cost is $\mathcal{O}(mP)$, where $m$ is the number of stored correction pairs (often $m \approx 10$). This is higher than SGD but vastly cheaper than full Newton's method.  
    \item \textbf{Adam(W)}: maintains first and second moment estimates, requiring $\mathcal{O}(P)$ memory for each. Per-step cost is $\mathcal{O}(P)$ with negligible overhead beyond SGD, making it the def facto method for many deep neural network training today.
\end{itemize}

These computational and theoretical limitations reveal a fundamental tension in deep learning optimization: methods with strong convergence guarantees prove impractical at scale, while scalable first-order methods lack formal understanding of their generalization properties. The field has thus converged on stochastic first-order methods with adaptive learning rates and ``momentum-ized'' algorithms that balance computational efficiency with empirical success, even as their theoretical foundations remain incomplete.

This gap between classical optimization theory and deep learning practice motivates the need for a new generation of algorithms. Rather than directly applying tools from pure convex optimization, modern approaches draw on insights from statistical learning theory, differential geometry, and empirical observations about neural network training dynamics. The following sections explore these advanced methods: algorithms that exploit problem structure through (approximate) curvature information, leverage techniques that are loosely coupled with statistical properties of overparameterized models, and incorporate inductive biases about feature learning --- all while maintaining computational tractability for billion-parameter networks.

\section{Curvature Matrices: Empirical Evidence and Theoretical Guarantees}\label{sec:curvmat}

The steepest descent framework of the previous section deliberately avoided second-order information at the risk of being computationally intractable. Instead, we settled for adaptive scaling methods with the wishful thinking that diagonal approximations to Hessian would be better than nothing. While this led to breakthroughs for training large models where the parameter space was extremely wide, we soon began to notice how brittle these solutions were and that there was a limit to informational gain by sticking to the diagonal.

However, we now ask: what if there was a cheap way to incorporate curvature information? This leads us to the classical idea of \emph{preconditioning} --- transforming the gradient using a matrix that approximates some notion of curvature. While conceptually moving toward second-order methods, modern preconditioned optimizers maintain computational efficiency by carefully approximating and factoring curvature matrices using various tricks from numerical linear algebra, information geometry, and other related fields.

The motivation for incorporating curvature is compelling: in high-dimensional non-convex optimization, different directions in parameter space exhibit vastly different curvatures. Gradient descent treats all directions equally, leading to slow progress in flat directions and instability in sharp ones. Adaptive descent only focuses on the higher order interactions between the same dimensions and ignores the cross terms (i.e. for a parameter vector $\theta$, how a change in $\theta_i$ affects the gradient of $\theta_j$). Preconditioning adapts the update to the local geometry, potentially accelerating convergence and improving generalization properties.


The classical preconditioned gradient descent update takes the form:
\[
\mathbf{w}_{t+1} = \mathbf{w}_t - \eta \, P_t^{-1} g_t
\]
where $P_t \in \mathbb{R}^{n \times n}$ is a \emph{preconditioner matrix} intended to approximate some notion of local curvature, and $\eta > 0$ is a step size. The preconditioner transforms the raw gradient $g_t$ into a direction $P_t^{-1} g_t$ that accounts for the geometry of the loss landscape. 

Luckily, we have already seen the first application of this idea straight from optimization theory: Newton's method is just ``preconditioned gradient descent'' with the Hessian of the loss function as the preconditioner! We also saw the online batch learning version of this setup earlier when exploring an improvement to SGD called Preconditioned Stochastic Gradient Descent \citep{li2017psgd}. We now turn to the generalization of this problem and explore curvature matrices that can help effectively train neural networks.

\subsection{The Zoo of Curvature Matrices}

In deep learning, several matrices can serve as preconditioners, each approximating different aspects of the loss landscape's curvature. We now survey the main candidates and their relationships.

\subsubsection{The Hessian Matrix}

The \emph{Hessian} of the loss function $\mathcal{L}(\mathbf{w})$ is the matrix of second partial derivatives (as discussed in the earlier sections):
\[
H(\mathbf{w}) = \nabla^2 \mathcal{L}(\mathbf{w}) = \frac{\partial^2 \mathcal{L}}{\partial w_i \partial w_j}
\]

The Hessian captures the local curvature exactly. We recall that Newton's method can be interpreted as preconditioning SGD with the Hessian  at the current iterate ($P_t = H(\mathbf{w}_t)$), yielding the update:
\[
\mathbf{w}_{t+1} = \mathbf{w}_t - \eta \, H(\mathbf{w}_t)^{-1} g_t
\]

This achieves quadratic convergence to minima in convex settings. However, computing and inverting the Hessian for a neural network with millions or billions of parameters is prohibitively expensive: $\mathcal{O}(n^2)$ memory and $\mathcal{O}(n^3)$ computation per step. Moreover, neural networks are inherently non-convex --- even composing two simple linear layers leads to cross-term interactions between variables. 

Beyond computational cost, the Hessian presents fundamental challenges in the non-convex regime typical of neural network training. The Hessian can be indefinite, possessing negative eigenvalues that cause $H^{-1}$ to point toward saddle points rather than minima. Additionally, the Hessian is sensitive to curvature in all parameter space directions, including those orthogonal to the gradient where the loss function is relatively flat. Stochastic mini-batch estimates of the Hessian are also extremely noisy, as they depend on second-order derivatives that exhibit high variance across different data samples. These pathologies motivate the development of approximations that retain useful curvature information while remaining computationally tractable and maintaining positive definiteness.

\subsubsection{The Generalized Gauss-Newton Matrix}

For a model with output $f(x; \mathbf{w}) \in \mathbb{R}^{d_y}$ and loss $\mathcal{L}(\mathbf{w}) = \mathbb{E}_{(x,y)} [\ell(f(x; \mathbf{w}), y)]$ decomposed into a per-example loss $\ell: \mathbb{R}^{d_y} \times \mathbb{R}^{d_y} \to \mathbb{R}$ and model prediction $f$, the \emph{generalized Gauss-Newton (GGN) matrix} is defined as:
\[
G(\mathbf{w}) = \mathbb{E}_{(x,y)} \left[ J_f(x; \mathbf{w})^\top H_\ell(f(x; \mathbf{w}), y) \, J_f(x; \mathbf{w}) \right]
\]
where $J_f(x; \mathbf{w}) = \frac{\partial f(x; \mathbf{w})}{\partial w} \in \mathbb{R}^{d_y \times n}$ is the Jacobian of the model output with respect to the $n$ parameters, and $H_\ell(f, y) = \frac{\partial^2 \ell(f, y)}{\partial f^2} \in \mathbb{R}^{d_y \times d_y}$ is the Hessian of the loss with respect to the model's output. 

\paragraph{Relationship to the Hessian.} Note the difference in the usage of ``Hessian'' here --- the Hessian with respect to the model parameters would be a matrix $\mathbb{R}^{n \times n}$, whereas the Hessian with respect to the model output is $\in \mathbb{R}^{d_y \times d_y}$. By the chain rule, the Hessian of the loss with respect to parameters decomposes as:
\[
H(\mathbf{w}) = \mathbb{E}_{(x,y)} \left[ J_f^\top H_\ell J_f + \sum_{i=1}^{d_y} \frac{\partial \ell}{\partial f_i} H_{f_i} \right]
\]
where $H_{f_i} = \frac{\partial^2 f_i}{\partial w^2}$ is the Hessian of the $i$-th output component with respect to parameters. The GGN matrix drops the second term, which involves second derivatives of the model itself. This approximation becomes exact in two important cases: when the loss is quadratic in the outputs (e.g., mean squared error, cross entropy), or when the model is close to a minimum where the loss gradients with respect to outputs satisfy $\partial \ell / \partial f_i \approx 0$ for all $i$ \citep{martens2020ngm}.

\paragraph{Advantages of the GGN.}
The GGN matrix is always positive semi-definite (PSD), which follows from its formulation as an expectation of products of the form $J_f^\top H_\ell J_f$ where $H_\ell$ is PSD for typical loss functions. This avoids the indefiniteness issues that plague the Hessian in non-convex settings. The GGN focuses on curvature that directly affects the model's outputs and, consequently, the loss value. It naturally ignores directions in parameter space that leave the function implemented by the network unchanged --- a form of curvature that is irrelevant for optimization. This ``loss-aware'' perspective makes the GGN more robust and better suited for preconditioning gradient descent in neural network training, but the actual GGN requires excessive computational costs. Rather, we will see that approximations to the GGN work well as actual preconditioners.

\subsubsection{The Fisher Information Matrix} \label{sec:fisher}

For a probabilistic model $p(y | x; \mathbf{w})$ with negative log-likelihood loss $\mathcal{L}(\mathbf{w}) = -\mathbb{E}_{x \sim p_{\text{data}}} [\log p(y | x; \mathbf{w})]$, the \emph{Fisher information matrix} (FIM) is defined as:
\[
F(\mathbf{w}) = \mathbb{E}_{x \sim p_{\text{data}}} \mathbb{E}_{y \sim p(y|x; \mathbf{w})} \left[ \nabla_w \log p(y|x; \mathbf{w}) \, \nabla_w \log p(y|x; \mathbf{w})^\top \right]
\]
The key distinguishing feature of the FIM is that the inner expectation is taken over the \emph{model distribution} $p(y|x; \mathbf{w})$, not the data distribution. This can equivalently be written as:
\[
F(\mathbf{w}) = \mathbb{E}_{x \sim p_{\text{data}}} \mathbb{E}_{y \sim p(y|x; \mathbf{w})} \left[ -\nabla_w^2 \log p(y|x; \mathbf{w}) \right]
\]

\paragraph{Connection to the GGN Matrix.}
For classification problems with cross-entropy loss or regression with Gaussian noise, the Fisher information matrix and the GGN matrix are mathematically equivalent \citep{martens2020ngm}. Specifically, both can be expressed as:
\[
F(\mathbf{w}) = G(\mathbf{w}) = \mathbb{E}_{x \sim p_{\text{data}}} \mathbb{E}_{y \sim p(y|x; \mathbf{w})} \left[ J_f(x; \mathbf{w})^\top H_\ell J_f(x; \mathbf{w}) \right]
\]
This equivalence holds because, for these loss functions, the Fisher information with respect to parameters can be decomposed through the model's Jacobian and the curvature of the log-likelihood with respect to outputs. Thus, in the context of neural network optimization with standard losses, the FIM and GGN matrix often refer to the same mathematical object, though they arise from different theoretical perspectives --- information geometry for the Fisher and loss function approximation for the GGN.

\paragraph{The Empirical Fisher.}
In practice, computing the true FIM requires sampling labels from the model distribution $p(y|x; \mathbf{w})$ for each input $x$, which adds computational overhead. Instead, practitioners commonly use the \emph{empirical FIM} matrix \citep{park2000ng}:
\[
\tilde{F}(\mathbf{w}) = \mathbb{E}_{(x,y) \sim p_{\text{data}}} \left[ \nabla_w \log p(y|x; \mathbf{w}) \, \nabla_w \log p(y|x; \mathbf{w})^\top \right]
\]
where the expectation is taken over the actual data distribution, using observed labels $y$ rather than samples from the model. The empirical FIM is computationally more efficient, as it can be computed alongside standard gradient calculations without additional forward-backward passes. In practice, given a dataset $\mathcal{D} = \{(x_i, y_i)\}_{i=1}^N$ (or, when doing batch approximations, this is defined as selecting subsets $\mathcal{D}_t$ from this dataset), the empirical FIM is computed as a sample average:
\[
\tilde{F}(\mathbf{w}) \approx \frac{1}{N} \sum_{i=1}^N \nabla_w \log p(y_i|x_i; \mathbf{w}) \, \nabla_w \log p(y_i|x_i; \mathbf{w})^\top
\]
This estimator directly follows from the definition of expectation as an average over the outer products of the gradients under the negative log-likelihood loss. These gradients will already be computed during standard backpropagation, so it's not significantly more expensive to compute. 

However, the empirical FIM and true Fisher can differ significantly, particularly early in training or when the model is poorly calibrated \citep{kunstner2019limitations}. The empirical FIM approaches the true FIM only under specific conditions: when the model distribution matches the data distribution (i.e., at a global optimum with perfect fit), or when using highly overparameterized models that can fit the training data exactly (as in deep learning). Despite these theoretical limitations, the empirical FIM has proven effective, both in practice and as motivation, for many second-order optimization methods.

\paragraph{Theoretical Foundations.}
Without going too deep into the complex mathematics underpinning the usage of this curvature matrix, we present brief theoretical foundations: first, the Fisher information matrix has deep connections to statistical learning and information geometry. In the framework of information geometry, the Fisher information defines the ``Riemannian metric'' on the statistical manifold of probability distributions parameterized by $w$ \citep{amari1998natural}. The Cramér-Rao bound establishes that the Fisher information controls the variance of any unbiased estimator, providing a fundamental limit on statistical efficiency. Natural gradient descent, which uses $F^{-1}$ to precondition gradients, is invariant to reparameterizations of the model, making the optimization trajectory independent of the specific parameter coordinates chosen. For generalized linear models and certain well-specified exponential family distributions, the Fisher information exactly captures the relevant curvature of the loss landscape, providing theoretical justification for its use in optimization.

\subsubsection{The AdaGrad Matrix} \label{sec:adagrad}

The Hessian, Fisher, and GGN matrix can be viewed as representing higher order information about the loss function under different metrics or assumptions. On a similar note, there's another curvature matrix whose approximated counterpart ended up being more famous and led to one of the most popular deep learning optimizers today: the Adam optimizer. Originally, the AdaGrad matrix operates on the principle of constructing a full matrix estimate of curvature through accumulated gradient outer products in an effort to preserve a notion of the second moment of the stochastic gradient \citep{duchi2011adaptive}.

\paragraph{Full Matrix Formulation.}
The core insight of AdaGrad is to build a running preconditioner based on the cumulative outer product matrix of gradients:
\[
G_t = \sum_{\tau=1}^t g_\tau g_\tau^\top
\]
where $g_t = \nabla f_t(x_t)$ is the gradient at time $t$. The theoretically motivated preconditioned update employs the full matrix:
\[
x_{t+1} = \Pi_{G_t^{1/2}} \left( x_t - \eta G_t^{-1/2} g_t \right)
\]
where $\Pi_A^X$ denotes projection onto the convex set $X$ under the Mahalanobis norm $\|\cdot\|_A = \sqrt{\langle \cdot, A \cdot \rangle}$. This update scales the gradient by the inverse square root of the cumulative outer product structure, effectively preconditioned by an estimate of the second moment of gradients --- a proxy for curvature.

However, computing $G_t^{1/2}$ and its inverse in high dimensions is computationally prohibitive, as it requires $\mathcal{O}(d^2)$ storage and $\mathcal{O}(d^3)$ computation per iteration. There have been methods such as the GGT algorithm proposed by Agarwal et. al that have tried to store and use the full AdaGrad matrix by devising a GPU-friendly way to apply the inverse square root operation to the low-rank version over a window of AdaGrad matrices (think using the last $r << n$ for $n$ total training steps) \citep{agarwal2020ggt}. While presenting impressive theoretical guarantees for non-convex stochastic optimization problems, these methods failed to scale as well and perform significantly better than a simple diagonal approximation of the AdaGrad matrix. 

\paragraph{Diagonal Approximation.}
Rather than attempting to estimate the full curvature matrix, the authors of the AdaGrad paper neatly found an estimate that would be far easier to compute; to make the algorithm scalable, (diagonal) AdaGrad restricts its computations to the diagonal elements of the full AdaGrad matrix:
\[
x_{t+1} = \Pi_{\text{diag}(G_t)^{1/2}} \left( x_t - \eta \text{diag}(G_t)^{-1/2} g_t \right)
\]
Equivalently, accumulating squared gradients element-wise:
\[
v_t = v_{t-1} + g_t \odot g_t
\]
and applying the per-coordinate update:
\[
x_{t+1} = x_t - \frac{\eta}{\sqrt{v_t + \epsilon}} \odot g_t
\]
where $\odot$ denotes the Hadamard product (element-wise multiplication), division is also element-wise, and $\epsilon > 0$ is a small regularization constant for numerical stability.

This diagonal simplification reduces both storage to $\mathcal{O}(d)$ and per-iteration computation to $\mathcal{O}(d)$ (and even less when gradients are sparse), making the method practical for high-dimensional problems.

\begin{figure}
    \centering
    \includegraphics[width=0.75\linewidth]{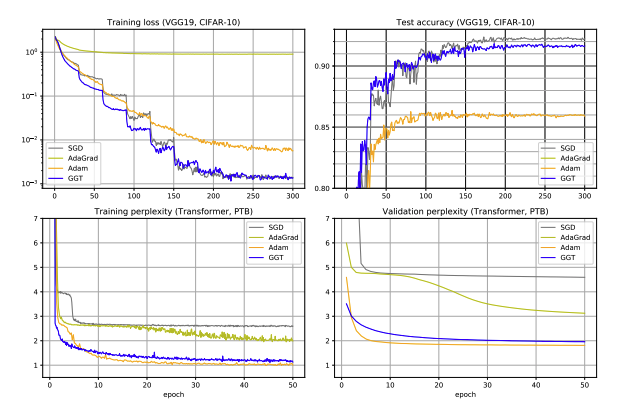}
    \caption{Figure 6 from \citep{agarwal2020ggt}. Plots comparing adaptive training algorithms on CIFAR-10 and PTB language-level modeling. AdaGrad seems unstable, while GGT doesn't outperform SGD or Adam significantly on test accuracy or validation perplexity.}
    \label{fig:ggt_adam}
\end{figure}

\paragraph{Connection to Curvature Estimation.}
The relationship between the full and diagonal variants illuminates how AdaGrad approximates curvature. For a loss function $\mathcal{L}(\mathbf{w}) = \mathbb{E}_{(x,y)} [\ell(f(x; \mathbf{w}), y)]$, the gradient outer product matrix captures second-order information:
\[
G(\mathbf{w}) = \mathbb{E}_{(x,y)} \left[ g(x,y,\mathbf{w}) \, g(x,y,\mathbf{w})^\top \right]
\]
where $g(x,y,\mathbf{w}) = \nabla_w \ell(f(x; \mathbf{w}), y)$ is the per-example gradient. AdaGrad's full-matrix variant builds an online estimate of this expectation through accumulated outer products $G_t = \sum_{\tau=1}^t g_\tau g_\tau^\top$, which under stationarity approximates $G(\mathbf{w})$. In a sense, the full AdaGrad matrix can be seen as tracking a smoothened version of the empirical FIM as both use the actual gradient outer products to form their curvature estimation.

By taking only the diagonal, AdaGrad obtains $v_t \approx \text{diag}(G_t)$, yielding a computationally efficient approximation where each coordinate $i$ satisfies $v_{t,i} \approx \sum_{\tau=1}^t g_{\tau,i}^2$. This diagonal element estimates the local coordinate-wise curvature (the variance of that parameter's gradient component) via the running accumulation of squared gradients. The inverse square root scaling, $v_t^{-1/2}$, corresponds to using the inverse square root of the curvature estimate, analogous to Newton-type methods that employ $H^{-1}$ (or approximations thereof) as the preconditioner.

The key trade-off is clear: the full matrix variant $G_t^{1/2}$ captures off-diagonal structure and parameter correlations, but becomes intractable at scale. The diagonal variant $\text{diag}(G_t)^{1/2}$ discards correlations but remains computationally efficient and often performs remarkably well in practice. Interestingly, theoretical regret bounds for the diagonal variant are only slightly weaker than the full matrix bounds, suggesting the lost correlation information is often not critical for convergence in many settings \citep{duchi2011adaptive}.

\paragraph{Adam as Smoothed AdaGrad.}
Adam extends AdaGrad by replacing the cumulative sum with exponential moving averages to mitigate issues arising from monotonic decay of learning rates in non-convex settings \citep{kingma2015adam}:
\[
m_t = \beta_1 m_{t-1} + (1 - \beta_1) g_t
\]
\[
v_t = \beta_2 v_{t-1} + (1 - \beta_2) (g_t \odot g_t)
\]
\[
x_{t+1} = x_t - \eta \frac{m_t}{\sqrt{v_t} + \epsilon}
\]
where typically $\beta_1 = 0.9$ and $\beta_2 = 0.999$. Through the curvature approximation lens, Adam can be understood as estimating a diagonal Hessian or Fisher matrix with temporal smoothing, effectively replacing the infinite-horizon accumulation of AdaGrad with a recent exponential window. This provides implicit warmup and noise reduction, preventing the aggressive learning rate decay that can occur in AdaGrad when early gradients are large. However, Adam forgoes some of AdaGrad's theoretical guarantees; while Adam admits alternative interpretations as approximate sign descent or $\ell_\infty$-normalized gradient descent (which we will discuss in later sections), it can diverge even on convex objectives --- a phenomenon not observed with AdaGrad, highlighting the depth of the curvature approximation perspective \citep{wilson2017marginal}.

\subsubsection{Alternate Stochastic Diagonal Hessian Approximations.}

While adaptive methods such as Adam focus on diagonal approximations through accumulated squared gradients, this estimation is inherently noisy due to its stochastic nature and sensitive to batch sizes. Moreover, the fundamental challenges that led to the abandonment of second-order methods in early neural network training remain relevant: computing stochastic approximations to the full Hessian requires prohibitive $\mathcal{O}(n^2)$ memory for storage and $\mathcal{O}(n^3)$ computation for inversion, where $n$ can reach billions in modern models. Additionally, the Hessian is generally indefinite in non-convex settings, meaning Newton search directions are not guaranteed to be descent directions. These issues motivated researchers to seek positive semi-definite approximations like the GGN matrix, or to restrict attention to diagonal elements that can be computed and inverted efficiently.

Sophia represents a recent attempt to leverage diagonal Hessian information while addressing these computational and stability concerns through a combination of lightweight stochastic Hessian estimation and element-wise clipping \citep{liu2023sophia}. Sophia maintains exponential moving averages of both gradients and diagonal Hessian estimates:
\[
m_t = \beta_1 m_{t-1} + (1-\beta_1) g_t
\]
\[
h_t = \beta_2 h_{t-1} + (1-\beta_2) \hat{h}_t
\]
where $\hat{h}_t$ is a stochastic estimate of the diagonal Hessian $\text{diag}(\nabla^2 \mathcal{L}(\mathbf{w}_t))$, computed infrequently (e.g., every 10 steps). The update is then:
\[
\mathbf{w}_{t+1} = \mathbf{w}_t - \eta \cdot \text{clip}\left(\frac{m_t}{\max(h_t, \epsilon)}, \rho\right)
\]
where $\text{clip}(\cdot, \rho)$ applies element-wise clipping: each coordinate is clipped to the range $[-\rho, \rho]$. This clipping mechanism serves multiple purposes: it controls worst-case update magnitude, guards against negative Hessian estimates (which could potentially yield ``ascent'' directions), and provides robustness to rapid Hessian changes along the trajectory.

Sophia proposes two estimators, both with computational cost comparable to a single gradient evaluation:
\begin{enumerate}
    \item \textbf{Hutchinson's estimator}: Draws $u \sim \mathcal{N}(0, I)$ and computes $\hat{h} = u \odot (\nabla^2 \mathcal{L}(\mathbf{w}) u)$ where $\nabla^2 \mathcal{L}(\mathbf{w}) u$ is a Hessian-vector product. This gives an unbiased estimate since $\mathbb{E}[u_i (\nabla^2 \mathcal{L}(\mathbf{w}) u)_i] = (\nabla^2 \mathcal{L}(\mathbf{w}))_{ii}$ by spherical symmetry.
    
    \item \textbf{Gauss-Newton-Bartlett (GNB) estimator}: For cross-entropy loss $\mathcal{L}(\mathbf{w}) = -\log p(y|x; \mathbf{w})$, leverages the Gauss-Newton decomposition and Bartlett identities to construct $\hat{h}$ by resampling labels from the model distribution. This is biased (approximates only the GGN term) but guarantees positive semi-definiteness and has lower variance in practice.
\end{enumerate}
Crucially, Sophia computes these estimates infrequently and uses only a small subset of the mini-batch (e.g., 32-240 examples), introducing less than 5\% average per-step overhead.

\citet{liu2023sophia} motivate Sophia by observing that deep learning loss landscapes exhibit heterogeneous curvature --- different parameter dimensions have vastly different local curvatures. AdaGrad and Adam's diagonal preconditioners provide only mild adaptation: they scale learning rates based on gradient magnitudes (approximately $1/\sqrt{\sum g^2}$), not curvature. In contrast, Sophia's diagonal Hessian preconditioner $1/h$ provides stronger differentiation between sharp dimensions (large $h$, small update) and flat dimensions (small $h$, large update), ensuring more uniform loss decrease across all parameters \citep{liu2023sophia}.

Diagonal preconditioning methods fundamentally ignore all off-diagonal structure in the curvature matrix, discarding information about correlations and interactions between different parameters. For a linear layer with weight matrix $W \in \mathbb{R}^{d_{\text{out}} \times d_{\text{in}}}$ flattened into a parameter vector of dimension $n = d_{\text{out}} \cdot d_{\text{in}}$, the Fisher information matrix exhibits rich block structure arising from the Kronecker product form of gradient covariances. Diagonal methods cannot capture this structure, which encodes how changes to different parameters jointly affect the loss. This limitation motivates structured approximations like KFAC and Shampoo, which we discuss in the next section, that maintain tractable representations of off-diagonal curvature information through factored forms.

\subsection{Why Approximate Curvature?}
The choice to approximate curvature matrices balances several competing objectives. Full Hessian computation and inversion scale as $\mathcal{O}(n^2)$ memory and $\mathcal{O}(n^3)$ time per iteration, where $n$ can reach billions for modern models. Even storing the Hessian becomes infeasible at scale. Structured approximations can reduce this computational burden by exploiting the layered structure of neural networks, making second-order methods tractable for deep learning.

A fundamental challenge beyond computational cost is the conditioning of curvature matrices. In neural networks, the Hessian typically exhibits condition numbers on the order of $10^9$ or higher, meaning the ratio of largest to smallest eigenvalues spans many orders of magnitude \citep{chaudhari2017entropy}. From classical optimization theory, the convergence rate of gradient descent deteriorates as $c = \left(\frac{\kappa-1}{\kappa+1}\right)^2$ where $\kappa = L/m$ is the condition number. With $\kappa \approx 10^9$ for some neural networks, this factor approaches 1, yielding glacially slow convergence. Even worse, numerical inversion of such ill-conditioned matrices amplifies noise and can produce unstable or divergent update directions \citep{zimmermann2015condition}. This extreme ill-conditioning fundamentally limits the reliability of exact second-order methods in neural network training, motivating approximations that implicitly regularize or dampen the curvature spectrum.

Beyond raw computational cost, preconditioned methods can exhibit better parallelization properties on modern accelerators. Computing second-order statistics can be distributed across devices or overlapped with forward and backward passes. On hardware where memory bandwidth and communication constitute primary bottlenecks, the reduction in iteration count afforded by better curvature information can offset the increased per-iteration cost.

In stochastic settings with mini-batches, exact Hessian estimates suffer from extreme noise. Approximations that aggregate information over many samples, such as exponential moving averages of covariances, provide substantially more stable curvature estimates. The bias introduced by approximation is often outweighed by the reduction in variance, leading to more reliable optimization.

The choice of approximation also affects numerical stability. The GGN and FIM are guaranteed to be positive semi-definite, unlike the Hessian which can have negative eigenvalues arising from non-convex regions. Ensuring positive semi-definite preconditioners avoids instabilities arising from negative curvature directions. Diagonal and Kronecker approximations naturally preserve this property when constructed from gradient outer/element-wise products, while also implicitly dampening extreme eigenvalues that plague exact curvature matrices.

Finally, approximations introduce inductive biases that can act as implicit regularization. For instance, the independence assumption underlying a Kronecker factorization, which we cover in great detail in Section \ref{sec:kfac}, favors solutions where layer inputs and outputs are decorrelated. Diagonal methods such as Adam or Sophia impose scaling preferences for the diagonal terms of the matrix and can lead to increasing gradient norms \citep{xie2023overlooked}. While these biases represent departures from the true curvature, they could also improve generalization by encoding structural priors about neural network optimization landscapes and by mitigating the pathological conditioning inherent to neural network Hessians. The practical success of these methods suggests that faithful curvature approximation may be less important than stable, well-conditioned preconditioners that provide consistent descent directions under stochastic gradients.

\subsection{Kronecker-Factored Approximate Curvature (KFAC)} \label{sec:kfac}

KFAC exploits the layered structure of neural networks to approximate the Fisher information matrix using Kronecker products \citep{martens2015kfac}. For each layer, rather than storing a full $n \times n$ matrix where $n$ is the total number of parameters, KFAC factors the curvature into smaller matrices whose Kronecker product approximates the full layer curvature. This factorization dramatically reduces both storage and computational requirements while maintaining much of the second-order information needed for effective optimization.

\subsubsection{The Block Structure of the Fisher Matrix}

For a neural network with parameters $\theta = [vec(W_1)^\top \; vec(W_2)^\top \; \cdots \; vec(W_\ell)^\top]^\top$, the Fisher information matrix can be viewed as an $\ell \times \ell$ block matrix:
\[
F = \begin{bmatrix}
F_{1,1} & F_{1,2} & \cdots & F_{1,\ell} \\
F_{2,1} & F_{2,2} & \cdots & F_{2,\ell} \\
\vdots & \vdots & \ddots & \vdots \\
F_{\ell,1} & F_{\ell,2} & \cdots & F_{\ell,\ell}
\end{bmatrix}
\]
where the $(i,j)$-th block is given by $F_{i,j} = \mathbb{E}[vec(DW_i) \, vec(DW_j)^\top]$ and the expectation is taken over the data distribution for inputs and the model's predictive distribution for outputs.

For a linear layer with weight matrix $W_i \in \mathbb{R}^{d_{\text{out}} \times d_{\text{in}}}$, the gradient with respect to $W_i$ computed during backpropagation is:
\[
DW_i = g_i \bar{a}_{i-1}^\top
\]
where $g_i = \frac{\partial \mathcal{L}}{\partial s_i} \in \mathbb{R}^{d_{\text{out}}}$ is the gradient with respect to the layer's pre-activations $s_i = W_i\bar{a}_{i-1}$, and $\bar{a}_{i-1} \in \mathbb{R}^{d_{\text{in}}+1}$ denotes the ``activities'' (post-activation outputs) from the previous layer. The bar is included to note that we append a homogeneous coordinate (value 1) to our activities $a_{i-1}$ to handle biases implicitly. Using the vectorization identity $vec(uv^\top) = v \otimes u$, we have:
\[
vec(DW_i) = \bar{a}_{i-1} \otimes g_i
\]
where $\otimes$ denotes the Kronecker product. For a more detailed overview on the theoretical setup for the Kronecker-factored approximation, please see Section 2.1 of the original KFAC paper: the notation should be consistent between the works \citep{martens2015kfac}.

\subsubsection{The Kronecker-Factored Approximation}

Now, we focus on the FIM for a single layer. Substituting the Kronecker product expression into the block formula yields:
\[
F_{i,j} = \mathbb{E}[(\bar{a}_{i-1} \otimes g_i)(\bar{a}_{j-1} \otimes g_j)^\top] = \mathbb{E}[\bar{a}_{i-1}\bar{a}_{j-1}^\top \otimes g_ig_j^\top]
\]

The KFAC approximation assumes that products of unit activities are statistically independent from products of unit input derivatives. More precisely, KFAC approximates:
\[
\mathbb{E}[\bar{a}_{i-1}\bar{a}_{j-1}^\top \otimes g_ig_j^\top] \approx \mathbb{E}[\bar{a}_{i-1}\bar{a}_{j-1}^\top] \otimes \mathbb{E}[g_ig_j^\top]
\]
\[
\mathbb{E}[A \otimes S] \approx \mathbb{E}[\bar{a}_{i-1}\bar{a}_{j-1}^\top] \otimes \mathbb{E}[g_ig_j^\top]
\]
This gives the block-wise approximation:
\[
F_{i,j} \approx A_{i-1,j-1} \otimes S_{i,j} = \tilde{F}_{i,j}
\]
where $A_{i-1,j-1} = \mathbb{E}[\bar{a}_{i-1}\bar{a}_{j-1}^\top] \in \mathbb{R}^{(d_{\text{in}}+1) \times (d_{\text{in}}+1)}$ is the covariance of layer activities and $S_{i,j} = \mathbb{E}[g_ig_j^\top] \in \mathbb{R}^{d_{\text{out}} \times d_{\text{out}}}$ is the covariance of backpropagated gradients.

It is important to note that this approximation does not assume independence between individual activities and gradients (which would be false, as activities directly influence gradients through the forward pass). Rather, it assumes that products of activities from potentially different layers are independent from products of gradients, which is a much weaker and more plausible assumption. Martens and Grosse provide a cumulant-based interpretation showing that the approximation error is small when higher-order cumulants of the joint distribution over these quantities are small, which tends to occur when this distribution is approximately Gaussian \citep{martens2015kfac}.

\paragraph{Further Approximation: Block-Diagonal Structure.}
Even with the Kronecker factorization, the full approximate Fisher $\tilde{F}$ remains expensive to invert. KFAC makes a further approximation by taking only the diagonal blocks, giving:
\[
\hat{F} = \text{diag}(\tilde{F}_{1,1}, \tilde{F}_{2,2}, \ldots, \tilde{F}_{\ell,\ell}) = \text{diag}(A_{0,0} \otimes S_{1,1}, A_{1,1} \otimes S_{2,2}, \ldots, A_{\ell-1,\ell-1} \otimes S_{\ell,\ell})
\]
This block-diagonal approximation is motivated by the observation that when predicting the derivative of weights in layer $i$ using a linear model, the derivatives of weights in the same layer are far more informative than those from distant layers. Martens and Grosse show empirically that the inverse Fisher exhibits strong block-diagonal (or block-tridiagonal) structure, even when the Fisher itself does not.

\paragraph{Efficient Inversion and Updates.}
The Kronecker product satisfies $(A \otimes B)^{-1} = A^{-1} \otimes B^{-1}$, allowing efficient computation of the inverse:
\[
\hat{F}^{-1} = \text{diag}(A_{0,0}^{-1} \otimes S_{1,1}^{-1}, A_{1,1}^{-1} \otimes S_{2,2}^{-1}, \ldots, A_{\ell-1,\ell-1}^{-1} \otimes S_{\ell,\ell}^{-1})
\]
Thus, inverting $\hat{F}$ reduces to inverting $2\ell$ much smaller matrices. For layer $i$ with weight matrix $W_i \in \mathbb{R}^{d_{\text{out}} \times d_{\text{in}}}$, storing $A_{i-1,i-1}$ and $S_{i,i}$ requires $\mathcal{O}(d_{\text{in}}^2 + d_{\text{out}}^2)$ memory instead of $\mathcal{O}((d_{\text{in}} d_{\text{out}})^2)$ for the full block. Computing their inverses requires $\mathcal{O}(d_{\text{in}}^3 + d_{\text{out}}^3)$ operations instead of $\mathcal{O}((d_{\text{in}} d_{\text{out}})^3)$. For typical layer dimensions where $d_{\text{in}} + d_{\text{out}} \ll d_{\text{in}} \cdot d_{\text{out}}$, this represents a massive computational savings.

Using the identity $(A \otimes B)\text{vec}(X) = \text{vec}(BXA^\top)$, the preconditioned gradient for layer $i$ can be computed as:
\[
\tilde{G}_i = S_i^{-1} G_i A_i^{-1}
\]
where $G_i$ is the gradient matrix for layer $i$. The weight update becomes:
\[
W_{i,t+1} = W_{i,t} - \eta \, S_{i,t}^{-1} G_{i,t} A_{i,t}^{-1}
\]

\subsubsection{Practical Considerations}

In practice, for a layer $l$, the matrices $A_l$ and $S_l$ are estimated using exponential moving averages of empirical covariances computed over mini-batches. For the activation covariances $A_l$, expectations are taken over the data distribution using forward pass activations. For the gradient covariances $S_l$, expectations are taken over the model's predictive distribution by sampling targets from the model and performing additional backward passes with these sampled targets.

The KFAC approximation described here represents the foundation of the method. Martens and Grosse further develop a block-tridiagonal approximation to the inverse Fisher which captures some off-diagonal structure while remaining efficiently invertible. Additionally, they emphasize that damping techniques --- particularly a combination of factored Tikhonov regularization and adaptive rescaling based on the exact Fisher --- are absolutely essential for KFAC to work well in practice. Without proper damping, the approximate quadratic model can become untrustworthy, leading to poor update proposals. These damping techniques compensate both for the local quadratic approximation of the objective and for the Fisher approximation itself.

\subsection{Empirical Evidence and Theoretical Backing for KFAC}

While KFAC and related Kronecker-factored methods have traditionally been motivated from an approximation-theoretic perspective --- aiming to efficiently approximate ideal curvature matrices like the Hessian or Fisher information --- recent work by \citet{zhang2025concurrence} provides a fundamentally different justification . Rather than viewing layer-wise preconditioning as a computational compromise, they demonstrate that these methods are ``provable necessary'' from a statistical learning perspective, emerging as the natural ``correct'' solution to feature learning problems under realistic (anisotropic) data distributions.

\subsubsection{The Statistical Necessity of Layer-wise Preconditioning}

Classical analyses of gradient-based optimization in neural networks often assume idealized conditions: isotropic input distributions $\mathbf{w} \sim \mathcal{N}(\mathbf{0}, \mathbf{I})$ and well-conditioned problem instances. \citet{zhang2025concurrence} demonstrate that when we relax these assumptions to more realistic settings --- anisotropic covariates $\mathbf{w} \sim \mathcal{N}(\mathbf{0}, \boldsymbol{\Sigma}_x)$ with $\boldsymbol{\Sigma}_x$ noisy --- standard SGD becomes a suboptimal feature learner. The authors study two prototypical problems for understanding feature learning in neural networks:

\paragraph{1. Linear Representation Learning.}
In this setting, a two-layer network learns to extract useful features from input data to solve a downstream task. The key finding is that under anisotropic input distributions, SGD's convergence rate degrades dramatically, exhibiting undesirable dependence on the conditioning of the input covariance matrix $\boldsymbol{\Sigma}_x$. Specifically, even with ideally tuned step sizes, SGD's ability to learn relevant features is bottlenecked by the eigenspectrum of $\boldsymbol{\Sigma}_x$.

\paragraph{2. Single-Index Learning.}
This problem involves learning a function of the form $f(\mathbf{w}) = \sigma(\mathbf{w}^* {}^\top \mathbf{w})$, where the goal is to recover the direction $\mathbf{w}^*$. Again, under mild anisotropy in the input distribution, SGD exhibits slow convergence and fails to efficiently align with the target direction when $\mathbf{w}^*$ is misaligned with the principal components of $\boldsymbol{\Sigma}_x$.

\subsubsection{KFAC as the Natural Solution}

The central theoretical contribution of \citet{zhang2025concurrence} is showing that Kronecker-factored preconditioning emerges as the \emph{first-principles solution} to the feature learning failures of SGD. Consider the linear representation learning problem where data is generated as $\mathbf{y}_i = \mathbf{F}_\star \mathbf{G}_\star \mathbf{w}_i + \boldsymbol{\varepsilon}_i$, with $\mathbf{w}_i \sim N(\mathbf{0}, \boldsymbol{\Sigma}_x)$ and the goal is to learn the low-dimensional feature space that $\mathbf{G}_\star$ maps to. \citet{zhang2025concurrence} demonstrate that standard SGD fails catastrophically when the input covariance $\boldsymbol{\Sigma}_x$ is poorly conditioned, converging only along the dominant eigendirections of $\boldsymbol{\Sigma}_x$ rather than learning the true feature space defined by $\mathbf{G}_\star$.

The authors reference a paper that previously derived an alternating descent algorithm that addresses this failure through bidirectional preconditioning \citep{zhang2024dfw}. For a two-layer predictor $f_{\mathbf{F},\mathbf{G}}(\mathbf{w}) = \mathbf{F}\sigma(\mathbf{G}\mathbf{w})$ trained against the MSE (mean squared error) regression objective, the De-bias Feature \& Whiten (DFW) update takes the form:
\[
\mathbf{G}_{+} = \mathbf{G} - \eta_G \mathbf{P}_\mathbf{G}^{-1} \nabla_\mathbf{G} \widehat{\mathcal{L}}(\mathbf{F}, \mathbf{G}) (\mathbf{Q}_\mathbf{G} + \lambda_G \mathbf{I}_{d_x})^{-1}
\]
\[
\mathbf{F}_{+} = \mathbf{F} - \eta_F \mathbf{P}_\mathbf{F}^{-1} \nabla_\mathbf{F} \widehat{\mathcal{L}}(\mathbf{F}, \mathbf{G}_{+}) \mathbf{Q}_\mathbf{F}^{-1}
\]
where the preconditioners are defined as:
\[
\mathbf{Q}_\mathbf{G} = \widehat{\boldsymbol{\Sigma}}_x = \widehat{\mathbb{E}}[\mathbf{w}\mathbf{w}^\top], \quad \mathbf{Q}_\mathbf{F} = \widehat{\boldsymbol{\Sigma}}_\mathbf{z} = \widehat{\mathbb{E}}[\mathbf{z}\mathbf{z}^\top]
\]
\[
\mathbf{P}_\mathbf{G} = \widehat{\mathbb{E}}\left[\left(\frac{\partial f_{\mathbf{F},\mathbf{G}}}{\partial \mathbf{h}}\right)^\top \frac{\partial f_{\mathbf{F},\mathbf{G}}}{\partial \mathbf{h}}\right] \text{ (or } \mathbf{I}_{d_h}\text{)}, \quad \mathbf{P}_\mathbf{F} = \mathbf{I}_{d_y}
\]
Here $\mathbf{h} = \mathbf{G}\mathbf{w}$ denotes the hidden layer pre-activations and $\mathbf{z} = \sigma(\mathbf{G}\mathbf{w})$ denotes the post-activations.

This update structure reveals a striking correspondence: when $\mathbf{P}_\mathbf{G} = \mathbf{I}$ and viewing the loss through the lens of the conditionally Gaussian model $p(\mathbf{y}|\mathbf{w}) = N(f_{\mathbf{F},\mathbf{G}}(\mathbf{w}), \sigma^2\mathbf{I})$, the DFW preconditioners align exactly with KFAC \citep{zhang2025concurrence}. The matrix form of the updates directly parallels KFAC's layer-wise preconditioning:
\[
(\mathbf{Q}_\ell \otimes \mathbf{P}_\ell)^{-1} \nabla_{\mathbf{w}_\ell} \mathcal{L}(\mathbf{w}) \iff \mathbf{P}_\ell^{-1} \nabla_{\mathbf{w}_\ell}\mathcal{L}(\mathbf{w}) \mathbf{Q}_\ell^{-1}
\]
This equivalence demonstrates that KFAC is not merely an approximation to the Fisher for computational convenience --- it is the geometrically natural preconditioner that whitens both the input space (via $\mathbf{Q}_\ell^{-1}$) and the gradient backpropagation space (via $\mathbf{P}_\ell^{-1}$), ensuring that learning is not bottlenecked by the condition numbers of either space.

\citet{zhang2025concurrence} prove various guarantees that demonstrate layer-wise preconditioned methods achieve convergence rates that do not degrade with the condition number of $\boldsymbol{\Sigma}_x$, unlike standard SGD. Moreover, these methods learn representations aligned with the optimal features for the task rather than merely capturing the dominant modes of the input distribution. Under appropriate regularity conditions, layer-wise preconditioning achieves statistically optimal rates for feature learning. Crucially, these guarantees are \emph{non-approximation-theoretic} --- they do not rely on how well KFAC approximates some idealized second-order method, but rather on KFAC's intrinsic properties as a feature learner.

\subsubsection{Empirical Verification of KFAC}

Through programming the update rules and setting up experiments, the authors discovered that KFAC outperforms the incumbent optimization techniques on these featured problems. In fact, a striking finding in \citet{zhang2025concurrence} is that KFAC not only matches but often exceeds the performance of the full second-order methods it was designed to approximate. This observation, first noted in the literature by Benzing, challenges the conventional approximation-theoretic narrative \citep{benzing2022imposing}. In carefully controlled experiments on linear representation learning and single-index models, the authors demonstrate that KFAC converges faster than Newton's Method and Natural Gradient Descent in terms of both iteration count and wall-clock time. Full second-order methods exhibit instability and divergence in settings where KFAC remains stable, particularly when the curvature estimates are noisy or the problem is poorly conditioned. Furthermore, KFAC often achieves lower test error than full second-order methods, suggesting it provides beneficial implicit regularization.

\begin{figure*}[t]
\centering
\begin{minipage}{0.32\textwidth}
    \centering
    \includegraphics[width=\linewidth]{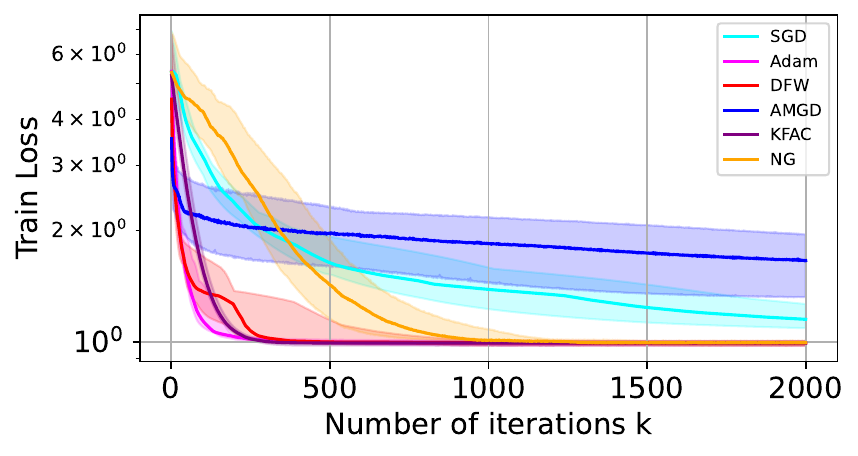}
\end{minipage}%
\begin{minipage}{0.32\textwidth}
    \centering
    \includegraphics[width=\linewidth]{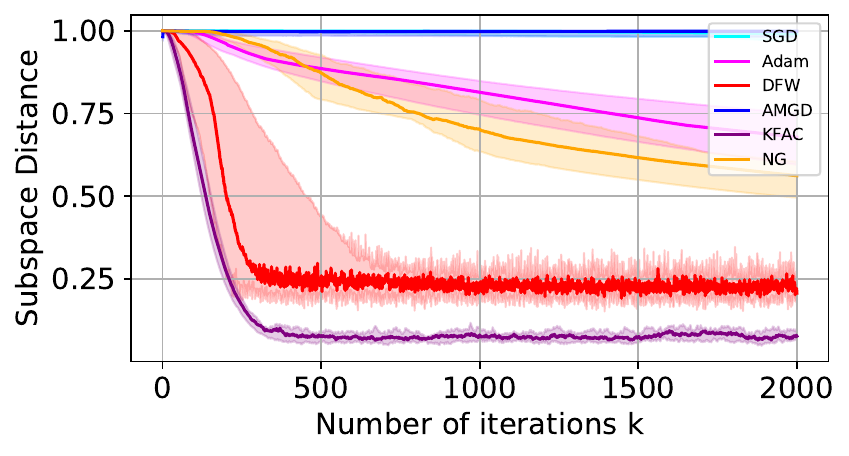}
\end{minipage}%
\begin{minipage}{0.32\textwidth}
    \centering
    \includegraphics[width=\linewidth]{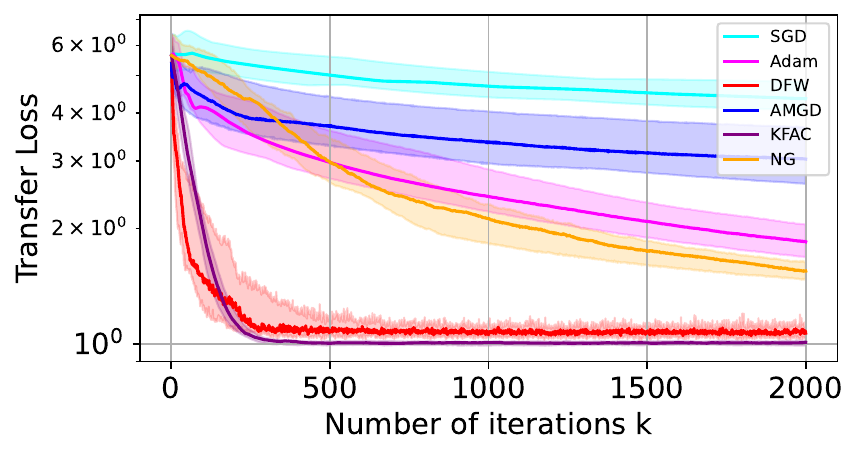}
\end{minipage}%
\caption{
Figure from \citep{zhang2025concurrence}: The training loss, subspace distance, and transfer loss by various algorithms on the linear representation learning task. KFAC not only outperforms SGD and Adam but even its inspiration, the natural gradient method.
}
\label{fig:headtohead}
\end{figure*}

This paradox --- that an ``approximation'' outperforms the ``ideal'' --- is resolved by recognizing that KFAC is not merely a computational shortcut, but rather implements a different (and potentially superior) optimization geometry tailored to feature learning. \citet{zhang2025concurrence}\ carefully verify that the benefits of layer-wise preconditioning cannot be replicated by existing deep learning algorithms and techniques. The authors show that Adam-like diagonal preconditioning provides only \emph{mild mitigation} of the feature learning issues identified for SGD. While Adam addresses per-coordinate scale imbalances, it cannot capture the critical off-diagonal structure --- the correlations between input dimensions and between neurons --- that Kronecker factorization explicitly models  \citep{zhang2025concurrence}. In the paper's studies, Adam converges faster than SGD but still exhibits suboptimal feature learning under anisotropic inputs, with the gap between Adam and KFAC widening as the conditioning of $\boldsymbol{\Sigma}_x$ worsens (see Figure \ref{fig:headtohead}). Empirically, they show that Adam fails to achieve the conditioning-independent guarantees that KFAC provides theoretically.

Batch normalization (BatchNorm) is sometimes viewed as an implicit preconditioner, as it whitens activations layer-by-layer \citep{ioffe2015batch}. However, \citet{zhang2025concurrence}\ demonstrate that BatchNorm does \emph{not} fix the fundamental feature learning issues in their models and can even \emph{hurt generalization} by introducing batch-dependent stochasticity that interferes with learning. These negative results for BatchNorm are particularly noteworthy, as they suggest that not all forms of normalization provide the same benefits as explicit layer-wise preconditioning of the gradient.

\subsubsection{The Deep Learning Regime: Why Curvature Matters at Scale}

The theoretical analysis in \citet{zhang2025concurrence} uses simplified two-layer models, but the authors argue that their findings extend to the deep learning regime through several mechanisms. Modern deep networks exhibit hierarchical feature learning, where each layer learns increasingly abstract representations. The anisotropy and conditioning issues identified in the two-layer case compound across layers: poor feature learning in early layers cascades to later layers, degrading overall performance.

Real-world data distributions are invariably anisotropic. Natural images, text embeddings, and sensor data all exhibit strong correlations, measurement noise, and imbalanced spectra. The assumption $\mathbf{w} \sim \mathcal{N}(\mu, \mathbf{I})$ is a mathematical convenience that rarely holds in practice, making KFAC's robustness to anisotropy critical for practical deep learning. While the theoretical results focus on simplified models, the paper's empirical validation extends to actual neural network training, confirming that the qualitative behaviors predicted by theory --- KFAC's superior feature learning, Adam's partial mitigation, and BatchNorm's limitations --- manifest in real deep learning tasks.

The work of \citet{zhang2025concurrence} provides a paradigm shift in how we understand layer-wise preconditioning methods like KFAC. Rather than viewing KFAC as merely a cheap substitute for an idealized second-order method, we now understand it as a statistically optimal solution to feature learning under realistic data distributions. Neither diagonal methods like Adam nor normalization schemes like BatchNorm can replicate KFAC's performance, as they miss essential structural information captured by the Kronecker factorization. Layer-wise preconditioning achieves conditioning-independent, feature-aligned convergence rates that are statistically optimal for the feature learning problems studied. Remarkably, KFAC outperforms not only simpler methods like Adam and SGD, but even full second-order methods like Newton's Method and Natural Gradient Descent, demonstrating that the right geometric structure matters more than crude approximation fidelity.

This theoretical backing, combined with the empirical success of curvature methods in competitions such as AlgoPerf, suggests that layer-wise preconditioning represents a fundamental advancement in our understanding of neural network optimization \citep{dahl2023algoperf}. The key insight is that effective optimizers must explicitly account for the statistical structure of feature learning, not merely approximate classical second-order methods designed for convex optimization.

\subsection{Implementation Details and Improvements to KFAC}

We first present the canonical implementation of KFAC, then analyze its computational benefits as well as some direct improvements to its empirical performance. Note that in order to facilitate quick comparison across the update rules following this section, we adopt this unified notation: $\mathbf{W}_t \in \mathbb{R}^{d_{\mathrm{out}} \times d_{\mathrm{in}}}$ denotes a weight matrix with gradient $\mathbf{G}_t = \partial \mathcal{L}/\partial \mathbf{W}$ at iteration $t$, layer input activations $\mathbf{h}_t \in \mathbb{R}^{d_{\mathrm{in}}}$, and backpropagated loss signals 
$\boldsymbol{\delta}_t = \partial \mathcal{L}/\partial \mathbf{a}_t \in \mathbb{R}^{d_{\mathrm{out}}}$. Here, $\mathbf{a}_t$ denotes the pre-activation (input to the nonlinearity) at step $t$. Wherever used in the algorithms below, we safely assume that $\boldsymbol{\delta}_t$ can be computed automatically by standard backpropagation routines in modern deep learning frameworks. The learning rate at step $t$ is $\eta_t > 0$.

\subsubsection{KFAC: The Baseline}

As discussed earlier, KFAC approximates the true Fisher information matrix via Kronecker factorization, storing activation and gradient covariances separately for each layer \citep{martens2015kfac}: 
\vspace{1em} 
\hrule
\noindent\textbf{Algorithm: KFAC Update}\label{alg:kfac}
\begin{algorithmic}[1]
\STATE \textbf{Input:} Current weights $\mathbf{W}_t$, gradient $\mathbf{G}_t$, activations $\mathbf{h}_t$, backprop $\boldsymbol{\delta}_t$
\STATE \textbf{Buffers:} Covariances of activations/gradients $\mathbf{A}_{t-1}$, $\mathbf{S}_{t-1}$
\STATE \textbf{Hyperparameters:} EMA decay $\beta_2$, damping $\lambda$, factorization $\pi_A$, $\pi_S$ where $\pi_A \pi_S = 1$, update frequency $T_{\text{inv}}$
\STATE
\STATE \textit{// Accumulate covariance statistics}
\STATE $\mathbf{A}_t \leftarrow \beta_2 \mathbf{A}_{t-1} + (1-\beta_2) \mathbf{h}_t \mathbf{h}_t^\top$
\STATE $\mathbf{S}_t \leftarrow \beta_2 \mathbf{S}_{t-1} + (1-\beta_2) \boldsymbol{\delta}_t \boldsymbol{\delta}_t^\top$
\STATE
\IF{$t \bmod T_{\text{inv}} = 0$}
    \STATE \textit{// Recompute damped inverses}
    \STATE $\tilde{\mathbf{A}}_t \leftarrow (\mathbf{A}_t + \pi_A \lambda \mathbf{I})$
    \STATE $\tilde{\mathbf{S}}_t \leftarrow (\mathbf{S}_t + \pi_S \lambda \mathbf{I})$
\ENDIF
\STATE
\STATE \textit{// Apply preconditioned update}
\STATE $\mathbf{W}_{t+1} \leftarrow \mathbf{W}_t - \eta_t \tilde{\mathbf{S}}_t \mathbf{G}_t \tilde{\mathbf{A}}_t$
\STATE \textbf{return} $\mathbf{W}_{t+1}$
\end{algorithmic}
\hrule 
\vspace{1em}

KFAC’s key insight is that the per-layer gradient matrix admits a Kronecker structure, \(\mathrm{vec}(\mathbf{G}_t) = \boldsymbol{\delta}_t \otimes \mathbf{h}_t\), which motivates the independence approximation that factorizes the Fisher as $\mathbf{F} \approx \mathbb{E}[\mathbf{h}\mathbf{h}^\top] \otimes \mathbb{E}[\boldsymbol{\delta}\boldsymbol{\delta}^\top] = \mathbf{A} \otimes \mathbf{S}$. In our implementation, the covariance factors $\mathbf{A}_t$ and $\mathbf{S}_t$ are maintained using exponential moving averages (EMA) rather than exact minibatch expectations. Similarly, we use fixed Tikhonov damping with a factored parameterization $\pi_A \pi_S = 1$, which preserves the Kronecker structure and enables
efficient inversion via $(\mathbf{A} \otimes \mathbf{S})^{-1} = \mathbf{A}^{-1} \otimes \mathbf{S}^{-1}$. To reduce computational cost, we follow standard KFAC practice and recompute the damped inverses only every $T_{\text{inv}}$ steps, reusing ``stale'' inverses in between updates. This assumption improves efficiency but can allow the eigenspectrum of the approximate Fisher to drift between updates. As demonstrated by \citet{george2018ekfac}, such drift can degrade the quality of the curvature approximation and motivate more frequent or spectrally aware updates (further discussed in Section \ref{alg:ekfac}).

\paragraph{Computational Benefits of KFAC.} KFAC's appeal stems from its ability to approximate second-order information while maintaining tractable computation. For a layer with weight matrix $\mathbf{W} \in \mathbb{R}^{d_{\text{out}} \times d_{\text{in}}}$, storing the full Fisher information matrix $\mathbf{F} \in \mathbb{R}^{(d_{\text{out}} d_{\text{in}}) \times (d_{\text{out}} d_{\text{in}})}$ requires $\mathcal{O}((d_{\text{out}} d_{\text{in}})^2)$ memory. KFAC's Kronecker factorization $\mathbf{F} \approx \mathbf{A} \otimes \mathbf{S}$ reduces this to $\mathcal{O}(d_{\text{in}}^2 + d_{\text{out}}^2)$ by storing only two smaller matrices: the activation covariance $\mathbf{A} = \mathbb{E}[\mathbf{h}\mathbf{h}^\top] \in \mathbb{R}^{d_{\text{in}} \times d_{\text{in}}}$ and the backpropagated gradient covariance $\mathbf{S} = \mathbb{E}[\boldsymbol{\delta}\boldsymbol{\delta}^\top] \in \mathbb{R}^{d_{\text{out}} \times d_{\text{out}}}$, where $\mathbf{h}$ denotes layer input activations and $\boldsymbol{\delta} = \frac{\partial \mathcal{L}}{\partial \mathbf{a}}$ is the backpropagated gradient with respect to pre-activations \citep{martens2015kfac}.

The Kronecker product structure enables efficient inversion through the identity $(\mathbf{A} \otimes \mathbf{S})^{-1} = \mathbf{A}^{-1} \otimes \mathbf{S}^{-1}$. This reduces matrix inversion from $\mathcal{O}((d_{\text{out}} d_{\text{in}})^3)$ to $\mathcal{O}(d_{\text{in}}^3 + d_{\text{out}}^3)$ --- a substantial reduction for typical layer dimensions. Furthermore, KFAC's incorporation of curvature information allows it to take larger, more confident steps than first-order methods, making it particularly effective in the large batch regime where data-parallel scaling is critical \citep{ba2017distributed}.

\paragraph{Fundamental Challenges.} While KFAC provides a theoretically elegant and empirically effective approximation to natural gradient descent, scaling it to modern deep learning workloads --- models with billions of parameters trained on distributed infrastructure --- presents significant computational and engineering challenges. First, computing the Kronecker factors $\mathbf{A}$ and $\mathbf{S}$, and especially their inverses, is still computationally expensive (albeit relatively cheap compared to natural gradient descent and full curvature matrix preconditioning). In practice, these quantities must be updated infrequently ---  \citet{martens2015kfac} suggest recomputing inverses every 10-100 iterations, justified by the heuristic that curvature changes relatively slowly. However, \citet{george2018ekfac} demonstrate empirically that the eigenspectrum of the KFAC approximation $\mathbf{G}_{\text{KFAC}}$ can diverge rapidly from the true Fisher $\mathbf{F}$ between inverse recomputations, suggesting that staleness introduces significant approximation error.

KFAC also introduces additional hyperparameters beyond the learning rate. To ensure numerical stability and positive definiteness, practitioners add Tikhonov damping $\lambda \mathbf{I}$ to the approximate Fisher. The factored damping approach preserves Kronecker structure by adding $\pi_A \lambda \mathbf{I}$ to $\mathbf{A}$ and $\pi_S \lambda \mathbf{I}$ to $\mathbf{S}$ with $\pi_A \pi_S = 1$ \citep{martens2015kfac}. Choosing appropriate values for $\lambda$, $\pi_A$, and $\pi_S$, along with update frequencies for factor estimation and inverse computation, requires careful tuning that is often problem-dependent.

Extending KFAC beyond fully-connected layers introduces additional complexity. For convolutional layers, \citet{martens2015kfac} develop the Kronecker Factors for Convolution (KFC) approximation, which makes three key assumptions: (1) independence between activations and derivatives, (2) spatial homogeneity, meaning statistical relationships depend only on relative spatial distance, and (3) spatially uncorrelated derivatives, which assumes pre-activation derivatives at different spatial locations are uncorrelated \citep{grosse2016kronecker}. Under these assumptions, the Fisher block for a convolutional layer decomposes as $\mathbf{F}_\ell \approx \mathbf{\Omega}_{\ell-1} \otimes \mathbf{\Gamma}_\ell$, where $\mathbf{\Omega}_{\ell-1}$ captures correlations across feature maps and spatial offsets, and $\mathbf{\Gamma}_\ell$ captures correlations between output feature maps. While assumption (3) may seem restrictive, empirical analysis by \citet{grosse2016kronecker} shows it holds remarkably well for networks using max-pooling, as the sparsifying effect of max-pooling naturally decorrelates spatial derivatives.

For recurrent neural networks, \citet{martens2018kfac} extend KFAC by modeling the statistical dependencies between gradient contributions across time steps using a chain-structured linear Gaussian graphical model. The key insight is that temporal correlations in RNN gradients can be captured by assuming the gradient contributions evolve according to $\mathbf{w}_t = \mathbf{\Psi}\mathbf{w}_{t-1} + \boldsymbol{\epsilon}_t$ where $\mathbf{\Psi}$ is a transition matrix with spectral radius less than 1, and $\boldsymbol{\epsilon}_t$ are i.i.d. Gaussian noise terms. Under this model, they derive two practical approximations: one assuming the transition matrix is symmetric, and another using a limiting formulation for long sequences, both of which preserve the Kronecker product structure necessary for efficient inversion. Experiments on Penn TreeBank language modeling and differentiable neural computers demonstrate that this approach significantly outperforms SGD and Adam, achieving comparable training loss in 10-20 times fewer iterations \citep{martens2018kfac}.

Despite these challenges, KFAC demonstrates clear advantages in certain regimes: medium-scale models where per-iteration overhead remains tolerable, large batch training where second-order information enables aggressive step sizes, optimization landscapes with highly imbalanced curvature, and sample-limited scenarios where reducing iteration count matters more than minimizing wall-clock time per step. These observations motivate the development of practical variants that retain KFAC's geometric insights while addressing its engineering limitations.

\subsubsection{EKFAC: Optimal Diagonal in Kronecker Eigenbasis}

While KFAC provides a strong foundation for Kronecker-factored methods, many have noted that infrequent inverse updates can lead to stale curvature estimates that degrade performance between recomputations. Eigenvalue-corrected KFAC (EKFAC) improves upon this by tracking eigenvalues cheaply while caching eigenvectors, yielding a provably better approximation \citep{george2018ekfac}.

\vspace{1em} 
\hrule
\noindent\textbf{Algorithm: EKFAC Update}\label{alg:ekfac}
\begin{algorithmic}[1]
\STATE \textbf{Input:} Current weights $\mathbf{W}_t$, gradient $\mathbf{G}_t$, activations $\mathbf{h}_t$, backprop $\boldsymbol{\delta}_t$
\STATE \textbf{Buffers:} Covariances $\mathbf{A}_{t-1}$, $\mathbf{S}_{t-1}$, eigenvectors $\mathbf{U}_A$, $\mathbf{U}_S$, diagonal estimates $\mathbf{s}^*_{t-1}$
\STATE \textbf{Hyperparameters:} EMA decay $\beta_2$, damping $\lambda$, eigenbasis update frequency $T_{\text{eig}}$
\STATE
\STATE \textit{// Update covariances via EMAs}
\STATE $\mathbf{A}_t \leftarrow \beta_2 \mathbf{A}_{t-1} + (1-\beta_2) \mathbf{h}_t \mathbf{h}_t^\top$
\STATE $\mathbf{S}_t \leftarrow \beta_2 \mathbf{S}_{t-1} + (1-\beta_2) \boldsymbol{\delta}_t \boldsymbol{\delta}_t^\top$
\STATE
\IF{$t \bmod T_{\text{eig}} = 0$}
    \STATE \textit{// Recompute eigenbasis (expensive, infrequent)}
    \STATE $\mathbf{U}_A, \boldsymbol{\Sigma}_A \leftarrow \text{eig}(\mathbf{A}_t)$
    \STATE $\mathbf{U}_S, \boldsymbol{\Sigma}_S \leftarrow \text{eig}(\mathbf{S}_t)$
\ENDIF
\STATE
\STATE \textit{// Project gradient to Kronecker eigenbasis (from Kronecker factored theorem)}
\STATE $\tilde{\mathbf{G}}_t \leftarrow \mathbf{U}_S^\top \mathbf{G}_t \mathbf{U}_A$
\STATE
\STATE \textit{// Update diagonal scaling estimates (every step)}
\STATE $\mathbf{s}^*_t \leftarrow \beta_2 \, \mathbf{s}^*_{t-1} + (1-\beta_2) \, (\tilde{\mathbf{G}}_t \odot \tilde{\mathbf{G}}_t)$
\STATE
\STATE \textit{// Apply diagonal preconditioning in eigenbasis}
\STATE $\hat{\mathbf{G}}_t \leftarrow \tilde{\mathbf{G}}_t / (\mathbf{s}^*_t + \lambda)$
\STATE
\STATE \textit{// Transform back to parameter space}
\STATE $\mathbf{W}_{t+1} \leftarrow \mathbf{W}_t - \eta_t \, \mathbf{U}_S \hat{\mathbf{G}}_t \mathbf{U}_A^\top$
\STATE \textbf{return} $\mathbf{W}_{t+1}$
\end{algorithmic}

\hrule
\vspace{1em} 

EKFAC addresses KFAC's staleness by separating eigenbasis computation (expensive, infrequent) from eigenvalue tracking (cheap, frequent). The key insight is that the diagonal of the true Fisher in the Kronecker eigenbasis is exactly the second moment of the projected gradients:
\begin{equation}
    \text{diag}(\tilde{\mathbf{F}}) = \mathbb{E}[\tilde{\mathbf{g}} \odot \tilde{\mathbf{g}}]
\end{equation}
where $\tilde{\mathbf{g}} = (\mathbf{U}_A \otimes \mathbf{U}_S)^\top \text{vec}(\mathbf{G})$. This follows directly from $\tilde{\mathbf{F}} = \mathbb{E}[\tilde{\mathbf{g}}\tilde{\mathbf{g}}^\top]$, whose diagonal entries are $\mathbb{E}[\tilde{g}_{ij}^2]$. Thus EKFAC's running estimate $\mathbf{s}^*$ targets the true Fisher diagonal, not the Kronecker-approximated eigenvalues $\sigma_A^{(i)} \sigma_S^{(j)}$ that KFAC implicitly uses. By updating $\mathbf{s}^*$ every step, EKFAC tracks curvature changes that KFAC misses between inverse recomputations. The diagonal rescaling $\tilde{\mathbf{G}} / (\mathbf{s}^* + \lambda)$ is equivalent to applying $\text{diag}(\tilde{\mathbf{F}} + \lambda \mathbf{I})^{-1}$ --- no explicit matrix inverse is needed! \citet{george2018ekfac} demonstrate empirically that EKFAC's eigenspectrum remains closer to the true Fisher than KFAC's throughout training.

\subsection{Shampoo: A New Perspective on Layer-wise Preconditioning}

Both KFAC and EKFAC rely on the independence assumption between activations and backpropagated gradients, which may not hold in practice and can be notably unstable during usage. Shampoo offers an alternative approach to approximating the empirical Fisher matrix --- and by extension the GGN --- by accumulating gradient outer products  \citep{gupta2018shampoo}. As established in Section \ref{sec:curvmat}, the empirical Fisher matrix serves as a tractable approximation to the GGN (via its equivalency to the true Fisher for typical deep learning losses), inheriting its positive semi-definite structure and curvature information without requiring second-order derivatives: Shampoo implicitly targets this same curvature structure by accumulating gradient  products while avoiding the explicit independence assumptions required by KFAC. Furthermore, Shampoo employs a matrix power of $p = 1/4$ for its Kronecker factors, occupying a middle ground between the full inversion ($p = 1$) used by KFAC and the square root inverse ($p = 1/2$) of AdaGrad; this partial preconditioning provides a tempered curvature correction that empirically improves performance on generalization while still capturing meaningful second-order information. 

Shampoo leverages the Kronecker inverse product rule to maintain separate left and right preconditioners $L_t \in \mathbb{R}^{m \times m}$ and $R_t \in \mathbb{R}^{n \times n}$, which accumulate row-space and column-space statistics of the gradients. This Kronecker factorization reduces both memory and computational costs from $O(m^2n^2)$ (for full AdaGrad-like conditioning) to $O(m^2 + n^2)$. We present the Shampoo update rule in the same notation used in Section \ref{alg:kfac} for convenience:

\vspace{1em} 
\hrule
\noindent\textbf{Algorithm: Shampoo Update}
\label{alg:shampoo}
\begin{algorithmic}[1]
\STATE \textbf{Input:} Current weights $\mathbf{W}_t$, gradient $\mathbf{G}_t$
\STATE \textbf{Buffers:} Left/right preconditioners $\mathbf{L}_{t-1}$, $\mathbf{R}_{t-1}$, cached inverses $\tilde{\mathbf{L}}_{t-1}$, $\tilde{\mathbf{R}}_{t-1}$
\STATE \textbf{Hyperparameters:} EMA decay $\beta_2$, damping $\lambda$, update frequency $T_{\text{prec}}$
\STATE
\STATE \textit{// Accumulate left and right preconditioners}
\STATE $\mathbf{L}_t \leftarrow \beta_2 \mathbf{L}_{t-1} + (1-\beta_2) \mathbf{G}_t \mathbf{G}_t^\top$
\STATE $\mathbf{R}_t \leftarrow \beta_2 \mathbf{R}_{t-1} + (1-\beta_2) \mathbf{G}_t^\top \mathbf{G}_t$
\STATE
\IF{$t \bmod T_{\text{prec}} = 0$}
    \STATE \textit{// Compute matrix power (typically $p=1/4$)}
    \STATE $\tilde{\mathbf{L}}_t \leftarrow (\mathbf{L}_t + \lambda \mathbf{I})^{-1/4}$
    \STATE $\tilde{\mathbf{R}}_t \leftarrow (\mathbf{R}_t + \lambda \mathbf{I})^{-1/4}$
\ENDIF
\STATE
\STATE \textit{// Apply preconditioned update}
\STATE $\mathbf{W}_{t+1} \leftarrow \mathbf{W}_t - \eta_t \tilde{\mathbf{L}}_t \mathbf{G}_t \tilde{\mathbf{R}}_t$
\STATE \textbf{return} $\mathbf{W}_{t+1}$
\end{algorithmic}
\hrule \vspace{1em}

Shampoo's use of $\mathbf{L}_t = \mathbb{E}[\mathbf{G}\mathbf{G}^\top]$ and $\mathbf{R}_t = \mathbb{E}[\mathbf{G}^\top\mathbf{G}]$ differs from KFAC's $\mathbf{A} = \mathbb{E}[\mathbf{h}\mathbf{h}^\top]$ and $\mathbf{S} = \mathbb{E}[\boldsymbol{\delta}\boldsymbol{\delta}^\top]$, directly capturing gradient second moments rather than separately modeling activations and backpropagation. 

Recent theoretical work reveals that the \textit{square} of Shampoo's preconditioner approximation is equivalent to a single step of the power iteration algorithm for computing the optimal Kronecker product approximation of these matrices, providing explicit justification for why Shampoo's approximation is close to optimal \citep{morwani2024Shampoo}. Moreover, when EMA is disabled ($\beta_2 = 0$), Shampoo without accumulation reduces to orthogonalized gradient descent, which is an important idea which we discuss in Section \ref{sec:modnorm}. Specifically, substituting the SVD $\mathbf{G}_t = \mathbf{U}_t \boldsymbol{\Sigma}_t \mathbf{V}_t^\top$ into the update yields $\mathbf{W}_{t+1} = \mathbf{W}_t - \eta_t \mathbf{U}_t \mathbf{V}_t^\top$, which is precisely the projection onto semi-orthogonal matrices and thus steepest descent under the spectral norm \citep{bernstein2024normanthology}.

A critical advantage of Shampoo over KFAC emerges in the infinite-width limit: Shampoo has theoretical guarantees that seem to preserve feature learning capabilities as the network grows. Under maximal update parameterization $\mu P$, KFAC can exhibit an implicit bias toward the ``neural network Gaussian process'' regime, particularly with zero initialization of the output layer --- a common practice in modern implementations \citep{ishikawa2024local}. The use of exponent $p = 1/4$ plays a crucial role in this behavior. While early work cited the use of $1/4$ primarily for numerical stability, the connection to power iteration revealed by \citet{morwani2024Shampoo} suggests a deeper reason: this choice balances between approximating the optimal Kronecker factorization (which would require full convergence of power iteration) and maintaining computational efficiency (achieved with a single iteration). Empirically, some work finds $p=1/2$ as the better choice \citep{anil2020scalable}. Further work is being done to investigate what other roles does this fraction play in Shampoo's behavior.

\paragraph{Distributed Shampoo.}
While the theoretical properties of Shampoo are compelling, native implementations can incur 50-75\% slower wall-clock time per iteration compared to diagonal adaptive gradient methods due to the replacement of element-wise operations with matrix multiplications. \citet{shi2023distributed}\ address this gap with Distributed Shampoo, a PyTorch implementation optimized for multi-GPU distributed data-parallel training. The key insight is to distribute both the memory and computation of preconditioners across workers rather than replicating them: each worker computes only a subset of search directions based on a greedy load-balancing assignment, followed by an ``AllGather'' primitive to synchronize the full search directions. This reduces per-worker memory by approximately a factor of the process group size and achieves at most a 10\% increase in per-step wall-clock time compared to Adam.

A critical practical contribution is the incorporation of \textit{learning rate grafting}, which maintains a parallel ``grafted'' optimizer (typically Adam or SGD) running on the same iterate sequence. At each step, the Shampoo search direction is rescaled to match the norm of the grafted method's search direction on a per-layer basis:
\begin{equation}
    \mathbf{w}^{(i)}_{t+1} = \mathbf{w}^{(i)}_t - \eta_t \frac{\|\mathbf{P}^{(i)}_{t,\text{graft}}\|_F}{\|\mathbf{P}^{(i)}_{t,\text{Shampoo}}\|_F} \mathbf{P}^{(i)}_{t,\text{Shampoo}}.
\end{equation}
This enables practitioners to transfer pre-existing learning rate schedules from well-tuned baselines, making Shampoo significantly easier to adopt without extensive hyperparameter search. Additional heuristics include delaying Shampoo preconditioning until the preconditioners stabilize (using the grafted method in early iterations), periodic root inverse computation to amortize costs, and blocking/merging strategies for handling large-dimensional tensors. On ImageNet ResNet50, Distributed Shampoo achieves the same validation accuracy in 60 epochs that SGD with Nesterov requires 90 epochs to reach, yielding a 1.35$\times$ improvement in overall wall-clock time. Notably, this implementation of Shampoo was used to achieve top rankings in the AlgoPerf optimization competition, demonstrating its practical effectiveness at scale. We cover more engineering practices that enable the use of curvature-aware methods in Section \ref{sec:eng_dlo}.

We now examine a family of optimizers that are derived from the Shampoo update rule, each making different trade-offs between approximation fidelity, computational efficiency, and practical usability. We will use the same notation from above to facilitate easy comparison of these methods.

\subsubsection{SOAP: ``Adam'' in Shampoo's Rotated Space}

Despite Shampoo's effectiveness and theoretical justification, its matrix power computations can still impose significant computational overhead, motivating the search for more even more efficient approximations. Extending the hygiene lexicon, ``SOAP'' establishes that Shampoo with $p = 1/2$ is equivalent to running full AdaGrad in the eigenbasis of the preconditioner, then exploits this to reduce computational cost \citep{vyas2024soap}:

\vspace{1em} 
\hrule
\noindent\textbf{Algorithm: SOAP Update}
\label{alg:soap}
\begin{algorithmic}[1]
\STATE \textbf{Input:} Current weights $\mathbf{W}_t$, gradient $\mathbf{G}_t$
\STATE \textbf{Buffers:} Second moment $\mathbf{V}_{t-1}$, left/right covariances $\mathbf{L}_{t-1}$, $\mathbf{R}_{t-1}$, eigenvector matrices $\mathbf{U}_L$, $\mathbf{U}_R$
\STATE \textbf{Hyperparameters:} EMA decays $\beta_1, \beta_2$, epsilon $\epsilon$, eigenbasis update frequency $T_{\text{prec}}$
\STATE
\STATE $\mathbf{L}_t \leftarrow \beta_2 \mathbf{L}_{t-1} + (1-\beta_2) \mathbf{G}_t \mathbf{G}_t^\top$
\STATE $\mathbf{R}_t \leftarrow \beta_2 \mathbf{R}_{t-1} + (1-\beta_2) \mathbf{G}_t^\top \mathbf{G}_t$
\STATE
\IF{$t \bmod T_{\text{prec}} = 0$}
    \STATE \textit{// Recompute eigenbases (infrequent)}
    \STATE $\mathbf{U}_L \boldsymbol{\Lambda}_L \mathbf{U}_L^\top \leftarrow \text{eig}(\mathbf{L}_t)$
    \STATE $\mathbf{U}_R \boldsymbol{\Lambda}_R \mathbf{U}_R^\top \leftarrow \text{eig}(\mathbf{R}_t)$
\ENDIF
\STATE
\STATE \textit{// Transform gradient to eigenbasis}
\STATE $\tilde{\mathbf{G}}_t \leftarrow \mathbf{U}_L^\top \mathbf{G}_t \mathbf{U}_R$
\STATE
\STATE \textit{// Update second moment in eigenbasis (element-wise)}
\STATE $\mathbf{V}_t \leftarrow \beta_2 \mathbf{V}_{t-1} + (1-\beta_2) \tilde{\mathbf{G}}_t \odot \tilde{\mathbf{G}}_t$
\STATE
\STATE \textit{// Apply Adam-like rescaling in eigenbasis}
\STATE $\hat{\mathbf{G}}_t \leftarrow \tilde{\mathbf{G}}_t / (\sqrt{\mathbf{V}_t} + \epsilon)$
\STATE
\STATE \textit{// Transform back to parameter space}
\STATE $\mathbf{W}_{t+1} \leftarrow \mathbf{W}_t - \eta_t \mathbf{U}_L \hat{\mathbf{G}}_t \mathbf{U}_R^\top$
\STATE \textbf{return} $\mathbf{W}_{t+1}$
\end{algorithmic}
\hrule \vspace{1em}

SOAP's conceptual insight is that Shampoo's preconditioner $(\mathbf{L} + \lambda \mathbf{I})^{-1/2} \mathbf{G} (\mathbf{R} + \lambda \mathbf{I})^{-1/2}$ can be rewritten as $\mathbf{U}_L \boldsymbol{\Lambda}_L^{-1/2} \mathbf{U}_L^\top \mathbf{G} \mathbf{U}_R \boldsymbol{\Lambda}_R^{-1/2} \mathbf{U}_R^\top$, which is precisely a coordinate transformation to the eigenbasis $(\mathbf{U}_L, \mathbf{U}_R)$, followed by diagonal rescaling. SOAP maintains this eigenbasis but updates the diagonal rescaling every step using Adam-style second moment estimation, mitigating the staleness problem identified in KFAC. The method reduces computational overhead from Shampoo's frequent matrix root computation to infrequent eigendecomposition plus cheap rotation and element-wise operations. \citet{vyas2024soap}\ demonstrate 40\% iteration reduction and 35\% wall-clock speedup over AdamW on language model pretraining (360M-660M parameters) on batch sizes of 256k, with approximately 20\% improvements over standard Shampoo.

SOAP's key insight is that Shampoo's preconditioner
\[
(\mathbf{L} + \lambda \mathbf{I})^{-1/2} \mathbf{G} (\mathbf{R} + \lambda \mathbf{I})^{-1/2}
\]
can be equivalently expressed as
\[
\mathbf{U}_L \boldsymbol{\Lambda}_L^{-1/2} \mathbf{U}_L^\top \mathbf{G} \mathbf{U}_R \boldsymbol{\Lambda}_R^{-1/2} \mathbf{U}_R^\top,
\]
which corresponds to a coordinate transformation to the eigenbasis $(\mathbf{U}_L, \mathbf{U}_R)$ followed by a diagonal rescaling \citep{vyas2024soap}. SOAP leverages this structure by maintaining the eigenbasis infrequently and updating the diagonal rescaling every step using Adam-style second moment estimation, effectively mitigating the staleness problem identified in KFAC. This approach drastically reduces the computational cost of Shampoo: rather than computing 
matrix square roots at every step, SOAP only performs infrequent eigendecompositions, followed by cheap rotations and element-wise updates in the rotated space. In large-scale language model pretraining experiments (360M–660M parameters, batch size 256k),  \citet{vyas2024soap} report approximately 40\% fewer iterations and a 35\% wall-clock speedup compared to AdamW, with roughly 20\% improvement over standard Shampoo. 

Moreover, an important distinction to note here is that SOAP is applied only to 2D parameters, such as the weight matrices of fully-connected or  linear layers, because the eigenbasis decomposition and rotation operations scale quadratically with the dimension of the parameter matrix, making them prohibitively expensive for higher-dimensional tensors such as embeddings or convolutional kernels. Additionally, embeddings and output layers often have extremely large dimensions (e.g., vocabulary size in language models), where computing and storing the full preconditioner would exceed memory limits and offer little optimization 
benefit due to their sparse usage patterns \citep{vyas2024soap}. For these layers, SOAP relies on standard AdamW updates, which are both memory- and computation-efficient, while retaining the advantages of second-moment rescaling. This design choice balances the benefits of Shampoo-style preconditioning on typical 2D layers with practical feasibility for very large-scale models.

\subsubsection{SPlus: Stable Whitening via Bounded Updates}

Although SOAP successfully reduces computational overhead through cached eigenbases, it can still exhibit training instabilities from stale preconditioners (as originally noted in Shampoo development) and require careful learning rate tuning when scaling to different model sizes. SPlus addresses three key instabilities in naive Shampoo: divergence from stale preconditioners, poor learning rate transfer, and noise amplification at high learning rates \citep{frans2025splus}:

\vspace{1em} 
\hrule
\noindent\textbf{Algorithm: SPlus Update}
\label{alg:splus}
\begin{algorithmic}[1]
\STATE \textbf{Input:} Current weights $\mathbf{W}_t$, gradient $\mathbf{G}_t$
\STATE \textbf{Buffers:} Iterate average $\bar{\mathbf{W}}_{t-1}$, left/right covariances $\mathbf{L}_{t-1}$, $\mathbf{R}_{t-1}$, eigenvector matrices $\mathbf{U}_L$, $\mathbf{U}_R$
\STATE \textbf{Hyperparameters:} Learning rate $\eta_t$, EMA decay $\beta_2$, eigenbasis update frequency $T_{\text{prec}}$, averaging $\alpha$
\STATE
\STATE $\mathbf{L}_t \leftarrow \beta_2 \mathbf{L}_{t-1} + (1-\beta_2) \mathbf{G}_t \mathbf{G}_t^\top$
\STATE $\mathbf{R}_t \leftarrow \beta_2 \mathbf{R}_{t-1} + (1-\beta_2) \mathbf{G}_t^\top \mathbf{G}_t$
\STATE
\IF{$t \bmod T_{\text{prec}} = 0$}
    \STATE \textit{// Update eigenbasis (historical curvature)}
    \STATE $\mathbf{U}_L \boldsymbol{\Lambda}_L \mathbf{U}_L^\top \leftarrow \text{eig}(\mathbf{L}_t)$
    \STATE $\mathbf{U}_R \boldsymbol{\Lambda}_R \mathbf{U}_R^\top \leftarrow \text{eig}(\mathbf{R}_t)$
\ENDIF
\STATE
\STATE \textit{// Instantaneous normalization (key innovation)}
\STATE $\tilde{\mathbf{G}}_t \leftarrow \mathbf{U}_L^\top \mathbf{G}_t \mathbf{U}_R$ \COMMENT{Project to eigenbasis}
\STATE $\hat{\mathbf{G}}_t \leftarrow \tilde{\mathbf{G}}_t / \|\tilde{\mathbf{G}}_t\|_F$ \COMMENT{Normalize instantaneously}
\STATE
\STATE \textit{// Shape-aware scaling}
\STATE $\eta_{\text{eff}} \leftarrow \eta_t \cdot \sqrt{d_{\text{out}} / d_{\text{in}}}$ \COMMENT{Dimension-dependent factor}
\STATE
\STATE \textit{// Transform back and update}
\STATE $\mathbf{W}_{t+1} \leftarrow \mathbf{W}_t - \eta_{\text{eff}} \mathbf{U}_L \hat{\mathbf{G}}_t \mathbf{U}_R^\top$
\STATE
\STATE \textit{// EMA for descent direction}
\STATE $\bar{\mathbf{W}}_t \leftarrow \alpha \bar{\mathbf{W}}_{t-1} + (1-\alpha) \mathbf{W}_{t+1}$
\STATE \textbf{return} $\mathbf{W}_{t+1}$
\end{algorithmic}
\hrule \vspace{1em}

\begin{figure}
    \centering
    \includegraphics[width=0.75\linewidth]{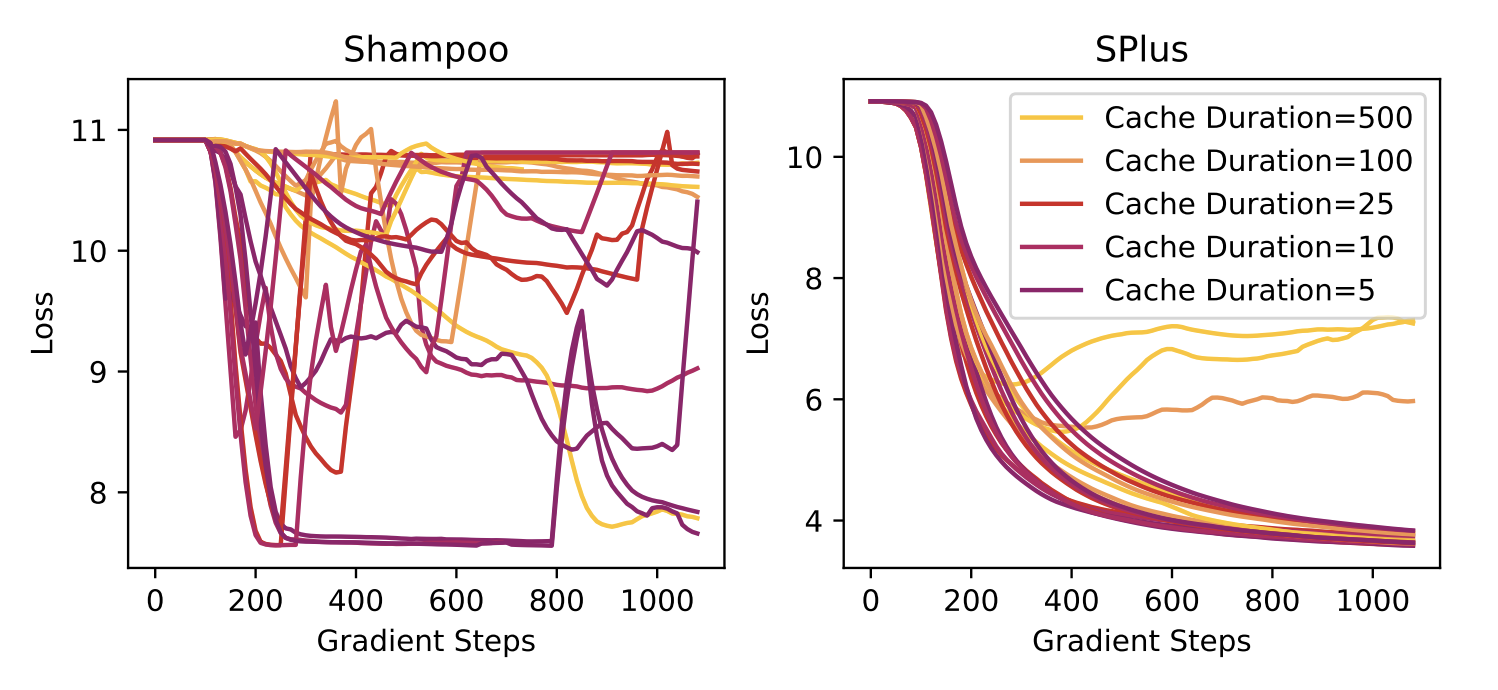}
    \caption{Figure 2 from \citep{frans2025splus} depicting the instability of Shampoo descent compared to SPlus over various cache durations.}
    \label{fig:splus_shampoo}
\end{figure}

SPlus introduces three innovations to stabilize Shampoo. First, instantaneous normalization in the eigenbasis prevents divergence by bounding the update magnitude without requiring a fresh inverse computation; the historical eigenbasis captures accumulated curvature while normalization in the Frobenius norm prevents explosion when that basis becomes stale (see Figure \ref{fig:splus_shampoo} for a study on stability over cache durations). Second, shape-aware scaling incorporates a dimension-dependent factor (which is analogous to the RMS$\to$RMS norm's $\sqrt{d_{\text{in}}/d_{\text{out}}}$ from the modular norm framework that we plan to discuss in a further section) to improve the learning rate transfer across network widths. This prevents the need to do learning rate sweeps as you increase the number of parameters in the network. Third, the EMA maintains a smoothed parameter estimate $\bar{\mathbf{W}}_t$ used for the final set of weights, which dampens high-frequency noise introduced by aggressive learning rates earlier in the training process without reducing the effective step size. \citet{frans2025splus} demonstrate that on a pointed transformer benchmark spanning language modeling, image classification, and diffusion, SPlus reaches AdamW's validation performance in approximately 44-58\% of gradient steps and 62-83\% of wall-clock time on average, with significantly improved stability across hyperparameter sweeps compared to naive Shampoo.

\subsubsection{Muon: Momentum with Efficient Orthogonalization}

The idea, hinted by Shampoo, that orthogonalized gradient descent was viable and effective at training neural networks. However, the primary concern with exploring alternative algorithms that directly compute orthogonal matrices is the high cost of SVD ($\mathcal{O}(n^3)$ for $n >> m$). To remedy this, Muon combines momentum with gradient orthogonalization, using Newton-Schulz iteration to avoid expensive SVD computations \citep{jordan2024muon}.

\vspace{1em} 
\hrule
\noindent\textbf{Algorithm: Muon Update}
\label{alg:muon}
\begin{algorithmic}[1]
\STATE \textbf{Input:} Current weights $\mathbf{W}_t$, gradient $\mathbf{G}_t$
\STATE \textbf{Buffers:} Momentum buffer $\mathbf{M}_{t-1}$
\STATE \textbf{Hyperparameters:} Learning rate $\eta_t$, momentum $\beta_1 = 0.95$, NS steps $T_{\text{NS}} = 5$, NS coefficients $(a,b,c)$
\STATE
\STATE \textit{// Nesterov momentum}
\STATE $\mathbf{M}_t \leftarrow \beta_1 \mathbf{M}_{t-1} + \mathbf{G}_t$
\STATE $\tilde{\mathbf{M}}_t \leftarrow \mathbf{M}_t + \beta_1 (\mathbf{M}_t - \mathbf{M}_{t-1})$ \COMMENT{Nesterov lookahead}
\STATE
\STATE \textit{// Newton-Schulz orthogonalization}
\STATE $\mathbf{X}_0 \leftarrow \tilde{\mathbf{M}}_t / \|\tilde{\mathbf{M}}_t\|_F$ \COMMENT{Frobenius normalization}
\IF{$d_{\text{out}} > d_{\text{in}}$}
    \STATE $\mathbf{X}_0 \leftarrow \mathbf{X}_0^\top$ \COMMENT{Transpose for efficiency}
\ENDIF
\FOR{$i = 1$ to $T_{\text{NS}}$}
    \STATE $\mathbf{A} \leftarrow \mathbf{X}_{i-1} \mathbf{X}_{i-1}^\top$
    \STATE $\mathbf{B} \leftarrow b \mathbf{A} + c \mathbf{A}^2$ \COMMENT{$b = -4.7750$, $c = 2.0315$}
    \STATE $\mathbf{X}_i \leftarrow a \mathbf{X}_{i-1} + \mathbf{B} \mathbf{X}_{i-1}$ \COMMENT{$a = 3.4445$}
\ENDFOR
\IF{$d_{\text{out}} > d_{\text{in}}$}
    \STATE $\mathbf{X}_{T_{\text{NS}}} \leftarrow \mathbf{X}_{T_{\text{NS}}}^\top$ \COMMENT{Transpose back}
\ENDIF
\STATE
\STATE \textit{// Apply orthogonalized momentum update}
\STATE $\mathbf{W}_{t+1} \leftarrow \mathbf{W}_t - \eta_t \mathbf{X}_{T_{\text{NS}}}$
\STATE \textbf{return} $\mathbf{W}_{t+1}$
\end{algorithmic}
\hrule \vspace{1em}

Muon's key innovation is applying momentum \emph{before} orthogonalization, which empirically outperforms the reverse ordering used in prior work \citep{tuddenham2022orthogonal}. The Newton-Schulz iteration approximates the matrix sign function $\text{msign}(\mathbf{M})$  via a 5-step polynomial iteration of the form
\[
X_0 = \frac{\tilde{\mathbf{M}}_t}{\|\tilde{\mathbf{M}}_t\|_F}, \qquad
X_{k+1} = a X_k + \big( b X_k X_k^\top + c (X_k X_k^\top)^2 \big) X_k, 
\quad k = 0,\dots,T_\text{NS}-1,
\]
with tuned coefficients $(a, b, c) = (3.4445, -4.7750, 2.0315)$, which enables convergence in just 5 steps while remaining stable even in bfloat16 precision. This is a critical advantage over the coupled Newton iteration used in Shampoo implementations, which historically required float32 precision. 

Despite performing several operations per iteration, the FLOP overhead is minimal: for typical transformer dimensions ($m = 768$, batch size $B = 524288$), the overhead is
\[
\frac{T_\text{NS} \cdot m}{B} = \frac{5 \cdot 768}{524288} < 1\%,
\]
which is negligible in practice. As Keller Jordan explains in his popular blog, Muon can be viewed as ``instantaneous Shampoo'' (without EMA) with efficient orthogonalization. Moreover, it can only be applied to 2D weight matrices and is typically not used for input/output layers, which means it still requires the usage of another optimizer (popularly AdamW) for the rest of the weights of the network \citep{jordan2024muon}. This is partly due to fact that the orthogonalization procedure in Muon relies on Newton-Schulz iterations that converge well for matrices where singular values are in a reasonable range. For very rectangular matrices (such as embeddings), the optimization dynamics would be vastly different, and the iterations may not converge properly or require many more steps. Muon quickly gained traction in the optimization community as it quickly broke the NanoGPT record for training GPT-2 to 3.28 cross entropy loss on the FineWeb dataset \citep{nanogpt}. Empirically, industry researchers find that it works well in other language model pre-training tasks \citep{shah2025practical}.

\subsection{Engineering Curvature Methods at Scale}\label{sec:eng_dlo}

While KFAC, Shampoo, and other higher-order methods offer superior convergence properties compared to first-order optimizers, their practical deployment at scale has historically been limited by computational and memory constraints. Recent engineering advances have systematically addressed these bottlenecks through algorithmic innovations, distributed system designs, and numerical techniques that preserve the theoretical benefits while achieving practical efficiency. The following engineering strategies have enabled second-order methods to match or exceed first-order methods in wall-clock time on modern hardware.

\paragraph{Memory efficiency.} Mixed-precision storage reduces preconditioner memory while maintaining numerical stability, with CPU offloading of inverse and eigendecomposition operations exploiting heterogeneous hardware architectures \citep{anil2020scalable}. Selective application strategies apply second-order updates only to convolutional and fully-connected layers, leaving embeddings and normalization layers to first-order methods.

\paragraph{Computational amortization.} SOAP decouples moment tracking from eigenbasis computation, performing expensive eigendecompositions only every 100-1000 steps while updating second moments at each iteration, keeping overhead below 5\% \citep{vyas2024soap}. Muon eliminates eigendecompositions entirely through Newton-Schulz iteration in bfloat16, requiring only 5 fixed-point iterations with sub-1\% FLOP overhead for typical batch sizes \citep{jordan2024muon}.

\paragraph{Distributed synchronization.} Asynchronous computation of Fisher inverses (for KFAC-style methods) on CPU while GPUs compute gradients provides 40-50\% wall-clock speedup, with statistics workers computing second moments on separate data shards independently of gradient workers \citep{ba2017distributed}. Hierarchical $\textit{allreduce}$ patterns aggregate Kronecker factors less frequently than gradients (every 20-100 steps), exploiting the slow evolution of curvature compared to first-order statistics. This work helped motivate the development of Distributed Shampoo, which we showed earlier was important to getting the original Shampoo method to work effectively \citep{shi2023distributed}.

\paragraph{Numerical stability.} Factored Tikhonov damping preserves Kronecker structure while adding $\lambda I$ through dimension-aware scaling factors, avoiding expensive eigendecompositions required by standard damping \citep{martens2015kfac,ba2017distributed}. SPlus introduces instant-sign normalization that bounds updates through orthonormal eigenbases and sign matrices, preventing divergence even with cached preconditioners used for hundreds of steps \citep{frans2025splus}.

These engineering advances collectively demonstrate that second-order optimization is no longer merely a theoretical curiosity but a practical alternative for large-scale neural network training. The progress suggests that a mixture of theoretical advancements coupled with engineering tricks will narrow the gap between promised and empirical performances; in addition, hardware architectures evolve to better support the matrix operations fundamental to these methods. While this section was intentionally brief, there is still lots of work left to be done to truly realize the benefits of curvature-aware methods at scale.

\section{Theoretical Optimizer Design: All You Need is the Right Norm}\label{sec:modnorm}

While these methods seem mature and well-defined, our collective theoretical understanding of these algorithms is still a growing field of study. Specifically, it remains a mystery on how to choose the right algorithm for the model components and whether or not classic neural network transformations (e.g. normalization, residual connections, convolutions) affect this decision. Bernstein's modular norm framework reframes optimizer design as the problem of choosing appropriate geometric structure for each component of a neural network.

First, classical optimization theory reminds us that the gradient is not itself a vector in the primal parameter space, but rather a \emph{dual vector} that must be mapped back before being applied as an update. In conventional deep learning practice, this distinction is ignored: we simply subtract the raw gradient from the weights. \citet{bernstein2024normanthology} argue that such an approach amounts to a type error, since it fails to respect the geometry of the underlying loss surface.

Their proposal is to construct explicit \emph{duality maps} that translate gradients from the dual space into the appropriate update directions in the primal space. This view reframes the design of optimizers as the problem of choosing the right duality map for each layer, thereby tailoring the update rule to the local geometry induced by the architecture and its operator norms. From this perspective, optimizers become \emph{modular}: each layer is equipped with its own natural map, and the global optimization scheme is assembled from these layerwise components.  

The promise of this approach is two-fold. First, it offers a principled unifying framework for existing methods, many of which can be interpreted as partial approximations to particular duality maps. Second, it opens the door to a new generation of optimizers with formal guarantees on update scale and stability, designed to be fast and scalable while remaining faithful to the mathematics of dual spaces.

\subsection{The Steepest Descent Framework}

Before diving into specific optimizers, we will first re-establish the foundational framework of steepest descent through the lens of the modular norm literature \citep{bernstein2024normanthology}. First, steepest descent considers the problem of minimizing a local quadratic model:
\[
\arg\min_{\Delta w \in \mathbb{R}^n} \left[ g^\top \Delta w + \frac{\lambda}{2} \|\Delta w\|^2 \right]
\]
where $g$ is the gradient, $\lambda > 0$ is a sharpness parameter, and $\|\cdot\|$ is a chosen norm. This framework admits a dual perspective that separates step size from step direction:

\paragraph{Steepest Descent Decomposition.} The first proposition in \citet{bernstein2024normanthology}: For any gradient $g \in \mathbb{R}^n$, sharpness $\lambda$, and norm $\|\cdot\|$ with dual norm $\|\cdot\|_*$:
\[
\arg\min_{\Delta w \in \mathbb{R}^n} \left[ g^\top \Delta w + \frac{\lambda}{2} \|\Delta w\|^2 \right] 
= -\frac{\|g\|_*}{\lambda} \cdot \arg\max_{\|t\|=1} g^\top t
\]

This decomposition reveals two design choices:
\begin{enumerate}
    \item \textbf{Step size:} $\eta = \|g\|_* / \lambda$, determined by the dual norm of the gradient
    \item \textbf{Step direction:} The unit vector that maximizes inner product with the gradient
\end{enumerate}

Crucially, this is a \emph{first-order} method since the norm and sharpness are chosen a priori without computing (approximate) Hessians. The ``art'' lies in selecting norms that respect the geometry of the loss landscape. However, under certain norms, we observe that we can recover various learning algorithms that, among other interpretations, attempt to approximate a curvature matrix. This ``coincidence'' will be well-explored after setting up the context to interpret them.

\subsection{Dualized Updates}  

Bernstein emphasizes that gradients are naturally elements of the \emph{dual space} of the parameter space. Thus, the proposal is to always dualize gradients prior to updates, ensuring that updates respect the geometry of the loss function. Formally, if $g = \nabla \ell(\mathbf{w})$ lies in the dual space $W^*$, then the update applied to the primal weight space $W$ should be
\[
\mathbf{w}_{t+1} = \mathbf{w}_t - \eta \, \mathrm{dualize}(g)
\]

\paragraph{Dual Spaces and Basic Theorems.}  
Let $\mathcal{V}$ be a vector space with norm $\|\cdot\|$, and $\mathcal{V}^*$ its dual space consisting of linear functionals on $\mathcal{V}$. The associated dual norm is defined by
\[
\|g\|_* = \max_{\|v\| \leq 1} v^\top g
\]
Given a norm $\|\cdot\|$, the corresponding duality map is
\[
\mathrm{dualize}_{\|\cdot\|}(g) = \arg\max_{\|v\| \leq 1} v^\top g
\]

We now present some classical examples of its application. 

\paragraph{Example: Linear Model with Euclidean Norm.}  
For a linear model with parameters in $\mathbb{R}^d$ and the Euclidean norm $\|\cdot\|_2$, the dual norm is again $\|\cdot\|_2$. The duality map simply normalizes the gradient:
\[
\mathrm{dualize}_{2}(g) = \frac{g}{\|g\|_2}
\]
Thus, the update rule becomes
\[
\mathbf{w}_{t+1} = \mathbf{w}_t - \eta \frac{g}{\|g\|_2}
\]
which ensures scale-invariant updates across the parameter vector. Note that this essentially re-derives normalized gradient descent or the RMSProp optimizer in 1D without momentum of the squared gradient over iterations.

\paragraph{Example: Matrix Model with RMS$\to$RMS Induced Operator Norm.}
For a vector $v\in\mathbb{R}^d$, define the RMS norm $\|v\|_{\mathrm{RMS}}=\|v\|_2/\sqrt{d}$. 
For a matrix $W\in\mathbb{R}^{d_{\mathrm{out}}\times d_{\mathrm{in}}}$, the induced operator norm resolves to a rescaled spectral norm:
\[
\|W\|_{\mathrm{RMS}\to\mathrm{RMS}}
= \sqrt{\frac{d_{\mathrm{in}}}{d_{\mathrm{out}}}}\;\|W\|_*
\]
where $\|\cdot\|_*$ is the standard spectral norm ($\ell_2 \to \ell_2$ operator norm).  
Given a gradient $G\in\mathbb{R}^{d_{\mathrm{out}}\times d_{\mathrm{in}}}$ with reduced Singular Value Decomposition (SVD) $G=U\Sigma V^\top$, the corresponding duality map is
\[
\mathrm{dualize}_{\|\cdot\|_{\mathrm{RMS}\to\mathrm{RMS}}}(G)
= \sqrt{\frac{d_{\mathrm{out}}}{d_{\mathrm{in}}}}\; U V^\top
\]
and the layer update is
\[
\mathbf{w}_{t+1} \;=\; \mathbf{w}_t \;-\; \eta\, \mathrm{dualize}_{\|\cdot\|_{\mathrm{RMS}\to\mathrm{RMS}}}(G)
\]

This update performs a projection of the gradient matrix onto the space of semi-orthogonal matrices, scaled appropriately by dimension. As we will see, this directly connects to the Shampoo optimizer without preconditioner accumulation.

\subsection{Reinterpreting Classical Optimizers as Norm Choices}

We now demonstrate how popular optimizers can be understood as steepest descent under specific norm choices. This reframing reveals implicit design decisions and suggests paths for principled improvements.

\subsubsection{Adam: Steepest Descent under the Max-of-Max Norm}

The Adam optimizer, with exponential moving averages (EMA) disabled ($\beta_1 = \beta_2 = 0$), reduces to \emph{sign gradient descent}:
\[
\mathbf{w}_{t+1} = \mathbf{w}_t - \eta \cdot \mathrm{sign}(g_t)
\]

At first glance, this appears to be steepest descent under the vector $\ell_\infty$ norm on the flattened weight space. However, this interpretation discards the structured, layered nature of neural networks. The deeper insight is that the $\ell_\infty$ norm enjoys a special property: \emph{a max of a max is a max}.

For a neural network with weight matrices $W_1, \ldots, W_L$, consider the induced $\ell_1 \to \ell_\infty$ operator norm on each matrix. The max-of-max property states:
\[
\|w\|_\infty = \max_l \max_r \|\mathrm{row}_r(W_l)\|_\infty = \max_l \|W_l\|_{\ell_1 \to \ell_\infty}
\]

This reveals that layerwise sign descent actually solves a \emph{matrix-aware} steepest descent problem, as noted in Bernstein's work:

\begin{theorem}[Adam as Steepest Descent \citep{bernstein2024normanthology}]
For gradient matrices $G_1, \ldots, G_L$ and sharpness $\lambda > 0$:
\[
\arg\min_{\Delta W_1, \ldots, \Delta W_L} \left[ \sum_{l=1}^L \langle G_l, \Delta W_l \rangle + \frac{\lambda}{2} \max_{l=1}^L \|\Delta W_l\|^2_{\ell_1 \to \ell_\infty} \right]
\]
is solved by layerwise sign descent: $\Delta W_l = -\eta \cdot \mathrm{sign}(G_l)$ for all $l$, with step size $\eta = \frac{1}{\lambda} \sum_{l=1}^L \|G_l\|^*_{\ell_1 \to \ell_\infty}$.
\end{theorem}

\paragraph{Derivation of the Adam Update.}
The $\ell_1 \to \ell_\infty$ operator norm of a matrix is the maximum $\ell_\infty$ norm over its columns:
\[
\|W\|_{\ell_1 \to \ell_\infty} = \max_j \|\mathrm{col}_j(\mathbf{w})\|_\infty
\]
The dual norm is the sum of $\ell_1$ norms over columns:
\[
\|G\|^*_{\ell_1 \to \ell_\infty} = \sum_{ij} |G_{ij}| = \|G\|_1
\]
The duality map is:
\[
\mathrm{dualize}_{\ell_1 \to \ell_\infty}(G) = \arg\max_{\|T\|_{\ell_1 \to \ell_\infty} = 1} \langle G, T \rangle = \mathrm{sign}(G)
\]
since all entries must have unit magnitude and be gradient-aligned to maximize the inner product.

Sign descent performs implicit \emph{per-matrix gradient normalization}. Each layer receives an update whose magnitude is independent of the layer's gradient magnitude, since only the ``sign'' matters. This may explain why Adam and related sign-based methods outperform SGD in large language model training, where the size of gradient norms vary dramatically across layers.

\subsubsection{Shampoo: Steepest Descent under the Spectral Norm}

The Shampoo optimizer maintains left and right preconditioners for each layer:
\[
L_t = L_{t-1} + G_t G_t^\top, \quad R_t = R_{t-1} + G_t^\top G_t
\]
\[
\mathbf{w}_{t+1} = \mathbf{w}_t - \eta \cdot L_t^{-1/4} G_t R_t^{-1/4}
\]

When accumulation is disabled ($L_t = G_t G_t^\top$, $R_t = G_t^\top G_t$), Shampoo reduces to:
\[
\mathbf{w}_{t+1} = \mathbf{w}_t - \eta \cdot (G_t G_t^\top)^{-1/4} G_t (G_t^\top G_t)^{-1/4}
\]

Substituting the reduced SVD $G_t = U_t \Sigma_t V_t^\top$, this simplifies to:
\[
\mathbf{w}_{t+1} = \mathbf{w}_t - \eta \cdot U_t V_t^\top
\]

This is the projection of the gradient onto the space of semi-orthogonal matrices, which are matrices that satisfy $AA^\top = I$ or $A^\top A = I$.

\begin{theorem}[Shampoo as Steepest Descent \citep{bernstein2024normanthology}]
For gradient matrices $G_1, \ldots, G_L$ with reduced SVDs $G_l = U_l \Sigma_l V_l^\top$:
\[
\arg\min_{\Delta W_1, \ldots, \Delta W_L} \left[ \sum_{l=1}^L \langle G_l, \Delta W_l \rangle + \frac{\lambda}{2} \max_{l=1}^L \|\Delta W_l\|^2_{\ell_2 \to \ell_2} \right]
\]
is solved by $\Delta W_l = -\eta \cdot U_l V_l^\top$ with step size $\eta = \frac{1}{\lambda} \sum_{l=1}^L \mathrm{tr}(\Sigma_l)$.
\end{theorem}

\paragraph{Derivation of the Shampoo Update.}
The spectral norm (or $\ell_2 \to \ell_2$ operator norm) is the largest singular value:
\[
\|W\|_{\ell_2 \to \ell_2} = \sigma_{\max}(\mathbf{w})
\]
Its dual norm is the nuclear norm (sum of singular values):
\[
\|G\|^*_{\ell_2 \to \ell_2} = \sum_i \sigma_i(G) = \mathrm{tr}(\Sigma)
\]
The duality map finds the unit spectral norm matrix maximizing the Frobenius inner product:
\[
\mathrm{dualize}_{\ell_2 \to \ell_2}(G) = \arg\max_{\|T\|_{\ell_2 \to \ell_2} = 1} \langle G, T \rangle = U V^\top
\]
where $G = U \Sigma V^\top$ is the reduced SVD. This follows because:
\[
\langle G, T \rangle = \mathrm{tr}(G^\top T) = \mathrm{tr}(\Sigma V^\top U^\top T) = \sum_i \sigma_i \cdot u_i^\top T v_i \leq \sum_i \sigma_i
\]
which holds with equality when $T v_i = u_i$ for all $i$, or more simply, $T = U V^\top$.

\paragraph{Geometric Interpretation.}
The update $U V^\top$ is the \emph{closest semi-orthogonal matrix} to the gradient $G$ in Frobenius norm. Semi-orthogonal matrices have all singular values equal to 1, so this update ``normalizes away'' the gradient's singular value spectrum while preserving its left and right singular spaces. The spectral norm emerges naturally from loss landscape analysis. For a linear predictor with square loss, one can show:
\[
\mathcal{L}(W + \Delta \mathbf{w}) \leq \mathcal{L}(\mathbf{w}) + \langle \nabla \mathcal{L}(\mathbf{w}), \Delta W \rangle + \frac{1}{2} \cdot \frac{d_{\mathrm{in}}}{d_{\mathrm{out}}} \|\Delta W\|^2_{\ell_2 \to \ell_2}
\]
Minimizing this upper bound is precisely steepest descent under the spectral norm. This is an instance of \emph{majorization-minimization}, which is a principled, first-order design pattern requiring no Hessian approximations \citep{carlson2015stochastic}. Shampoo's success in the AlgoPerf competition suggests that this spectral norm captures important geometric structure in deep learning problem spaces \citep{dahl2023algoperf}.

\subsubsection{Prodigy: Automatic Step Size via Escape Velocity}

 While our current discussion of optimizers has largely focused on improving the descent direction, we remind the reader that the step size, or the learning rate, also plays a critical role in determining the next jump in parameter space. While we plan to devote a longer discussion to the role of learning rates in optimizer design, we first include an optimizer that aligns with principles from the modular norm theory and aims to reduce total hyperparameters needed to tune. 
 
 The Prodigy optimizer combines sign descent with an adaptive step size mechanism \citep{mishchenko2023prodigy}. With EMA disabled ($\beta_1 = \beta_2 = 0$), Prodigy simplifies to:
\[
\eta_{t+1} = \max\left(\eta_t, \frac{g_t^\top (w_0 - \mathbf{w}_t)}{\|g_t\|_1}\right)
\]
\[
\mathbf{w}_{t+1} = \mathbf{w}_t - \eta_t \cdot \mathrm{sign}(g_t)
\]

Prodigy uses the same norm as Adam (max-of-max / $\ell_\infty$) but automatically warms up the step size using a heuristic that Bernstein terms as \emph{escape velocity}. To be precise, we rewrite the step size update as:
\[
\eta_{t+1} = \max\left(\eta_t, \frac{\|g_t\|_2}{\|g_t\|_1} \cdot \|\mathbf{w}_t - w_0\|_2 \cdot \cos\theta\right)
\]
where $\theta$ is the angle between $g_t$ and $(w_0 - \mathbf{w}_t)$. If we make two assumptions:
\begin{enumerate}
    \item Gradients are dense: $\|g_t\|_2 / \|g_t\|_1 \approx 1/\sqrt{n}$
    \item Still in linear regime: $\cos\theta \approx 1$
\end{enumerate}
then this becomes $\eta_{t+1} \approx \max(\eta_t, \|\mathbf{w}_t - w_0\|_{\mathrm{RMS}})$.

The algorithm begins with a tiny initial step size $\eta_0 \ll \eta^*$ (far below the optimal value). At each iteration, it checks whether the weights have ``escaped'' the initial linearization of the loss around $w_0$. If the directional derivative $(\mathbf{w}_t - w_0)^\top g_t$ is still negative, the weights could have moved further, so the step size increases. Once the weights escape (directional derivative is near zero), the step size stops growing. Another way to interpret this is as an online line search problem \citep{bernstein2024normanthology}.

Prodigy demonstrates the modularity of this descent framework. While the max-of-max norm is chosen, in theory one could choose any norm and plug into the Prodigy optimizer. Moreover, the base step size can be determined through various means: fixed hyperparameter, line search, or heuristics, but the power comes in being able to automatically adapt the learning rate over time. Prodigy's automatic warm-up removes one degree of tuning freedom while maintaining the geometric benefits of sign descent. The idea of tuning the learning rate along with the optimizer will be more closely explored in Section \ref{sec:lrs}.

\subsection{The Modular Norm Framework}

As one may have noted by now, Bernstein takes a first-principles approach on deep learning optimization: each layer in a neural network may admit its own natural duality map, determined by its operator norm and semantics. Thus, one can build up the network based on their desired features and automatically generate a recipe for training the network. This observation generalizes via the \emph{modular norm}:

\begin{definition}[Modular Norm \citep{bernstein2024normanthology}]
Given scalar coefficients $s_1, \ldots, s_L > 0$ and norms $\|\cdot\|_1, \ldots, \|\cdot\|_L$, the modular norm is:
\[
\|(W_1, \ldots, W_L)\| = \max\{s_1 \|W_1\|_1, \ldots, s_L \|W_L\|_L\}
\]
\end{definition}

\begin{theorem}[Steepest Descent under the Modular Norm \citep{bernstein2024normanthology}]
The steepest descent problem under the modular norm:
\[
\arg\min_{\Delta W_1, \ldots, \Delta W_L} \left[ \sum_{l=1}^L \langle G_l, \Delta W_l \rangle + \frac{\lambda}{2} \max_{l=1}^L s_l^2 \|\Delta W_l\|_l^2 \right]
\]
is solved by:
\[
\Delta W_l = -\frac{\eta}{s_l} \cdot \arg\max_{\|T_l\|_l = 1} \langle G_l, T_l \rangle
\]
with global step size $\eta = \frac{1}{\lambda} \sum_{k=1}^L \frac{1}{s_k} \|G_k\|^*_k$.
\end{theorem}

This framework separates three design choices:
\begin{enumerate}
    \item \textbf{Per-layer norms} $\|\cdot\|_l$: capture the geometric role of each layer
    \item \textbf{Scalar weights} $s_l$: balance update magnitudes across layers
    \item \textbf{Global step size} $\eta$: determined by dual norms or learned adaptively
\end{enumerate}

\subsection{Principled Norm Choices for Common Layer Types}

The modular framework demands that we assign norms based on the \emph{role} each layer plays, not merely its parameter shape. We now derive appropriate norms for common layer types.

\subsubsection{Linear Layers}

Linear layers map vectors to vectors, typically with roughly unit RMS norm on inputs and outputs (especially under careful initialization schemes like Kaiming or Xavier). The natural choice is the RMS$\to$RMS operator norm:
\[
\|W\|_{\mathrm{RMS} \to \mathrm{RMS}} = \sqrt{\frac{d_{\mathrm{in}}}{d_{\mathrm{out}}}} \|W\|_{\ell_2 \to \ell_2}
\]

As shown earlier, the corresponding duality map is:
\[
\mathrm{dualize}_{\mathrm{RMS} \to \mathrm{RMS}}(G) = \sqrt{\frac{d_{\mathrm{out}}}{d_{\mathrm{in}}}} U V^\top
\]
This provides dimension-aware normalization while respecting the matrix structure.





\subsubsection{Convolutional Layers}

Convolutional layers can be viewed as structured linear operators. For a conv layer with kernel $K \in \mathbb{R}^{c_{\mathrm{out}} \times c_{\mathrm{in}} \times k_h \times k_w}$, one approach is to reshape into a matrix and apply the RMS$\to$RMS norm:
\[
\|K\|_{\mathrm{conv}} = \left\|\mathrm{reshape}(K) \in \mathbb{R}^{c_{\mathrm{out}} \times (c_{\mathrm{in}} k_h k_\mathbf{w})}\right\|_{\mathrm{RMS} \to \mathrm{RMS}}
\]

Alternatively, one can use the induced norm from spatial input tensors to output tensors, accounting for the circulant structure. The appropriate choice may depend on whether the convolution preserves or changes spatial resolution. This idea is an active area of research and could definitely be further explored.

\subsection{The Modular Optimizer Conjecture}

We now articulate the central hypothesis driving this line of work. A deep neural network trained with per-layer optimizers, where each layer's optimizer is derived from steepest descent under a norm chosen to match that layer's geometric role, will exhibit:
\begin{enumerate}
    \item \textbf{Learning rate transfer across scales}: Optimal step sizes remain stable as model size increases
    \item \textbf{Improved training stability}: Update magnitudes adapt naturally to heterogeneous curvature
    \item \textbf{Faster convergence}: Respecting layer-specific geometry reduces wasted gradient information
    \item \textbf{Reduced hyperparameter sensitivity}: Principled norms provide better default behavior
\end{enumerate}

The conjecture essentially conveys that a principled  ``toolkit'' of modular dualizations could provide many benefits that are critical to efficient and scalable neural network training. In this view, the next generation of optimizers would not be ``monolithic'' but \emph{modular}, assembling the right optimizer for each layer to ensure robust downstream performance for a diverse array of neural network architectures.

Early empirical evidence supports this conjecture. \citet{yang2023spectral} demonstrate learning rate transfer using spectral normalization in large vision models. The Shampoo variant that won the AlgoPerf external tuning track leverages spectral geometry. Muon, which can also be interpreted as a modular optimizer as it combines sign descent for embeddings with orthogonalized updates for linear layers, achieves competitive performance on language modeling benchmarks \citep{jordan2024muon}.

\subsection{Practical Implementation: The \texttt{modula} Library}

To facilitate experimentation with modular optimizers, Large, Bernstein and collaborators have developed the \texttt{modula} library, which provides a practical toolkit for implementing norm-based optimization methods \citep{large2024modular,modula-docs}. The library's design philosophy centers on making theoretically-principled optimizer design accessible to practitioners while maintaining computational efficiency.

The library provides implementations of layer-specific duality maps corresponding to different induced operator norms. For linear layers, it implements the RMS$\to$RMS operator norm (equivalent to rescaled spectral norm), which as discussed in Section \ref{alg:shampoo} corresponds to projecting gradients onto semi-orthogonal matrices. For embedding layers, it provides the $\ell_1 \to \ell_2$ operator norm appropriate for mapping one-hot vectors to dense representations. The library also supports custom norm definitions, allowing researchers to experiment with novel geometric structures.

A key practical contribution is the implementation of efficient algorithms for computing these duality maps. For the spectral norm case, the library provides Newton-Schulz iterations (as discussed in the Muon algorithm) as well as randomized SVD approximations. These methods avoid the computational expense of full eigendecomposition while maintaining the essential geometric properties of the updates. The library's modular architecture allows practitioners to mix and match different optimizers for different layer types within a single model, directly implementing the modular norm framework proposed in the theoretical work.

Beyond computational tools, the \texttt{modula} library has enabled new empirical insights into the behavior of norm-based optimization. In a striking demonstration of the qualitative difference between standard and dualized gradient descent, Bernstein replicates the ``weight erasure'' experiment of \citet{jesus2020effect}. The original experiment showed that when training wide neural networks with standard gradient descent, a watermark image embedded in the initial weights remains visible after training, suggesting that weights barely move from initialization. However, when the same experiment is conducted with dualized gradient descent --- that is, applying the duality map corresponding to the spectral norm --- the watermark is completely erased during training.

This difference can be understood through the stable rank of weight updates, defined as $\text{srank}(M) = \|M\|_F^2 / \|M\|_2^2$. Raw gradients typically exhibit very low stable rank, meaning their singular values are highly imbalanced with most signal concentrated in a few directions. The duality map for spectral norm sets all non-zero singular values to unity, dramatically increasing the stable rank to equal the matrix rank. Consequently, dualized updates have substantially larger typical entry magnitudes for the same spectral norm: $|M_{ij}|_{\text{typ}} \sim \|M\|_2 / \sqrt{\text{srank}(M)}$. Experiments in the \texttt{modula} documentation show that gradient stable rank remains small across network widths, while dualized gradient stable rank grows with width until plateauing at the batch size, which serves as an upper bound of the rank \citep{modula-docs}.

This observation connects back to the central narrative of norm-based optimization. The choice of norm not only determines the geometric direction of steepest descent but also fundamentally alters the numerical trajectory of training. Networks optimized under the spectral norm explore parameter space in a qualitatively different manner than those trained with standard gradient descent, moving ``further'' from initialization in the sense of typical entry-wise displacement. Whether this property contributes to the empirical success of methods like Shampoo and Muon remains an open question, but the weight erasure experiments demonstrate that norm choice has consequences extending beyond convergence rates to the basic numerical behavior of neural networks.

\paragraph{Expressivity of Optimizer Solutions.} The optimizer is not merely a vehicle for faster descent but a mechanism that shapes which solutions are reachable during training. In neural networks, different learning rules trace different paths through parameter space and converge to minima with meaningfully different properties. Consequently, studies have shown that the choice of optimizer encodes inductive biases and changes the effective expressivity of a fixed architecture, restricting the set of functions that are practically reachable from initialization \citep{pascanu2025optimizers}. On this view, optimizer design should be treated as a first-class lever, alongside architecture and data, for inducing target properties of the learned solution. Rather than assessing methods solely by convergence speed, the community should systematically study the biases of existing algorithms and deliberately construct new optimizers that steer training toward solutions with desired characteristics such as sparsity, structured representations, and robustness to interference.

Preconditioning and duality maps makes these ideas concrete by altering how gradients are transformed before updates, thereby modifying the quality and nature of solutions that training can discover. Pascanu et. al showed that non-diagonal preconditioners that capture cross-parameter correlations may reduce interference, improve forward transfer, and mitigate catastrophic forgetting relative to diagonal or first-order methods trained from the same initialization, illustrating a qualitative shift in the learned representations rather than a mere change in speed \citep{pascanu2025optimizers}. A second case study in the paper demonstrates that we can induce other desirable properties, such as sparsity, directly via the optimizer at the cost of slower convergence. This furthers the central claim that preconditioners are levers for injecting inductive bias and shaping solution properties beyond standard generalization metrics.

\subsection{Limitations of the Modular Norm Framework}

While the modular norm framework provides elegant theoretical insights and practical algorithmic innovations, several fundamental limitations and open questions warrant careful discussion. These limitations reveal important gaps between the idealized theoretical framework and the messy reality of training modern deep networks.

\subsubsection{Missing Theory on EMA}

Perhaps the most significant theoretical gap in the modular norm framework concerns the role of exponential moving averages (EMA). The framework's analysis focuses on instantaneous norms applied to current gradients, characterizing steepest descent updates in terms of well-defined geometric quantities. However, all practical implementations of the optimizers derived from this framework --- Adam, Shampoo, SOAP, and variants --- rely heavily on EMAs for stability and convergence. The theory largely ignores this critical component.

Consider Adam: the modular norm theory establishes that sign descent (Adam with $\beta_1 = \beta_2 = 0$) implements steepest descent under the max-of-max ($\ell_1 \to \ell_\infty$) operator norm \citep{bernstein2024normanthology}. This geometric interpretation is intellectually satisfying and mathematically rigorous. However, practitioners never use Adam without momentum. The $\beta_1$ parameter (typically 0.9) maintains an EMA of the gradient or first moment, while $\beta_2$ (typically 0.999) maintains an EMA of the squared gradient, or the second moment of the gradient \citep{kingma2015adam}. These moving averages fundamentally transform the optimizer's behavior, yet their role in the geometric framework remains unclear.

The question is not merely one of variance reduction. Recent theoretical work by \citet{ahn2024adamema} demonstrates that Adam with model EMA achieves optimal convergence rates in nonconvex settings, and that the analysis ``crucially relies'' on the momentum and discounting factors. This suggests EMA plays a deeper role than simple noise smoothing --- it may fundamentally alter the effective geometry of the optimization problem. The modular norm framework does not yet provide tools to reason about this phenomenon.

Bernstein and Newhouse acknowledge this gap in their ``Norm Anthology'' paper, noting that understanding EMA's role in the framework is ``perhaps still an open problem'' \citep{bernstein2024normanthology}. Without resolution, the framework provides limited guidance for hyperparameter selection: Should $\beta_2$ be larger or smaller for spectral norm optimizers versus infinity norm optimizers? How should EMA parameters scale with network depth or batch size? These remain empirical questions despite the framework's theoretical rigor.

The interaction between EMA and duality maps presents additional complications. When we compute the spectral duality map $UV^\top$, are we computing it for the instantaneous gradient $G_t$ or for an EMA-smoothed version? If the latter, does the smoothing respect the geometric structure imposed by the norm? The Muon optimizer uses Newton-Schulz iterations to approximate the duality map, but applies these to momentum-accumulated gradients without theoretical justification for this composition of operations.

\subsubsection{Intractability for Complex Architectures}

Modern neural networks exhibit architectural complexity that strains the modular norm framework. While the theory elegantly handles networks composed of simple sequential layers (Linear $\to$ ReLU $\to$ Linear), contemporary architectures like Transformers interleave multiple computational paths, normalization schemes, and skip connections in ways that resist clean geometric characterization.

Consider the standard Transformer encoder layer:
\begin{align*}
&x \to \text{LayerNorm} \to \text{MultiHeadAttention} \to \text{Add}(x, \cdot) \to \\
&\quad\quad\text{LayerNorm} \to \text{FeedForward} \to \text{Add}(x, \cdot) \to \text{output}
\end{align*}

Each component introduces geometric structure: \textit{LayerNorm} projects activations onto a $(d-1)$-dimensional sphere, fundamentally changing the geometry mid-layer. \textit{MultiHeadAttention} involves four separate linear projections (Query, Key, Value, Output) with interdependent forward-pass semantics but independent gradient computations. \textit{Residual connections} create multiple gradient pathways, mixing contributions from different geometric spaces. Deriving the ``correct'' induced norm for compositions that is evidently challenging. The modular norm framework provides a recursive construction for sequential architectures for basic attention operators, but it does not work for all complicated Transformer architectures \citep{large2024modular}. When gradients from the attention sublayer and the residual pathway merge, they have passed through different geometric transformations. Should we dualize them separately before addition, or add them and then dualize?

The specific case of Query-Key-Value projections in attention mechanisms exemplifies the ambiguity. These three matrices share input features but serve distinct semantic roles: Query and Key engage in similarity computation via dot product, while Value carries information to be aggregated. A principled norm-based approach might assign different norms to these three matrices based on their roles. However, in practice they typically share the same embedding dimension, and existing implementations treat them uniformly. Is this optimal, or does it reflect a limitation of current theory?

Layer normalization parameters present another puzzle. LayerNorm involves scale ($\gamma$) and shift ($\beta$) parameters that operate elementwise on normalized activations. These parameters have unusual geometry: they modulate already-normalized quantities and interact multiplicatively with the forward pass. The appropriate induced norm for such parameters remains unclear. Should we treat them as scalar multipliers and use $\ell_\infty$ norm? Or does their post-normalization role suggest a different structure?

The modular norm framework shows greatest success on architectures with clear hierarchical structure: CNNs, MLPs, and simple RNNs. For these, the recursive norm construction proceeds naturally. But for modern architectures with complex connectivity patterns --- Transformers with cross-attention, graph neural networks, even mixture-of-experts models --- the framework provides less guidance. This suggests the need for either extensions to the theoretical framework that handle non-hierarchical computation graphs or acceptance that some architectural components may not admit principled geometric characterization.

\subsubsection{Interaction with Other Training Techniques}

Modern deep learning practitioners employ a sophisticated ensemble of training techniques beyond the base optimizer. The modular norm framework, by focusing narrowly on the geometric structure of gradient descent, provides limited guidance on how norm-based optimizers interact with these complementary methods. In some cases, interactions may be beneficial; in others, they may undermine the framework's theoretical guarantees.

\paragraph{Gradient clipping} represents the most direct interaction. Many training pipelines clip gradients by global norm before applying optimizer updates. When combined with modular norm optimizers, this creates a conceptual tension. The modular framework carefully constructs update directions based on the geometry of specific norms (spectral, infinity, etc.). Global gradient clipping then imposes a different geometric constraint --- an $\ell_2$ ball in the concatenated parameter space --- that cuts across the modular structure. Does clipping before dualization negate the benefits of geometric updates? Should we clip after dualization instead? Or does clipping address orthogonal concerns (training stability) that don't interfere with geometric principles?

\paragraph{Learning rate schedules} complicate the picture further. The modular norm framework derives natural step sizes from dual norms, suggesting that geometric structure should inform learning rate selection. However, practitioners universally employ schedules --- warmup, cosine decay, linear decay --- that override these theoretical step sizes. Prodigy attempts to address this by automatically warming up learning rates using ``escape velocity'' heuristics, but this still represents an incomplete solution \citep{mishchenko2023prodigy}.

The interaction between schedules and geometric updates raises questions: If we've chosen norms to respect the loss landscape's geometry, why do we need warmup? If warmup is necessary (perhaps due to batch statistics instability in early training), does it indicate that our norm choice is wrong? Should schedule-free optimization be the goal, or are schedules addressing fundamentally different concerns (e.g., exploration vs. exploitation tradeoffs)? Schedules deserve their own analysis, which follows in Section \ref{sec:lrs}. 

\paragraph{Mixed-precision training} adds another layer of complexity. Training in FP16 or BF16 requires loss scaling to prevent gradient underflow. When gradients are scaled before dualization, does this alter the geometric structure? For spectral norm updates involving SVD, numerical precision matters: small singular values might be lost to quantization, changing the effective rank of duality maps. The Modula documentation notes that Newton-Schulz iterations (used for efficient spectral duality) can suffer from numerical instability in low precision. However, systematic guidance on precision choices for different norms is absent.

\subsubsection{Final Remarks}

These limitations reveal a fundamental tension: the modular norm framework aspires to provide a complete, principled theory of neural network optimization, yet modern deep learning consists of a complex interplay of techniques --- architectural innovations, regularization schemes, training tricks --- that evolved semi-independently. Integrating norm-based optimization theory with this ecosystem requires either extending the theory to encompass all techniques within a unified geometric framework or developing clear interfaces that delineate where geometric principles apply and where empirical practices and expert advice is followed.

\section{Modern Training Techniques Beyond the Optimizer}\label{sec:toolkitdl}

So far, we have explored the vast space of optimization algorithms for deep learning, focusing solely on the scheme by which we update our weights in every training iteration. We saw from first principles how these optimizers have been theoretically motivated by the steepest descent problem under various norms and how to preserve the correct geometry of the gradient space via duality maps. However, learning in neural networks can be hardly be attributed solely to the optimizer itself; there are various ``tools'' that we employ alongside the optimizer to ensure that we achieve our training goals, whether that be generalization, hyperparameter transfer, low-norm solutions, etc. This section is dedicated to exploring the ``toolkit'' used in practice to scale up neural network training. We will focus on the interactions between these tools and the optimizer as well as how they can be effectively understood in today's deep learning regime. 

\subsection{Maximal Update Parameterization ($\mu$P)}

The Maximal Update Parameterization ($\mu$P) represents a principled reparameterization framework for neural networks that establishes specific scaling relationships between network architecture and optimization hyperparameters \citep{yang2021tensorIV,yang2022tensorprogramsV}. Unlike standard parameterization (SP), which suffers from scale-dependent dynamics as network width grows, $\mu$P prescribes width-dependent scaling rules for weight initialization and learning rates that stabilize training dynamics across model scales. This theoretical framework has profound implications for understanding the interplay between network parameterization and optimization algorithms.

\subsubsection{Theoretical Framework and Feature Learning}

From a theoretical perspective, $\mu$P emerges as the unique parameterization that achieves \emph{maximal feature learning} in the infinite-width limit \citep{yang2022tensorprogramsV}. This distinguishes it fundamentally from both standard parameterization --- where network components either diverge or cease learning --- and from Neural Tangent Kernel (NTK) parameterization, which exhibits only kernel-like behavior with frozen features. The mathematical foundation rests on balancing two competing width-dependent effects in neural network training:
\begin{enumerate}
    \item \textbf{Hidden layer gradient dilution}: Each neuron's gradient contribution scales as $\mathcal{O}(1/w)$ as width $w$ increases
    \item \textbf{Output layer gradient aggregation}: The output layer accumulates contributions from $w$ neurons, yielding gradients that scale as $\mathcal{O}(w)$
\end{enumerate}

$\mu$P's scaling rules are specifically designed to counteract these effects, ensuring that parameter update magnitudes remain at $\Theta(1)$ regardless of width. This is formalized through an parametrization framework that specifies initialization variances and learning rate scalings as functions of network width (as measured by the size of corresponding weight matrices).

\subsubsection{Hyperparameter Transfer and $\mu$Transfer}

A central practical consequence of $\mu$P's width-invariant dynamics is that optimal hyperparameters remain stable across model scales, enabling the \emph{$\mu$Transfer} paradigm. Empirically, this allows practitioners to tune hyperparameters on small proxy models and transfer them to substantially larger architectures without retuning. The original work demonstrates this capability through experiments showing successful transfer from 13M to 350M parameters (BERT-large) and from 40M to 6.7B parameters (GPT-3), with the latter requiring only 7\% of the full pretraining cost for hyperparameter optimization.

The set of $\mu$Transfer-able hyperparameters encompasses learning rates (both global and layer-specific), initialization scales, optimizer momentum coefficients, and regularization parameters including dropout and weight decay. Notably, architectural choices, batch size, and training duration are not guaranteed to be $\mu$Transfer-able, requiring separate consideration during scaling.

\subsubsection{Optimizer-Dependent Parameterization}

A critical but often underappreciated aspect of $\mu$P is its optimizer dependence: the appropriate parameterization scheme varies depending on the optimization algorithm employed. This was rigorously demonstrated by \citet{ishikawa2024local}, who analyzed $\mu$P in the context of local learning algorithms and revealed that different optimization methods induce distinct scaling requirements.

For first-order methods, the situation is relatively well understood. SGD(+M) follows the canonical $\mu$P scaling derived in the original $\mu$P paper. However, adaptive gradient methods such as Adam introduce additional complexity: the entry-wise normalization in Adam's update rule necessitates modified scaling relationships to maintain stable dynamics across widths. In Tensor Programs V, \citet{yang2022tensorprogramsV} extended the $\mu$P framework to account for Adam's adaptive preconditioning, ensuring that $\mu$Transfer holds for this widely-used optimizer.

The interaction between $\mu$P and second-order optimization methods represents a particularly rich area of ongoing research. \citet{ishikawa2024local} derived $\mu$P-compatible parameterizations for two prominent higher-order optimizers: KFAC and Shampoo. Their analysis reveals that proper scaling of not only learning rates and initialization but also damping terms --- which regularize the curvature matrix inversion --- is essential for stable feature learning at large widths.

More fundamentally, \citet{ishikawa2024local} discovered a striking phenomenon in their analysis of predictive coding networks: depending on the parameterization scheme, the effective gradient can interpolate between pure first-order gradients and Gauss-Newton-like gradients. Specifically, in deep linear networks trained with predictive coding under different parameterizations, the effective update incorporates varying degrees of second-order curvature information. Under standard parameterization, predictive coding reduces to first-order gradient descent, while under $\mu$P with appropriately scaled inference steps, it approximates Gauss-Newton updates that precondition gradients with layer-wise curvature.

This interpolation behavior suggests a deeper principle: parameterization can fundamentally alter the optimization geometry. The choice of how to scale learning rates and initialization with width determines not merely the step size but the direction and curvature-awareness of parameter updates. This finding challenges the conventional view of parameterization as a purely practical concern (motivating many deep learning teams to just blindly sweep over large hyperparameter sets) and elevates it to a fundamental component of optimization algorithm design.

\subsubsection{Local Learning and Alternative Training Paradigms}

The extension of $\mu$P to local learning algorithms demonstrates its generality beyond standard backpropagation \citep{ishikawa2024local}. Local learning methods, which train networks through layer-wise local losses rather than global backpropagation, have historically suffered from instability and hyperparameter sensitivity due to the additional degrees of freedom introduced by locality. The application of $\mu$P to these algorithms stabilizes training across widths and enables hyperparameter transfer, suggesting that width-stable parameterization is a universal principle applicable to diverse training methodologies.

Notably, the analysis reveals unique properties in local learning under $\mu$P that are absent in conventional backpropagation. The local loss structure interacts with parameterization to produce optimization dynamics with qualitatively different characteristics, including the aforementioned gradient interpolation between first-order and Gauss-Newton regimes.

\paragraph{Unit-Scaled $\mu$P.}

The recently proposed unit-scaled $\mu$P (u-$\mu$P) extends the framework by incorporating Unit Scaling principles, which enforce unit variance for all activations, weights, and gradients at initialization \citep{blake2024uMuP}. This synthesis addresses several practical limitations:
\begin{itemize}
    \item \textbf{Simplified hyperparameter space}: u-$\mu$P eliminates the need for user-specified base shapes and reduces interdependencies among hyperparameters
    \item \textbf{Near-optimal defaults}: The unit-scale principle provides a natural anchor for default hyperparameter values
    \item \textbf{Numerical stability}: Unit-scale initialization improves conditioning for low-precision training (e.g., FP8)
    \item \textbf{Improved width scaling}: Corrections to learning rate scaling yield better performance as width increases
\end{itemize}

From a theoretical standpoint, u-$\mu$P can be viewed as adding an additional constraint (unit-scale dynamics) to the $\mu$P optimization problem, further restricting the space of viable parameterizations while providing stronger guarantees about numerical behavior.

\subsubsection{Open Questions and Future Directions}

Despite substantial progress, several fundamental questions remain open regarding the interaction between $\mu P$ and optimizers:

\begin{enumerate}
    \item \textbf{Higher-order methods}: While extensions to K-FAC and Shampoo exist, the full space of quasi-Newton and natural gradient methods remains incompletely characterized. How should parameterization scale for methods that explicitly maintain curvature approximations?
    
    \item \textbf{Discrete optimization}: $\mu$P theory is derived from continuous-time analysis of gradient flow. How do discrete optimization effects (e.g., learning rate schedules, gradient clipping) interact with width scaling?
    
    \item \textbf{Non-width scaling dimensions}: Most $\mu$P theory focuses on width scaling. Recent work has begun exploring depth scaling and other architectural dimensions, but a unified theory remains elusive.
    
    \item \textbf{Beyond feature learning}: The maximal feature learning criterion that motivates $\mu$P is one possible objective. Are there alternative scaling criteria that optimize for other desirable properties (e.g., generalization, sample efficiency)?
    
    \item \textbf{Practical deviations}: Empirical studies have identified cases where $\mu$Transfer breaks down in modern training setups (e.g., learning rate warmup and decay, specific normalization schemes). Understanding and resolving these deviations remains an active research area.
\end{enumerate}

From a practitioner's perspective, $\mu$P should be understood not as a fixed prescription but as a design principle: proper alignment of parameterization with optimization algorithm and network architecture enables stable, predictable scaling behavior. As the field develops increasingly sophisticated optimization methods and explores alternative training paradigms, the principles underlying $\mu$P --- width-invariant dynamics, balanced updates, and optimizer-aware scaling --- will likely remain central to scalable deep learning.

\subsection{Learning Rate Schedules}\label{sec:lrs}

A central observation in modern optimization is that the learning rate schedule $\{\eta_t\}_{t=1}^T$ defines a ``macroscopic control law'' that shapes both convergence speed and generalization. While optimizers such as SGD or Adam control \emph{local geometry} (through momentum or adaptive scaling), the schedule defines the \emph{temporal geometry} of training, determining how aggressively we explore and how gradually we converge.

\subsubsection{Historical Context: From Convex Optimization to Deep Learning}

The theory of learning rate selection has deep roots in convex optimization, where classical methods provide strong convergence guarantees. In the convex setting, line search methods --- such as backtracking with Armijo or Wolfe conditions --- adaptively select step sizes to ensure sufficient decrease of the objective function \citep{nocedal2006numerical,armijo1966minimization}. These methods evaluate the objective or gradient at multiple candidate step sizes per iteration, bracketing an interval that satisfies descent criteria. For smooth strongly convex functions with condition number $\kappa$, gradient descent with exact line search achieves linear convergence $\mathcal{O}((1 - 1/\kappa)^t)$, and Newton's method with appropriate line search achieves quadratic convergence near the optimum.

However, this theoretical framework encounters fundamental obstacles when applied to modern deep learning:

\paragraph{Computational Intractability.} Line search requires multiple forward passes (and possibly backward passes) per iteration to evaluate candidate step sizes. For large-scale neural networks trained on massive datasets, this overhead is prohibitive. A single forward-backward pass on a modern language model may take minutes to hours; repeating this multiple times per update is infeasible.

\paragraph{Stochasticity and Mini-Batching.} Deep learning relies on stochastic gradients computed from mini-batches, introducing noise that violates the assumptions underlying classical line search convergence proofs. The objective function evaluated on different mini-batches differs, making it unclear how to apply Armijo-type sufficient decrease conditions meaningfully.

\paragraph{Non-Convexity and Saddle Points.} Neural network loss landscapes are highly non-convex with abundant saddle points and local minima. Classical convex theory provides no guidance on learning rate schedules in such settings, and empirical practice has diverged sharply from theoretical prescriptions.

\paragraph{The Gap Between Theory and Last-Iterate Behavior.} Classical convergence analysis typically bounds the average iterate $\bar{x} = \frac{1}{T}\sum_{t=1}^T x_t$ or the best iterate $\min_{t \le T} f(x_t)$, but practitioners care about the \emph{last iterate} $x_T$ returned by the algorithm. \citet{pan2021eigencurve} demonstrate that even for simple quadratic objectives, optimizing upper bounds on average-iterate error can yield suboptimal last-iterate performance. They propose Eigencurve, a learning rate schedule that achieves optimal convergence rates for the last iterate on quadratic problems when the Hessian's eigenvalue distribution is skewed. However, this method requires knowledge of the full eigenspectrum of the Hessian, making it impractical for high-dimensional neural networks where computing or even approximating the Hessian is computationally prohibitive. This highlights a fundamental tension: theoretically optimal schedules often demand problem-specific information (curvature, condition numbers, eigenspectra) that is inaccessible in practice.

These challenges have motivated the development of fixed, problem-agnostic schedules that work well empirically across diverse architectures and tasks, even without curvature information.

\paragraph{General Schedule Families.}
We denote $t$ as the training step and $T$ as the total number of updates. Three common functional forms appear in large-scale deep learning:

\[
\eta_t = 
\begin{cases}
\eta_{\max} \frac{t}{T_w}, & t \le T_w \quad \text{(warmup)}\\
\eta_{\max}, & T_w < t \le T_s \quad \text{(stable phase)}\\
\eta_{\max} \phi\!\left(\frac{t - T_s}{T - T_s}\right), & t > T_s \quad \text{(decay/cooldown)}
\end{cases}
\]

where $\phi(\cdot)$ may be linear, cosine, or square-root.  
Warmup mitigates early instability; decay phases act as implicit regularizers controlling the final descent into flatter minima.

\subsubsection{Optimal Linear Decay}

 demonstrate that linear decay schedules are theoretically near-optimal for a wide class of convex and smooth non-convex objectives under first-order dynamics \citep{defazio2023optimal}.  
By expressing the expected suboptimality bound as 
\[
\mathbb{E}[f(x_T)] - f^\star \le \mathcal{O}\!\left(\frac{1}{T}\sum_{t=1}^T \eta_t\right),
\]
they show that the linearly decaying schedule minimizes this upper bound under mild assumptions, even without exact curvature information.  
This helps reconcile empirical evidence for the simplicity and robustness of linear decay across large-scale training runs. 

Moreover, \citet{defazio2023optimal} present in Theorem 3 a refined version of their optimal-decay analysis that extends beyond SGD to preconditioned and coordinate-adaptive methods (like Adam). This allows automatic schedule refinements that adapt to observed gradient norms, producing warmup and late-stage annealing behavior without manual tuning. In other words, Theorem 3 suggests that if you can observe the sequence of gradient norms ahead of time (or perhaps by running a baseline schedule a priori), you can use them to compute an optimal schedule. Their empirical evaluation across 10 diverse deep learning problems, a series of LLMs, and logistic regression demonstrates that linear decay consistently outperforms cosine annealing and other common defaults, while adaptive refinements provide further gains.

\subsubsection{Constant + Linear Cooldown}

Recent work by \citet{schaipp2025surprising} reveals that the ubiquitous ``constant + cooldown'' schedule used in transformer pretraining aligns closely with non-smooth convex optimization theory.  
They show that for a stepwise schedule 
\[
\eta_t =
\begin{cases}
\eta_0, & t \le T_c \\
\eta_0 \!\left(1 - \frac{t - T_c}{T - T_c}\right), & t > T_c,
\end{cases}
\]
the resulting convergence bound avoids the logarithmic penalty appearing in standard diminishing step-size results. This theoretical alignment explains the empirical success of short cooldowns in large language model training. Moreover, they introduce LR transfer and continuation training, where tuning performed for one schedule can be re-used across architectures and scales. 

\citet{schaipp2025surprising} also derive a performance bound for the constant + linear cooldown schedule and show that, unlike many traditional diminishing-step bounds, their bound does not incur logarithmic penalty terms in $T$ (training rounds). In particular, the bound matches observed behavior in training large models, making cooldown analytically justified rather than merely heuristic. They further show that one can \emph{transfer} the optimal peak learning rate across architectures and then use \emph{continued training} beyond the nominal horizon with the same schedule to improve performance on 124M and 210M Llama-style models.  

\subsubsection{Warmup Stable Decay (WSD) and the Cooldown Effect}
\citet{dremov2025cooldown} empirically decompose the training dynamics of large transformers during the final cooldown phase of a WSD schedule. They show that a disproportionate amount of validation loss reduction occurs during the last 10-20\% of training, corresponding to small but highly structured steps in parameter space. Different decay shapes (linear, cosine, square-root) bias this late-stage trajectory differently: sharper decays favor convergence to sharper minima, while gentler ones (e.g., linear) yield flatter optima and improved generalization.

\begin{figure}
    \centering
    \includegraphics[width=0.75\linewidth]{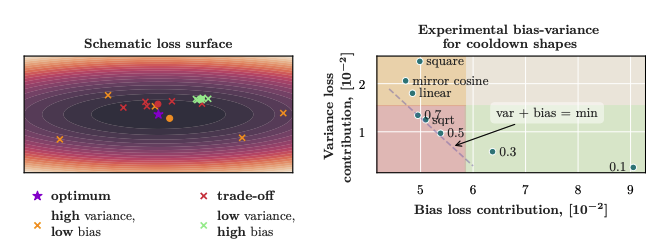}
    \caption{Figure 1 from \citep{dremov2025cooldown}. Left: Conceptual loss surface showing how different methods (colors) exhibit varying bias-variance tradeoffs, where crosses represent independent runs, circles show cluster means, and the star marks the optimum. Right: Empirical bias-variance characteristics of cooldown shapes, where bias measures distance to a better reference model and variance quantifies inconsistency across different data orderings; shapes like sqrt and 0.7 achieve optimal tradeoffs near the minimum of bias + variance.}
    \label{fig:bv_tradeoff_wsd}
\end{figure}

The authors formalize this intuition by proposing a bias-variance decomposition framework for cooldown shapes \citep{dremov2025cooldown}. For models trained from the same pre-cooldown checkpoint with different data permutations, they decompose the expected validation loss relative to a better reference model (obtained via longer training) as
\begin{equation}
\mathbb{E}_i[L(m_i) - L(m^*)] = \underbrace{L(\mathbb{E}_i[m_i]) - L(m^*)}_{\text{bias}} + \underbrace{\mathbb{E}_i[L(m_i)] - L(\mathbb{E}_i[m_i])}_{\text{variance}},
\end{equation}
where $m_i$ denotes the model obtained with data permutation $i$, $m^*$ is the averaged reference model, and expectations are taken over different data orderings. High learning rates during cooldown increase exploration of the loss surface, yielding high-variance low-bias solutions that differ substantially across runs but achieve lower average loss. Conversely, low learning rates encourage exploitation of the current basin, producing low-variance high-bias solutions that are more consistent but may sacrifice performance \citep{dremov2025cooldown}. Square-root and lowered-linear (with parameter 0.7) schedules empirically achieve the optimal bias-variance tradeoff, minimizing the sum of both components (see Figure \ref{fig:bv_tradeoff_wsd}).

This framework also emerges from a recency bias perspective: cooldown shapes that improve performance uniformly across all previously observed data points (low deviation) tend to exhibit higher bias, while shapes that achieve greater average improvement (low shift) but with uneven progress across the training data exhibit higher variance \citep{dremov2025cooldown}. Critically, the choice of AdamW hyperparameters $\beta_1$ and $\beta_2$ during cooldown yields performance variations comparable to those from cooldown shape selection itself, with surprisingly large values of $\beta_2$ (approaching 0.9999) consistently yielding improved results --- suggesting that longer momentum timescales help stabilize the descent into narrow basins during the final training phase.

\subsubsection{Interplay with $\mu$P and Optimizer Choice}

An open question is why learning rate schedules designed and analyzed for SGD often transfer effectively to adaptive methods like Adam. Defazio's extension to coordinate-adaptive methods provides partial theoretical justification, but the empirical robustness across optimizers remains not fully understood \citep{defazio2023optimal}. Additionally, the interaction between learning rate schedules and the Maximal Update Parameterization ($\mu$P) warrants investigation: does $\mu$P's width-invariant dynamics imply that schedule shapes should also remain stable across model scales? Preliminary evidence suggests that while $\mu$P enables transfer of peak learning rates, the \emph{shape} of the schedule (warmup duration, cooldown length) may still require scale-dependent adjustments, particularly for depth scaling where $\mu$Transfer has known limitations.

A related question concerns the interplay between learning rate schedules and the internal dynamics of second-order methods themselves. Recent work by \citet{eschenhagen2025pureshampoo} demonstrates that the empirical necessity of learning rate grafting in Shampoo stems from its role in compensating for stale and mis-scaled eigenvalues in the preconditioner, rather than from any fundamental property of the schedule shape. Specifically, they show that Shampoo's update magnitude exhibits dimension-dependent scaling ($m^{-p/2}n^{-p/2}$ factors) relative to full-matrix Adam, and that correcting the eigenvalues at every iteration --- while updating the eigenbasis only periodically --- eliminates the need for grafting entirely. Their eigenvalue-corrected variant (EShampoo) enables direct learning rate transfer from Adam without per-layer rescaling heuristics \citep{eschenhagen2025pureshampoo}. This raises a nuanced hypothesis: that learning rate schedules designed for SGD transfer effectively to adaptive methods not because the methods share identical loss landscape dynamics, but because grafting implicitly bridges the gap between their update magnitudes; furthermore, principled eigenvalue correction may reduce this schedule dependence. Whether such corrections remain sufficient under $\mu$P, where per-layer learning rate scaling is already governed by width-dependent initialization rules, remains as an open question. One could speculate that $\mu$P's coordinate-wise scaling interacts non-trivially with the eigenvalue structure of Kronecker-factored preconditioners: while $\mu$P ensures width-invariant peak learning rates, the \emph{shape} of the schedule may still require adjustment if the preconditioner's eigenvalue dynamics vary with depth or layer type.

\subsubsection{Synthesis}

Together, these findings indicate that learning rate schedules play a complementary role to optimizer design. First-order methods define directional dynamics in weight space, while the schedule defines temporal weighting over the optimization path. Linear decay emerges as a theoretically sound and computationally efficient default; constant + cooldown inherits strong convex guarantees; and WSD empirically exploits late-phase regularization, especially for adaptive optimizers like AdamW. In essence, schedules act as a global regularization mechanism that shapes both the rate and geometry of convergence in deep learning systems. 

The transition from convex line search methods, which require expensive per-iteration searches and curvature information, to fixed, pre-determined schedules reflects a pragmatic compromise: practitioners sacrifice per-step optimality for computational efficiency and empirical robustness. The success of this compromise hinges on careful empirical tuning and theoretical insights that identify broadly effective schedule families, even in the absence of problem-specific information.

\subsection{Exponential Moving Averages} \label{sec:ema}

As mentioned earlier, momentum have been shown to play a major role in improving the stability of neural network training. Moreover, they are well-regarded in stochastic optimization problems empirically, if not causing headaches to theorists who still look to prove how adapted concepts of momentum such as EMAs apply in nonconvex settings. We have already seen that EMAs provide a memory-efficient alternative to the uniform averaging schemes in dual averaging, while retaining the benefits of incorporating historical gradient information. Rather than storing and averaging all past gradients explicitly, EMAs maintain a single running statistic that exponentially discounts older gradients. This perspective has proven highly effective in deep learning, where memory efficiency and computational simplicity are paramount. However, interesting work is still being done to uncover how effective EMAs can really be for scaling neural network training to the billion-scale parameter regime.

\paragraph{Dual Averaging Revisited.}
\citet{jelassi2020dual} proposed \emph{Modernized Dual Averaging} (MDA), which matches SGD with momentum (SGD+M) and other popular deep learning algorithms on vision and NLP tasks. They interpret MDA as inducing a decaying, uncentered $\ell_2$ regularization relative to SGD+M. Because DA averages all past gradients uniformly, early gradients retain lasting influence, making the method more sensitive to the initial point and early optimization noise than exponentially weighted momentum. Moreover, the regularizing subproblem is anchored by $\lambda$ (often centered at the origin), yielding an implicit bias toward that center \citep{xiao2010dual}; with $\lambda(\mathbf{w})=\tfrac12\|w\|_2^2$, the iterates are pulled toward small-norm solutions, and the analysis in \citet{jelassi2020dual} identifies a decaying uncentered $\ell_2$ effect that interacts with initialization. In practice, warmup, careful stepsize schedules $\{\alpha_t\}$, or momentumized/adaptive DA can mitigate this sensitivity while retaining the benefits of averaging.

\paragraph{Bias Correction in Exponential Moving Averages.}
A fundamental challenge with EMAs is initialization bias: early in training, the moving average is dominated by its initialization (typically zero) rather than reflecting true gradient statistics. Adam addresses this through bias correction, computing $\hat{m}_t = m_t / (1 - \beta_1^t)$ and $\hat{v}_t = v_t / (1 - \beta_2^t)$, where $\beta_1$ and $\beta_2$ are the momentum coefficients for first and second moments respectively \citep{kingma2015adam}. This correction ensures that the early iterates are not unduly biased toward zero, particularly important when $\beta_1$ and $\beta_2$ are close to 1.

Recent work by \citet{block2025bema} proposes \emph{Bias-Eliminating Augmented Momentum Averaging} (BEMA), which fundamentally rethinks how to handle initialization bias in EMAs. Rather than applying multiplicative correction factors that grow inversely with $1 - \beta^t$, BEMA augments the momentum update with a carefully designed bias-elimination term. The key insight is that standard bias correction, while asymptotically correct, can introduce transient instabilities early in training when the correction factors are large.
BEMA maintains two auxiliary sequences alongside the primary parameters: a bias-tracking term $\alpha_t$ and an augmented EMA estimate $\tilde{\mu}$. At intervals of $\phi$ steps (after a burn-in period $\tau$), BEMA performs bias-corrected updates using:
\[
\alpha_t \leftarrow (\rho + \gamma t)^{-\eta} \quad \text{and} \quad \beta_t \leftarrow (\rho + \gamma t)^{-\kappa}
\]
\[
\tilde{\mu} \leftarrow (1 - \beta_t) \cdot \tilde{\mu} + \beta_t \cdot \mathbf{w}_t
\]
\[
\text{return} \quad \alpha_t (\mathbf{w}_t - \mathbf{w}_\tau) + \tilde{\mu}
\]
where $\mathbf{w}_t$ is the current parameter, $\mathbf{w}_\tau$ are the weights in the last burn-in round, $\tilde{\mu}$ is the exponentially moving average, $\alpha_t$ controls the bias correction strength (governed by bias power $\eta$), $\beta_t$ controls the EMA decay (governed by EMA power $\kappa$), $\rho$ is a lag parameter, $\gamma$ is a multiplier, and $\phi$ is the update frequency. The term $\alpha_t(\mathbf{w}_t - \mathbf{w}_\tau)$ explicitly corrects for initialization bias, with the correction strength decreasing over time as $\alpha_t$ decays.

\begin{figure}
    \centering
    \includegraphics[width=0.75\linewidth]{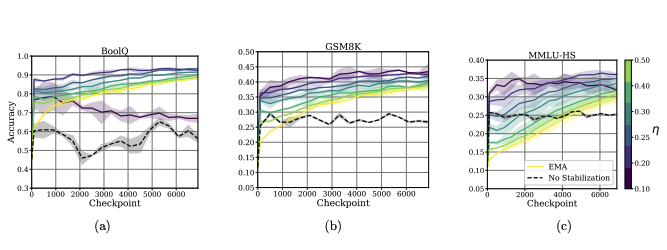}
    \caption{Figure 3 from \citep{block2025bema}: Effect of varying the $\eta$ hyperparameter (bias power) in BEMA while finetuning Qwen2.5-1.5B on (a) BoolQ, (b) GSM8K, and (c) MMLU-HS. Lower $\eta$ values strengthen the BEMA intervention over vanilla EMA, improving performance until $\eta$ becomes too small, causing collapse due to quadratic approximation failure}
    \label{fig:bema}
\end{figure}

The authors show that BEMA achieves convergence rate $\mathcal{O}(1/\sqrt{T})$ for non-convex stochastic optimization with the same per-iteration complexity as Adam, while exhibiting more stable training dynamics in the critical early phase. Empirically, BEMA matches or exceeds Adam and AdamW performance on vision and language modeling tasks while being more robust to the choice of momentum coefficients $\beta_1$ and $\beta_2$. This suggests that explicit bias elimination may be preferable to multiplicative correction when initialization effects are significant.

\paragraph{Practical Usage of EMAs.}
The choice between bias-corrected EMA (as in Adam), decoupled weight decay (as in AdamW, and explicit bias elimination (as in BEMA) depends on the specific optimization landscape and training regime. For large-batch training or when using aggressive momentum coefficients ($\beta_1 \geq 0.95$), the transient effects of bias correction can be substantial, making BEMA's approach more attractive. Conversely, for standard settings with moderate batch sizes and momentum coefficients, the simpler Adam or AdamW formulations often suffice. Recent empirical work suggests that the interaction between bias correction and learning rate schedules remains an active area where careful ablation studies are necessary to determine optimal configurations for specific problem domains \citep{block2025bema}.

\subsection{Weight Decay}

Another topic that we briefly touched but is considered elemental to the success of neural network training is weight decay, or the process of decreasing the magnitude of the weights during each update, thus pushing the weights closer to zero. Weight decay was originally conceived as explicit $\ell_2$ regularization, but we find that it plays a far more nuanced role in modern deep learning. Recent theoretical and empirical work reveals that weight decay fundamentally shapes the training dynamics of neural networks, particularly through its interaction with normalization layers and learning rate schedules. Rather than simply regularizing toward small weights, weight decay orchestrates a delicate balance in how different layers learn, while also introducing subtle pathologies that can emerge during extended training.

One of the earliest study on the interactions between weight decay and the optimization scheme was the development of AdamW. \citet{loshchilov2019decoupled} demonstrated that the initialization bias induced by EMAs can interact poorly with weight decay in Adam . They modified the base optimizer to decouple weight decay from the gradient-based update:
\[
\mathbf{w}_{t+1} = (1 - \eta \lambda) \mathbf{w}_t - \eta \cdot \frac{\hat{m}_t}{\sqrt{\hat{v}_t} + \epsilon},
\]
where $\lambda$ is the weight decay coefficient. This decoupling ensures that weight decay acts as true $\ell_2$ regularization rather than being entangled with the adaptive learning rate, improving generalization across many tasks.

\paragraph{Weight Decay and Rotational Equilibrium.}
\citet{kosson2024rotational} demonstrate that weight decay induces a steady state in the optimization dynamics they term \emph{rotational equilibrium}. In this regime, the magnitude and angular updates of a weight vector converge to a stable configuration where the norm-increasing effect of gradient updates precisely balances the norm-decreasing effect of weight decay. For a layer with weights $\mathbf{w}$ receiving gradient $G_t$, standard SGD with weight decay $\lambda$ and learning rate $\eta$ updates according to:
\[
\mathbf{w}_{t+1} = \mathbf{w}_t - \eta G_t - \eta \lambda \mathbf{w}_t.
\]
A crucial property of normalized layers (those followed by LayerNorm or BatchNorm) is that their gradients are orthogonal to their weights: $\langle G_t, \mathbf{w}_t \rangle = 0$. This orthogonality arises from the scale-invariance that normalization imposes --- scaling the weight vector does not affect the layer's output, so gradients must lie in the tangent space orthogonal to the radial direction.

Analyzing the squared norm dynamics under this orthogonality constraint, we find:
\[
\|\mathbf{w}_{t+1}\|^2 = \|\mathbf{w}_t - \eta G_t - \eta \lambda \mathbf{w}_t\|^2
\]
\[
= (1 - \eta\lambda)^2 \|\mathbf{w}_t\|^2
  - 2(1 - \eta\lambda)\eta \langle G_t, \mathbf{w}_t \rangle
  + \eta^2 \|G_t\|^2
\]
\[
= (1 - \eta\lambda)^2 \|\mathbf{w}_t\|^2
  + \eta^2 \|G_t\|^2
\]
where the cross term vanishes due to orthogonality. At rotational equilibrium, the weight norm stabilizes ($\|\mathbf{w}_{t+1}\| = \|\mathbf{w}_t\|$), and assuming small $\eta\lambda$ such that $(1 - \eta\lambda)^2 \approx 1 - 2\eta\lambda$, we obtain:
\[
\frac{\|G_t\|}{\|\mathbf{w}_t\|} \approx \sqrt{\frac{2\lambda}{\eta}}.
\]
This ratio represents the angular rotation magnitude per update, which serves as a proxy for the \emph{effective learning rate} of the layer. Remarkably, weight decay drives all normalized layers toward the same gradient-to-weight ratio, effectively balancing the rate of learning across different layers and neurons without requiring per-layer learning rate tuning. This homogenization partially explains why AdamW outperforms Adam with $\ell_2$ regularization --- the decoupled weight decay in AdamW preserves this balancing effect, while $\ell_2$ regularization entangles with the adaptive learning rates and disrupts the equilibrium.

The rotational equilibrium framework also illuminates the success of methods like LARS \citep{you2017large} and weight standardization, which explicitly normalize or control gradient-to-weight ratios. These techniques can be viewed as explicitly enforcing what weight decay achieves implicitly through its steady-state dynamics. This perspective reframes weight decay not as a regularizer in the traditional sense, but as an optimization tool that coordinates learning dynamics across the network architecture.

\paragraph{The Pathology of Gradient Norm Increase.}
While rotational equilibrium offers theoretical insight into weight decay's role, it also predicts a concerning pathology observed in practice: during long-duration training runs, particularly for large language models, gradient norms can increase dramatically near the end of training. This issue has been discovered and studied earlier by \citet{xie2023overlooked}, who identified that constant weight decay in the training process can lead to problematically large gradient norms at the final phase of training, particularly when using adaptive optimizers like Adam. Large gradient norms at convergence indicate both poor optimization (failure to reach a true stationary point) and poor generalization (correlation with sharp minima). To address this, \citet{xie2023overlooked} proposed \emph{Scheduled Weight Decay} (SWD), a gradient-norm-aware scheduler that dynamically adjusts the weight decay strength to penalize large gradient norms during training.

\begin{figure}
    \centering
    \includegraphics[width=0.75\linewidth]{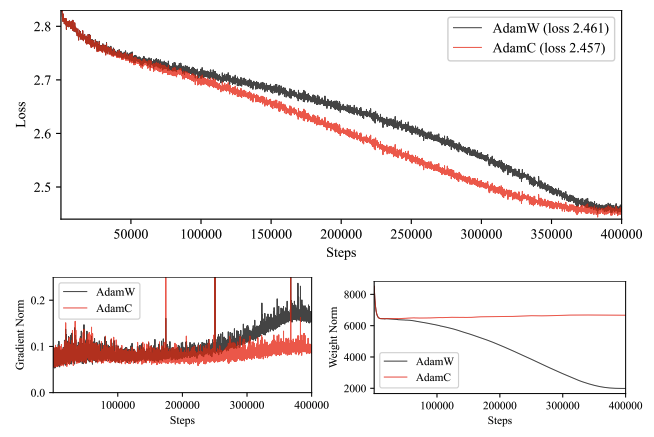}
    \caption{Figure 4 from \citep{defazio2025gradients} depicting head-to-head comparisons of Adam(C) and Adam(W) training a 120M Llama 3 on FineWeb-Edu dataset.}
    \label{fig:adamc_adamw}
\end{figure}

In a similar vein, \citet{defazio2025gradients} identifies this phenomenon as an unintended consequence of the interaction between weight decay, normalization layers, and time-varying learning rate schedules. The theoretical steady-state ratio $\|G_t\| / \|\mathbf{w}_t\| \approx \sqrt{2\lambda/\eta}$ suggests that when a decaying learning rate schedule is employed (such as cosine annealing where $\gamma_t \to 0$), the target gradient-to-weight ratio increases as $1/\sqrt{\eta}$. Although the reduced step size slows convergence to this moving target, the rapid decrease in $\gamma_t$ near the end of training causes gradient norms to spike as the system attempts to track the increasing theoretical ratio. AdamW only lightly mitigates this issue: Figure \ref{fig:adamc_adamw} illustrates this behavior in a 120M LLM training run, where gradient norms almost double during the final phase despite the loss continuing to decrease.

This pathology is problematic for several reasons. First, monitoring gradient norms is a standard diagnostic for training health, and such increases can be mistaken for training instability or divergence. Second, the increasing gradients interact with other training dynamics in unpredictable ways, potentially affecting final model quality. Third, this behavior complicates the transfer of hyperparameters across different training durations, as the endpoint dynamics differ substantially from mid-training behavior.

Defazio proposes a simple correction: rather than using a fixed weight decay coefficient $\lambda$, adjust it proportionally to the learning rate schedule to maintain a constant steady-state weight norm. Specifically, for AdamW, define:
\[
\lambda_t = \lambda_{\text{base}} \cdot \frac{\gamma_t}{\gamma_0},
\]
where $\gamma_0$ is the initial learning rate. This \emph{scheduled weight decay} ensures that the theoretical gradient-to-weight ratio $\sqrt{2\lambda_t/\gamma_t} = \sqrt{2\lambda_{\text{base}}/\gamma_0}$ remains constant throughout training, eliminating the pathological gradient norm increase. Empirically, this corrected Adam(W) variant, termed ``Adam(C),'' maintains stable gradient and weight norms throughout 200B token training runs while achieving lower final loss values than standard AdamW as depicted in Figure \ref{fig:adamc_adamw} \citep{defazio2025gradients}.

These studies present promising results on interpreting the role of weight decay in neural network training. AdamS optimizer (Adam with SWD) demonstrates that adaptive scheduling of weight decay can bridge the generalization gap between Adam and SGD, achieving comparable generalization performance without introducing additional hyperparameters. While the analysis in \citet{defazio2025gradients} later analysis provides the theoretical mechanism --- showing that the gradient-to-weight ratio increases as learning rate decays --- and proposes scheduling weight decay proportionally to the learning rate ($\lambda_t \propto \gamma_t$), \citet{xie2023overlooked} empirically demonstrated the importance of controlling gradient norms through weight decay scheduling, particularly for adaptive methods where the pathology is most pronounced. Moreover, while Defazio's analysis technically applies only to layers immediately followed by normalization, experiments show that treating most linear layers as normalized (excluding only the output layer) yields the predicted benefits. This suggests that the rotational equilibrium framework captures essential training dynamics even in architectures where not all layers strictly satisfy the orthogonality assumption.

\paragraph{Broader Implications.}
These findings fundamentally reshape our understanding of weight decay's role in modern optimization. Rather than viewing weight decay as a form of regularization that biases toward smaller weights, we should recognize it as a mechanism that:
\begin{itemize}
\item Controls the effective learning rate through gradient-to-weight ratios,
\item Coordinates learning dynamics across layers via balanced rotation,
\item Interacts critically with learning rate schedules in ways that require explicit management.
\end{itemize}
The rotational equilibrium perspective also connects to recent work on maximal update parameterization ($\mu$P) and learning rate scaling rules \citep{yang2022tensorprogramsV}. As network width increases, maintaining balanced rotation across layers requires careful tuning of weight decay alongside learning rates --- a consideration that becomes increasingly important for training at scale. The pathological gradient norm increase identified by Defazio highlights that even well-established training practices can harbor subtle issues that only manifest during extended training runs, underscoring the importance of understanding optimizer dynamics beyond asymptotic convergence guarantees.

\subsection{Final Remarks}

The interplay between parameterization, learning rate schedules, momentum averaging, and weight decay forms a complex but increasingly well-understood system at the heart of modern deep learning optimization. Through $\mu P$'s preservation of feature learning at scale, the emergence of practical scheduling strategies that balance theoretical guarantees with computational constraints, and the careful management of gradient aggregation and regularization dynamics, we can now train models at unprecedented scales with greater stability and predictability. These advances represent not just incremental improvements but fundamental shifts in how we think about optimizer design --- from ad-hoc hyperparameter tuning to principled, transferable configurations that work across model scales. The gap between theoretical understanding and practical success in deep learning optimization, while still present, continues to narrow through these unified frameworks.

For practitioners training large models, several well-backed recommendations emerge from these works. First, adopt $\mu P$ (or u-$\mu P$) parameterization before training billion parameter models as it enables hyperparameter transfer from small proxy models, dramatically reducing computational costs that would be required for tuning the larger model. Second, use a warmup-stable-decay schedule with linear or cosine decay, allocating roughly 5-10\% of training to warmup and 20---30\% to decay phase --- this balances early stability with late-stage refinement. Third, when using AdamW, ensure weight decay is properly decoupled from the adaptive learning rate, and consider scheduling weight decay proportionally with the learning rate (as in SWD/Adam(C)) to prevent gradient norm explosion during the decay phase. Fourth, for tasks requiring stable early training or when finetuning large pretrained models, consider BEMA over standard Adam to eliminate initialization-dependent biases while maintaining momentum benefits. Finally, monitor gradient norms throughout training as a diagnostic --- sudden spikes during learning rate decay often indicate the need for scheduled weight decay, while persistently small norms may suggest overly aggressive regularization or suboptimal parameterization choices.

To validate these theoretical insights and practical recommendations, we now turn to a comprehensive experimental evaluation that explores the intricate relationships between optimizers, schedules, and regularization strategies. We particularly focus on the dynamics of curvature-aware methods, the universality of certain scheduling patterns, and the critical gradient-norm pathologies that emerge when components are misaligned. Through these controlled studies, we aim to bridge a few of the remaining gaps between theoretical predictions and empirical observations, while also exploring novel hybrid approaches that leverage the strengths of multiple optimization paradigms.

\section{Experiments}\label{sec:exp}

We include this section to discuss the various experiments we ran to confirm the findings of the referenced papers as well as to build on their insights and highlight key results.

\begin{figure}
    \centering
    \includegraphics[width=0.5\linewidth]{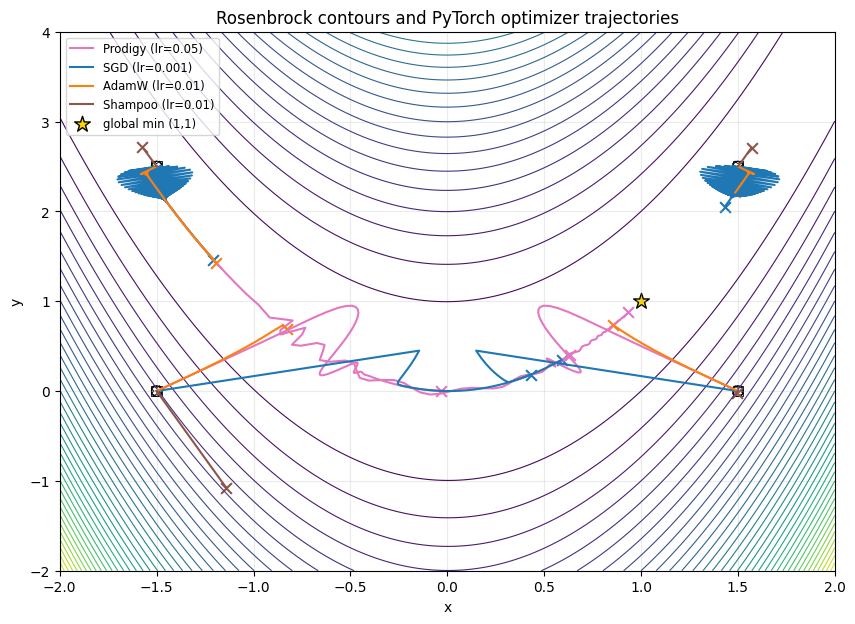}
    \caption{Rosenbrock descent for Shampoo, SGD, Adam, and Prodigy}
    \label{fig:rosenbrock}
\end{figure}

\paragraph{Rosenbrock Function Descent.}
The Rosenbrock function serves as a fundamental benchmark in optimization, presenting a challenging non-convex landscape with a narrow, curved valley. This classical test case allows us to isolate and evaluate how curvature-based methods perform relative to standard first-order approaches in a controlled setting before moving to more complex neural network training scenarios. See Figure \ref{fig:rosenbrock} to see the results of running SGD, AdamW, Shampoo, and Prodigy on the Rosenbrock function from 4 different starting points for 500 steps. The figure reinforces the idea that SGD is extremely sensitive to the gradients it experiences during training (especially without momentum), causing it to oscillate in the two upper starting points but progress reasonably in the lower two. Shampoo seems to diverge and remains unstable in all 4 cases, likely due to the noise in maintaining the preconditioner over a small input size. AdamW and Prodigy shine as their adaptive learning rates enable them to explore the full region, with Prodigy even making significant progress towards the global minimum in from 3 out of 4 starting points. Lastly, we see that the top right starting point of (1.5, 2.5) is bad for all of the optimizers, hinting at the fact that some initialization points even on small-scale non-convex problems can lead to poor optimizer performance.

\begin{figure}
    \centering
    \includegraphics[width=1\linewidth]{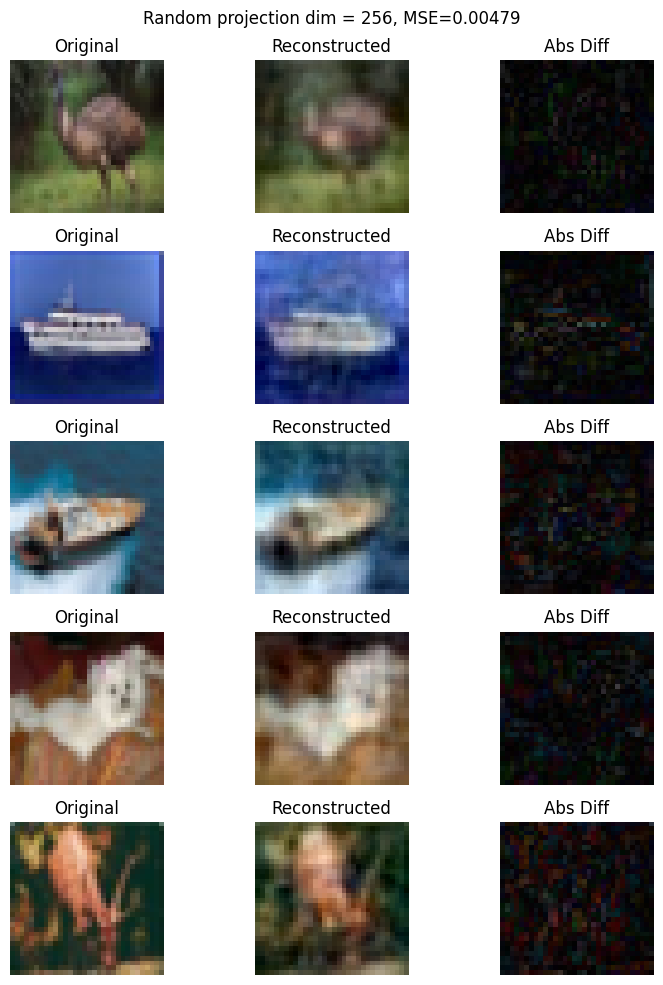}
    \caption{Examples of the Downsampled CIFAR-10 Data via Random Orthogonal Projections}
    \label{fig:ds_ex}
\end{figure}

\paragraph{Data Sampling Methods}
We take the canonical CIFAR-10 dataset, widely used for neural network evaluations, and apply dimensionality reduction via random projection \citep{krizhevsky2009cifar}. Specifically, we project the high-dimensional image data ($32 \times 32 \times 3 = 3072$) onto a lower-dimensional subspace ($d = 256$) using a random orthogonalized matrix via QR decomposition. This approach is theoretically justified by the Johnson-Lindenstrauss lemma, which guarantees that a set of $n$ points in high-dimensional Euclidean space can be embedded into $O(\log n / \varepsilon^2)$ dimensions while preserving all pairwise distances to within a factor of $(1 \pm \varepsilon)$ \citep{dasgupta2003jllemma}. Since classification fundamentally depends on the geometric relationships between data points, this distance-preserving property ensures that the essential structure relevant for learning is retained. This downsampling strategy enables computationally efficient training on standard CPU hardware without requiring GPU acceleration, while also introducing a controlled noise perturbation to a well-studied benchmark, allowing us to evaluate how robust these methods are under transformed representations. See Figure \ref{fig:ds_ex} to observe how the downsampling method affects the data samples while preserving just enough information to power classification.

\begin{figure}
    \centering
    \includegraphics[width=1\linewidth]{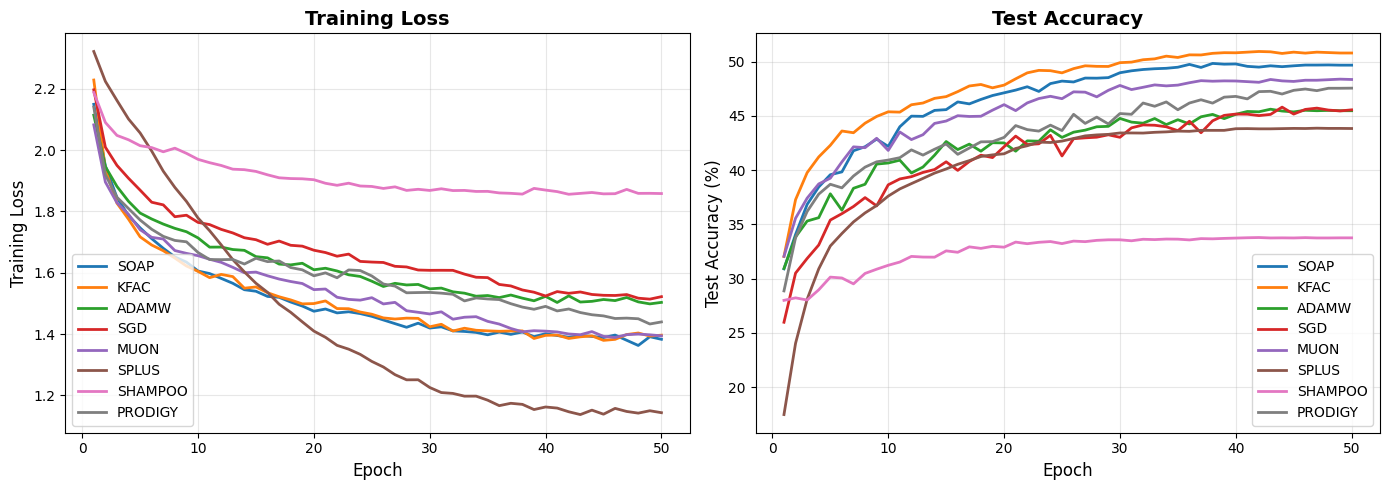}
    \caption{Head to Head Optimizer Studies on 3-layer 256-dim MLP}
    \label{fig:base_h2h}
\end{figure}

\paragraph{Head-to-Head Optimizer Studies.}
To establish a fair comparison baseline, we trained a simple 3-layer MLP with hidden dimension of 256 on the downsampled CIFAR-10 for 50 epochs. We swept hyperparameters for each major optimizer and for simplicity's sake chose the model that minimized the test loss across all permutations. We converged on using the cosine annealing learning rate schedule as this delivered the best performances across the other examples, although some methods may be better under different learning rates as we explore in another experiment. This controlled experiment isolates the algorithmic differences between optimizers by fixing the architecture, dataset, and hyperparameter search budget, allowing us to directly assess their relative performance under ideal conditions on a well-defined problem. See Figure \ref{fig:base_h2h} for the full plot of train loss and test accuracy.

Our results provide evidence that on smaller models trained on data with noise baked into its sampling method, curvature-aware optimizers perform feature learning remarkably fast, while Adam and SGD induce a clear suboptimality gap even under best-tuned hyperparameters. Notably, Shampoo failed to converge in this learning scenario; although others have reported success with Shampoo on larger models, our findings suggest that its preconditioners do not stabilize at this scale without modifications such as SOAP or SPlus. We observe that SPlus aggressively decreases training loss but begins to exhibit overfitting behavior over the tail end of its epoch count. In terms of training efficiency, Muon achieved the quickest wall-clock time, reaching 45\% test accuracy in approximately 1 minute 20 seconds (15 epochs), while SOAP hit this benchmark in only 10 epochs albeit with longer wall-clock time due to its higher per-iteration cost.

\begin{figure}
    \centering
    \includegraphics[width=1\linewidth]{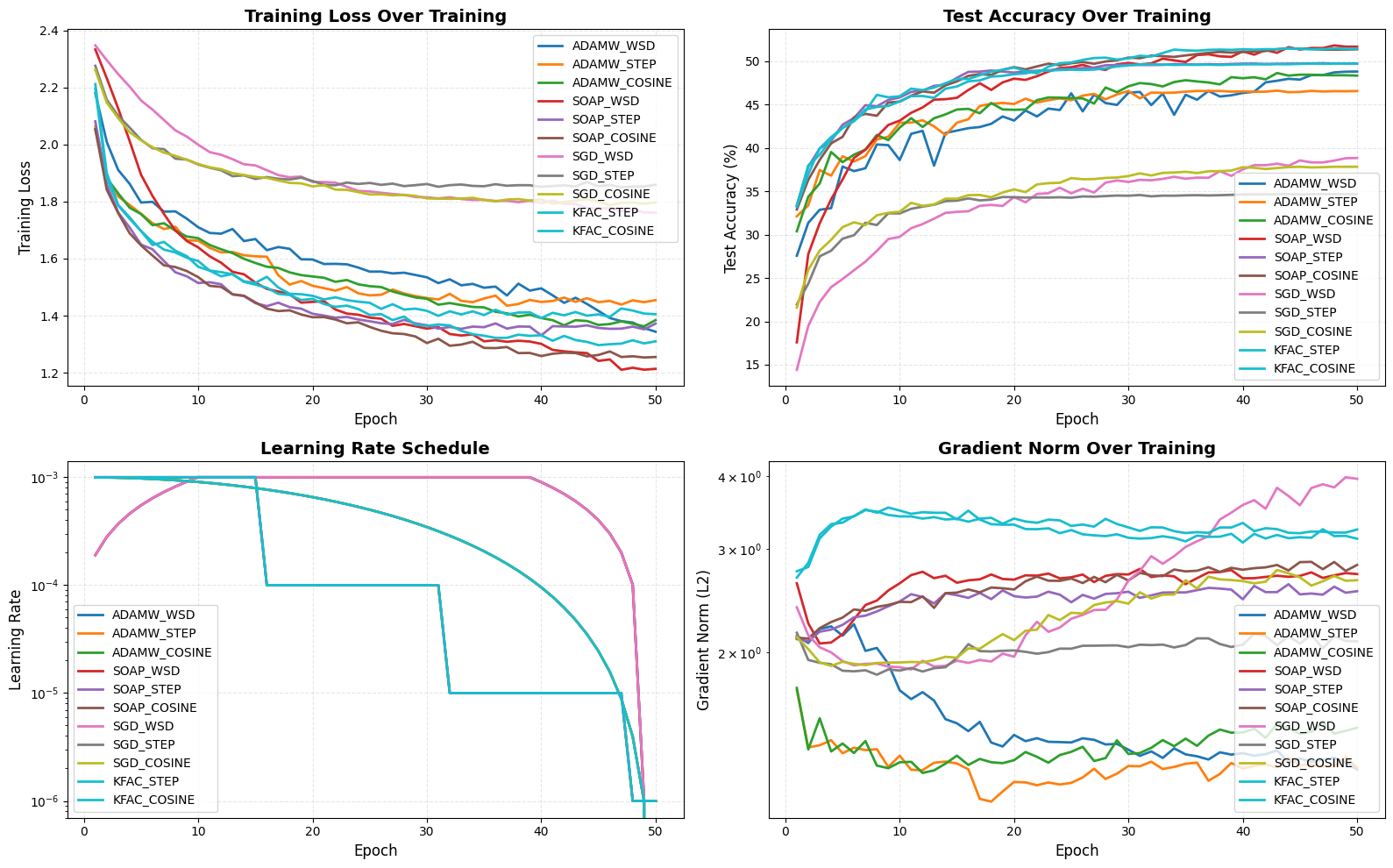}
    \caption{Learning Rate Schedules over 3-layer 256-dim MLP}
    \label{fig:lrs_h2h}
\end{figure}

\paragraph{Learning Rate Schedules.}
Learning rate schedules play a crucial role in optimization performance, yet their interaction with different optimizers remains underexplored. We systematically paired various optimization algorithms with common learning rate schedules --- cosine annealing, step decay, and warmup-stable-decay (WSD) --- to investigate whether certain schedules exhibit universal benefits across optimizer families or whether schedule effectiveness is optimizer-dependent. We used the same 3-layer 256-dim MLP from above to enable direct comparison of results. See Figure \ref{fig:lrs_h2h} for the full results.

Our findings suggest that learning rate schedules do not have a pronounced effect on training performance in this setting. We posit that for smaller-scale training tasks, the choice of optimizer is significantly more indicative of final performance than the learning rate schedule employed, a finding that may not generalize to larger models where schedule choice has been shown theoretically and empirically to be critical.

To better understand the optimization dynamics at play, we tracked the evolution of gradient norms throughout training. SGD exhibits steadily increasing gradient norms as training progresses, which can indicate difficulty navigating the loss landscape and may partly explain its suboptimality gap observed in our earlier experiments. Well-tuned Adam, by contrast, maintains stable gradient norms throughout training, likely due to its adaptive moment estimates providing implicit gradient scaling. However, the more sophisticated curvature-aware methods do not substantially outperform Adam on this metric in our downsampled data scenario, suggesting that while these methods offer faster convergence and better final performance, their advantage stems from improved update directions rather than superior gradient norm control.

\paragraph{Hybrid Approaches.}

\begin{figure}[!tbp]
    \centering
    \begin{subfigure}[b]{0.75\textwidth}
        \centering
        \includegraphics[width=0.9\textwidth]{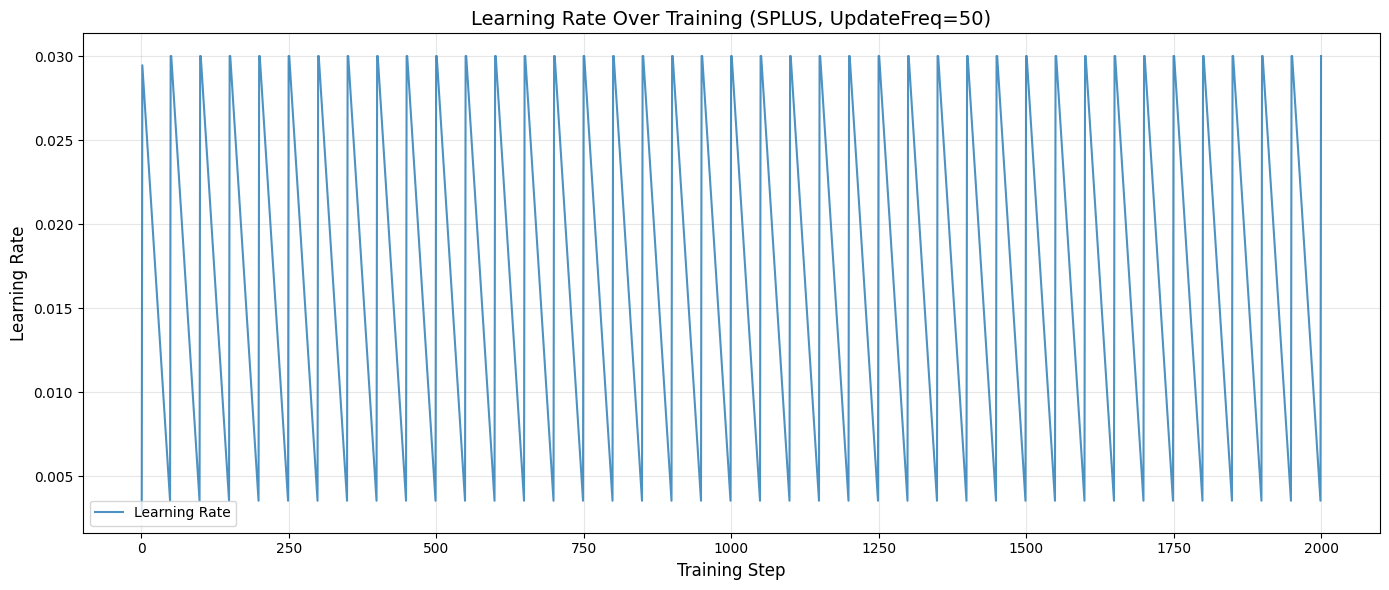}
        \caption{LR Spike of Factor 10 for SPlus at Update Frequency of 50}
        \label{fig:spike_ex}
    \end{subfigure}
    
    \vspace{0.5em}
    
    \begin{subfigure}[b]{\textwidth}
        \includegraphics[width=\textwidth]{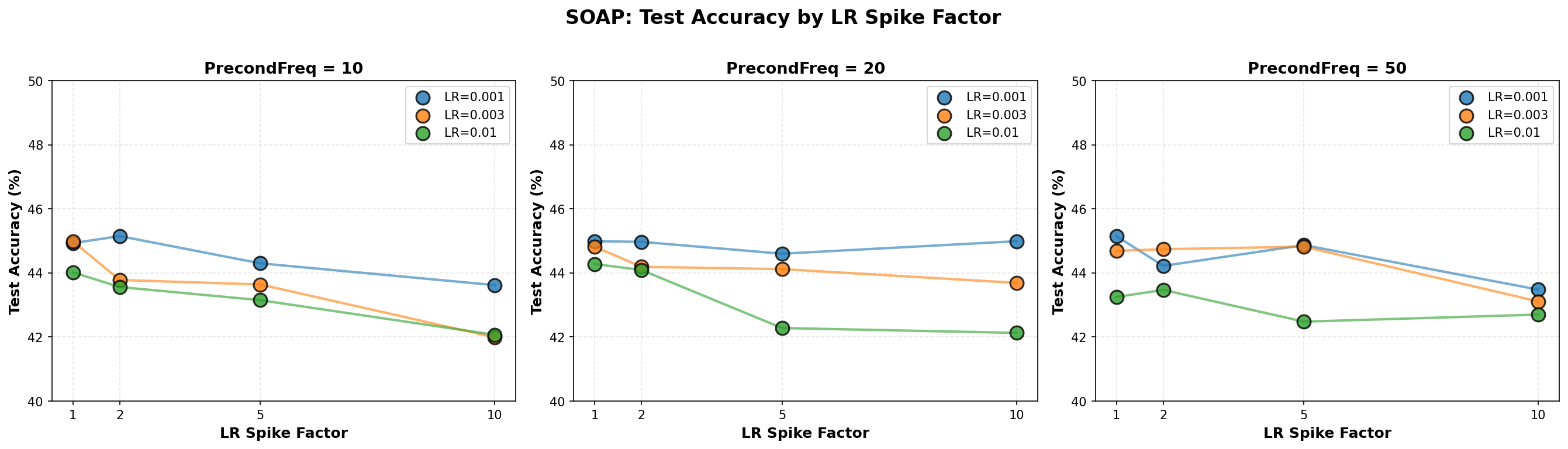}
        \caption{SOAP}
        \label{fig:soap_spike}
    \end{subfigure}
    
    \vspace{0.5em}
    
    \begin{subfigure}[b]{\textwidth}
        \includegraphics[width=\textwidth]{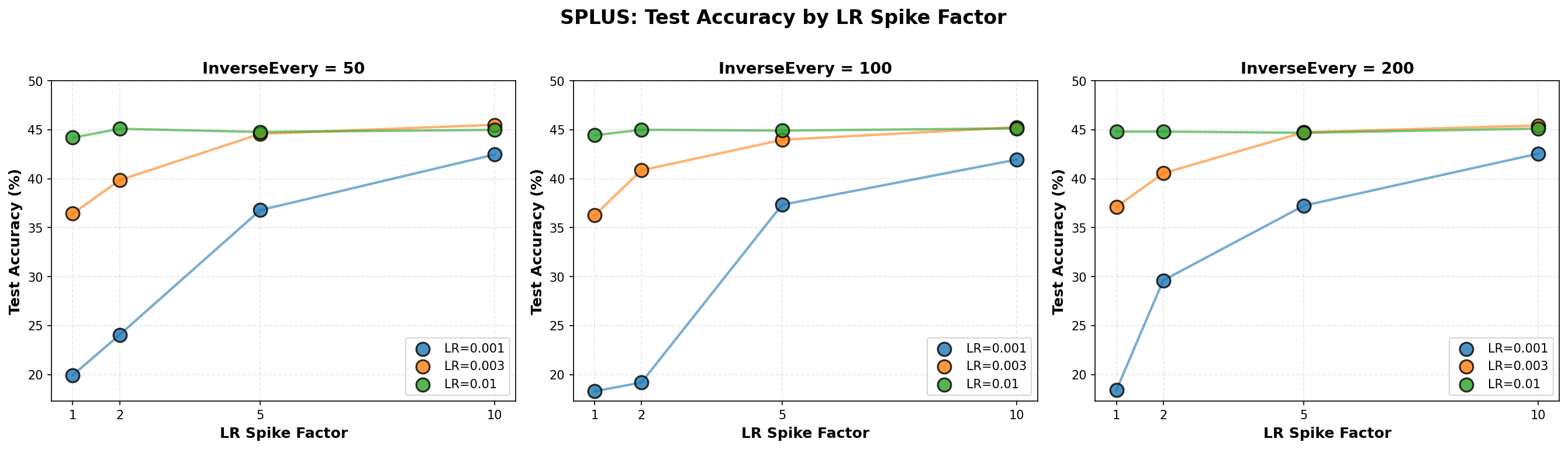}
        \caption{SPlus}
        \label{fig:splus_spike}
    \end{subfigure}
    \caption{Learning rate spike experiments. (a) Example of the LR spike schedule over training. (b, c) Final test accuracy by spike factor across various preconditioner update intervals.}
    \label{fig:spike_comparison}
\end{figure}

Recent work suggests that curvature information quality varies throughout training, motivating adaptive strategies that exploit high-confidence curvature updates. We explore hybrid learning rate schemes that synchronize learning rate ``spikes'' with periodic preconditioner update steps in SOAP and SPlus, hypothesizing that fresh curvature information warrants increased learning from these directions. See Figure \ref{fig:spike_ex} for an example of what the LR spike schedule looks like over a full training sequence. See Figure \ref{fig:soap_spike} for test accuracy by LR spike factor over various preconditioner update intervals for SOAP and Figure \ref{fig:splus_spike} for SPlus.

Contrary to our hypothesis, the learning rate spike during SOAP preconditioner updates does not appear to have a significant effect on test accuracy, which we use as a proxy for generalization. SPlus does exhibit modest benefits from spikes at small base learning rates (e.g., 0.0001), but equivalent performance can be elicited by simply using a higher learning rate consistently throughout training, obviating the need for the added complexity of synchronized spikes.

We offer two potential explanations for these null results. First, the scale of our experiment may be insufficient to surface the benefits of curvature-synchronized updates; larger models with richer loss landscapes may present more opportunities for exploiting fresh curvature information. Second, in a mini-batch setting, the preconditioner update is inherently noisy and may not provide a sufficiently accurate approximation of local curvature to justify aggressive learning rate increases at update steps. Further investigation on larger-scale tasks would be needed to determine whether this technique offers any practical benefits.

\begin{figure}
    \centering
    \includegraphics[width=1\linewidth]{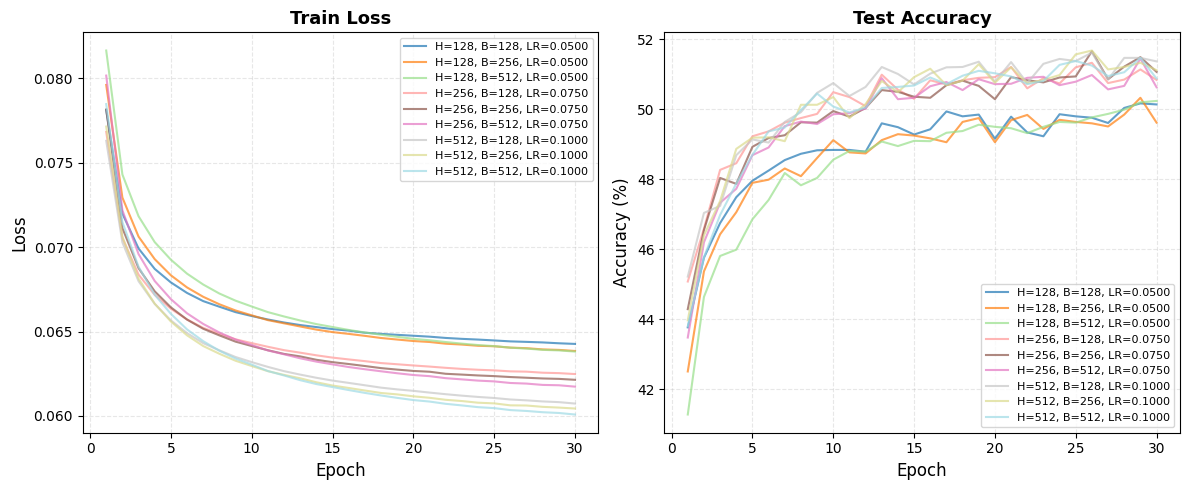}
    \caption{Modula Head-to-Head Runs on 3-layer 256-dim MLP}
    \label{fig:modula_h2h}
\end{figure}

\paragraph{Modula Framework Validation.}
We validate the \texttt{modula} optimization library by reproducing key results from the literature and demonstrating its extensibility to modern architectures. We trained 3-layer \texttt{modula} networks on downsampled CIFAR-10 data while varying the hidden dimension, base learning rate, and batch size. See Figure \ref{fig:modula_h2h} for results. Following the library defaults, we use a base target norm of 1 for linear layers, which corresponds to the max-of-max norm discussed in Section \ref{sec:modnorm} as the update rule analogous to Adam \citep{modula-docs}.

The results are striking: \texttt{modula} achieves comparable performance to even the second-order optimizers evaluated on the same dataset and architecture, with essentially no hyperparameters to tune. Some runs reach 50\% test accuracy in just 10 epochs, matching or exceeding the convergence speed of heavily-tuned baselines. This provides an extremely simple entry point for metrized deep learning \citep{modula-docs}. Due to limited computational resources, we were unable to explore \texttt{modula}'s recent support for Transformer architectures, which we hope to investigate in future work. Additionally, the library's easy-to-use framework opens the door to Shampoo-like update rules and momentum incorporation, which remain promising directions for extension.

\begin{figure}
    \centering
\includegraphics[width=1\linewidth]{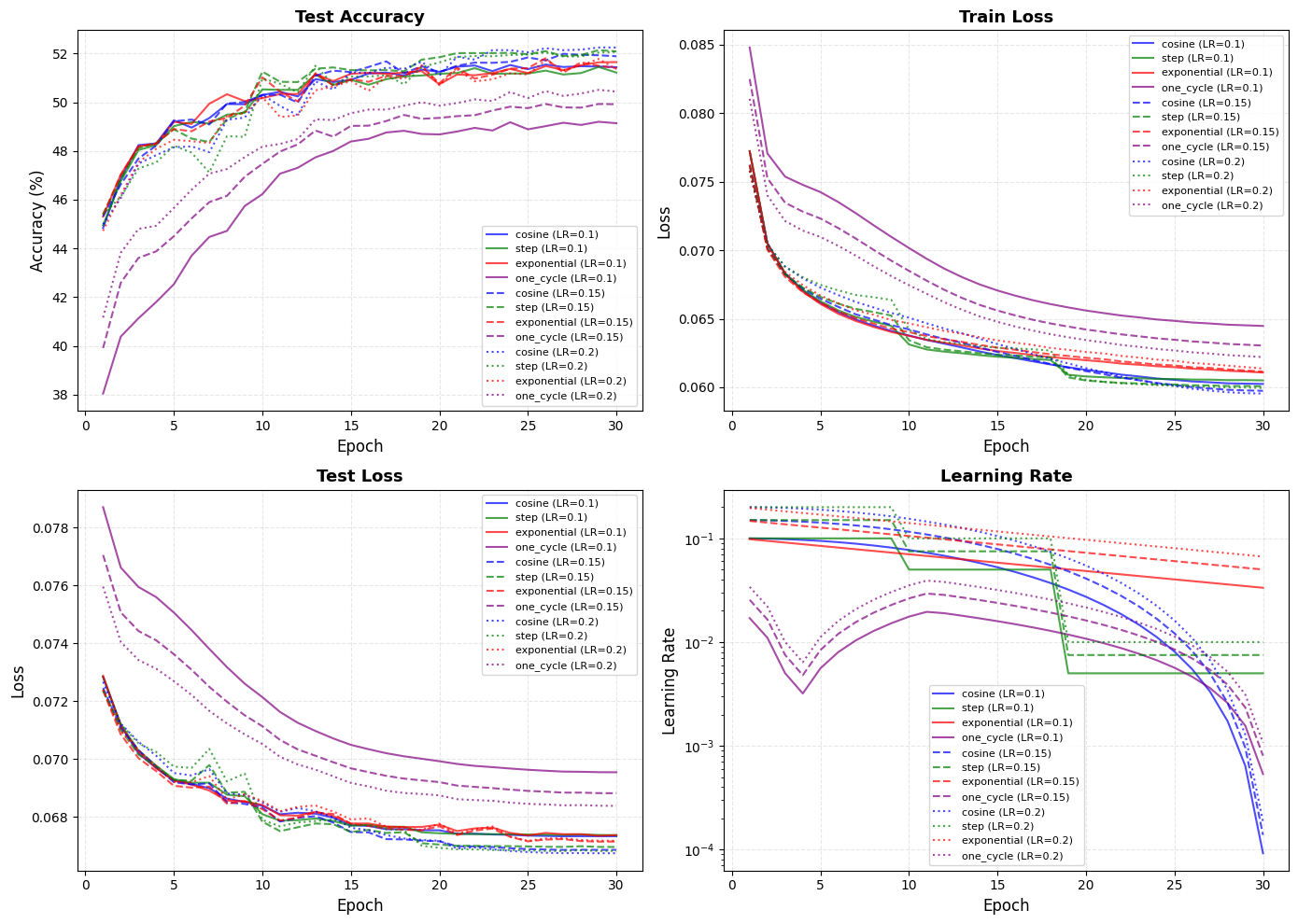}
    \caption{Modula 3-layer 256-dim MLP Adapted with Common Learning Rate Schedules}
    \label{fig:lrs_modula}
\end{figure}

We also applied various learning rate schedules on top of \texttt{modula}'s built-in optimizer to identify potential easy wins. We varied starting learning rates and applied cosine annealing, step decay, exponential decay, and one-cycle schedules. See Figure~\ref{fig:lrs_modula} for results. One-cycle performs significantly worse than the other schedules, likely due to its gradual warmup phase and steep drop-off behavior conflicting with \texttt{modula}'s internal dynamics. Beyond this, \texttt{modula} appears largely schedule-invariant in our setting: this is mostly consistent with Bernstein's results on the initial tests of the library where he uses a linear decay schedule, which empirically achieves similar performance to any schedule that decays toward the end of training. A natural follow-up would be to explore schedule-free methods or automatic learning rate adaptation approaches such as Prodigy~\citep{mishchenko2023prodigy}.

\paragraph{Concluding Remarks}

The experiments presented in this work were not designed to introduce novel techniques or uncover new phenomena, but rather to verify and consolidate results from the recent optimization literature in a simple, controlled setting. By using a downsampled CIFAR-10 dataset and modest 3-layer MLPs, we intentionally prioritized accessibility and reproducibility over state-of-the-art performance. This testbed allowed us to quickly validate core claims across multiple optimizer families, from classical methods like SGD and Adam to modern curvature-aware approaches like KFAC, Shampoo, and Muon, without the confounding factors that could be introduced by large-scale training.

We acknowledge that more time could have been spent on better datasets, extensive hyperparameter tuning, and pushing absolute performance numbers. However, such efforts would have detracted from our primary goal: providing a unified, easy-to-validate exploration of the ideas surveyed in this paper. Each individual work covered here contains more rigorous experiments specifically tailored to their contributions, and we encourage readers to consult those original papers to fully appreciate the power of these methods in their intended applications.

We hope this work serves as a practical starting point for practitioners looking to understand the landscape of modern optimizers, and that our experimental framework provides a foundation for future studies extending these ideas to larger models and more challenging tasks.

\section{Conclusions}

The question of which optimizer to use for deep learning is not one that admits a single universal answer, but rather depends critically on the training context, model architecture, and computational constraints at hand. For foundation model pretraining --- where features are deep, condition numbers are high, and training runs span weeks or months --- higher-order methods such as KFAC and Muon have emerged as particularly well-suited choices \citep{liu2025muon}. These methods leverage structured approximations of curvature information to achieve both faster convergence and improved compute efficiency.

For large language model pretraining specifically, Muon has demonstrated remarkable superiority over AdamW. Recent work shows that Muon explicitly expands the Pareto frontier over AdamW on the compute-time tradeoff, retaining data efficiency at large batch sizes far beyond the critical batch size while remaining computationally efficient \citep{shah2025practical}. The key to scaling Muon lies in adding weight decay and carefully adjusting per-parameter update scales, enabling the optimizer to work out-of-the-box on large-scale training without extensive hyperparameter tuning. Scaling law experiments indicate that Muon achieves approximately $2\times$ computational efficiency compared to AdamW under compute-optimal training conditions \citep{liu2025muon}. Given these substantial advantages, it would be naive to assume that frontier laboratories are still relying solely on AdamW or your classic first-order algorithm. The competitive dynamics of the AI industry virtually guarantee that leading labs have already incorporated Muon or related proprietary second-order methods into their training stacks. What the research community discovers and publicizes today, industry likely deployed months ago; the "secret sauce" behind the most capable models almost certainly includes optimization innovations that will never appear in any pre-print until it's well known throughout the community.

One might ask, given the immediate success of LLMs in many generalized tasks, whether optimization is effectively "solved" for pretraining and that all we need is better training data for fine-tuning models for specialized use cases. While recent research advances and growing interest in the startup world have substantially narrowed the gap between theory and applied AI, the implications of training efficiency improvements extend far beyond academic interest. In an era where data centers consume unprecedented amounts of energy, even modest improvements in training time translate directly into reduced environmental impact and computational cost. Moreover, the frontier of foundation model development continues to expand into new domains --- robotics, world models, embodied AI --- where the optimization challenges remain far from settled. Critically, the optimization problem is still demonstrably unsolved for post-training and fine-tuning regimes: methodologies such as supervised fine-tuning and reinforcement learning from human feedback face distinct challenges such as catastrophic forgetting and the need to preserve pretrained representations while adapting to new objectives \citep{aslam2025llm}. When models fail to capture features appropriately during scaling, the choice of optimizer becomes essential for maintaining training stability and final performance. The interplay between optimization algorithms and parameter-efficient fine-tuning methods like LoRA remains an active area of investigation.

Adam and its variants retain important advantages for downstream tasks and rapid learning scenarios where deep feature extraction is not the primary concern. The heterogeneous block structure of Transformer architectures creates loss landscapes where different parameter blocks exhibit dramatically different Hessian spectra  \citep{zhang2024transformers}. This heterogeneity fundamentally hampers SGD-like optimizers, while Adam's adaptive learning rates naturally accommodate such structural variation. For fine-tuning on top of foundation models, Adam's adaptivity may get it most of the way there, but whether its downstream performance can be attributed to suboptimal feature learning during fine-tuning stages or from the foundation model remains to be  explored. 

It is essential to emphasize that modern deep learning optimization is not "plug-and-play." Different weight geometries demand different optimization strategies, and the raw optimizer update rule represents only one component of a successful training recipe. Weight decay, learning rate schedules, exponential moving average implementations, gradient clipping, and architecture-specific modifications all interact in complex ways to determine final model quality. $\mu P$ offers principled hyperparameter transfer across scales, but practical deployment requires careful attention to these auxiliary techniques. Nevertheless, when practitioners seek a single recommendation for out-of-the-box performance with minimal tuning, \textit{Muon} emerges as the most practical choice for large-scale pretraining. However, properly tuned AdamW --- particularly with schedule-free variants or warmup-stable-decay schedules --- remains highly competitive. The AlgoPerf benchmark demonstrates that well-configured AdamW continues to perform excellently across diverse workloads, even achieving state-of-the-art results in the self-tuning track of recent algorithmic efficiency competitions \citep{dahl2023algoperf}. The choice between these methods often reduces to practical considerations: available compute, tuning budget, and domain-specific requirements. Note that despite Adam's widespread adoption, it still suffers from well-documented generalization limitations \citep{keskar2017improving,wilson2017marginal,zhang2025concurrence}. Even on competitive benchmarks, smart variants of SGD with proper tuning can outperform Adam on tasks where generalization is paramount. This goes to demonstrate that one can't simply apply Adam-like optimizers to their deep learning problems without a deep understanding of what they require in performance.

Through this survey, we have traced the evolution of neural network optimization: from the foundations of momentum methods and quasi-Newton approaches through the development of adaptive gradient algorithms that addressed the limitations of vanilla SGD, and finally with curvature-aware methods that leverage structured approximations of second-order information along with architecture-aware optimization. We hope to have contributed to transforming neural network optimization from an empirical art into a principled science, providing both theoretical insights and baseline prescriptions for the next generation of deep learning systems.

\section{Future Work}

Several directions that were brought up in this work remain unexplored, each offering opportunities to further bridge theory and practice in deep learning optimization:

\paragraph{Feature Normalization.}
This work has focused on weight-space geometry while largely setting aside the complementary question of how to deal with the activation-space. Feature normalization techniques such as BatchNorm and LayerNorm are now ubiquitous, yet their interaction with the optimization landscape remains incompletely understood \citep{ioffe2015batch,ba2016layernorm}. While input whitening accelerates learning by decorrelating activations, it risks discarding informative structure. Lubana et al.\ provide a unified analysis revealing distinct mechanisms --- activation stabilization, informative forward propagation, and gradient norm control --- that vary across normalization variants \citep{lubana2021beyondbatchnorm}. A systematic study of how these feature-space transformations interact with parameter-space preconditioning could yield optimization strategies that jointly exploit both sources of structure.

\paragraph{Extension to Non-Euclidean Geometries.}
Our treatment has assumed Euclidean parameter spaces, yet many neural network components possess inherent geometric constraints: rotation matrices lie on orthogonal groups, covariance matrices must remain positive definite, and attention weights live on probability simplices. Extending the principled optimization framework developed here to respect such manifold structure is a natural next step. Recent advances in Riemannian coordinate descent provide efficient update rules across diverse matrix manifolds (Stiefel, Grassmann, symmetric positive definite), while adaptive Riemannian gradient methods offer principled step-size selection on curved spaces \citep{han2024riemanniancd,xiao2025adaptive}. Integrating these techniques could yield geometry-respecting optimizers that maintain feasibility without projection while preserving the benefits of curvature-aware updates. Moreover, attention layers involve distinct projections --- queries, keys, and values --- that each fundamentally different computational roles \citep{vaswani2017attention}. Recent theoretical analysis by Yao et al.\ demonstrates that these matrices exhibit "unequal importance" during fine-tuning: optimizing the value matrix $\mathbf{W}_v$ yields significantly better performance than optimizing on the key matrix $\mathbf{W}_k$, with $\mathbf{W}_q$ and $\mathbf{W}_v$ together often matching or exceeding full attention tuning \citep{yao2025finetuning}. This asymmetry suggests that uniform treatment across attention components is suboptimal, and that norm choices reflecting the distinct geometry of query-key matching versus value projection could substantially improve training dynamics.

\paragraph{Theoretical Convergence Guarantees.}
While this thesis has emphasized practical effectiveness and intuitive understanding, establishing rigorous convergence rates for the methods discussed --- particularly in non-convex settings characteristic of deep learning --- remains largely open. Recent work on generalized gradient norm clipping provides convergence analysis under $(L_0, L_1)$-smoothness conditions for hybrid steepest descent methods \citep{pethick2025clipping,kornilov2025signop}. Extending such analyses to architecture-aware preconditioning would provide theoretical grounding for hyperparameter selection and potentially reveal conditions under which these methods offer provable advantages over standard approaches.

\paragraph{Computational Advances.}
The SVD computations underlying spectral preconditioning introduce non-trivial overhead, particularly for the large weight matrices in modern architectures. Randomized SVD methods with power iteration offer controllable accuracy-efficiency tradeoffs, while matrix sketching techniques enable memory-efficient gradient approximation \citep{feng2024dashSVD}. Investigating when such approximations suffice and whether approximation error can be absorbed into the optimization dynamics without degrading convergence could substantially reduce per-iteration costs. Furthermore, hardware-aware implementations that exploit structure in iterative orthogonalization may yield additional speedups on modern accelerators, making these principled methods practical at the largest scales \citep{dreier2022hardware}.

\clearpage

\clearpage
\bibliographystyle{plainnat}
\bibliography{refs}

\end{document}